\documentclass[10pt,twocolumn,letterpaper]{article}

\usepackage[pagenumbers]{cvpr} % To force page numbers, e.g. for an arXiv version

\definecolor{darkgreen}{rgb}{0, 0.7, 0}
\definecolor{darkblue}{RGB}{41,119,191}

\usepackage{multirow}
\usepackage{setspace}
\usepackage{algorithm}
\usepackage{algorithmic}
\usepackage[accsupp]{axessibility}  % Improves PDF readability for those with disabilities.

\definecolor{cvprblue}{rgb}{0.21,0.49,0.74}
\usepackage[pagebackref,breaklinks,colorlinks,allcolors=cvprblue]{hyperref}

\def\paperID{3102} % *** Enter the Paper ID here
\def\confName{CVPR}
\def\confYear{2025}

\title{HiPART: Hierarchical Pose AutoRegressive Transformer for \\ Occluded 3D Human Pose Estimation}

\author{
Hongwei Zheng\footnotemark[1], \  Han Li\footnotemark[1], \  Wenrui Dai\footnotemark[2], \  Ziyang Zheng\footnotemark[2], \  Chenglin Li, \  Junni Zou, \  Hongkai Xiong\\
Shanghai Jiao Tong University, Shanghai, China
\\
{\tt\small \{1424977324, qingshi9974, daiwenrui, zhengziyang, lcl1985, zoujunni, xionghongkai\}@sjtu.edu.cn }
}

\begin{document}
\maketitle
 
\renewcommand{\thefootnote}{\fnsymbol{footnote}}
\footnotetext[1]{Equal Contribution}
\footnotetext[2]{Corresponding Authors} 

\begin{abstract}

Existing 2D-to-3D human pose estimation (HPE) methods struggle with the occlusion issue by enriching information like temporal and visual cues in the lifting stage. In this paper, we argue that these methods ignore the limitation of the sparse skeleton 2D input representation, which fundamentally restricts the 2D-to-3D lifting and worsens the occlusion issue. To address these, we propose a novel two-stage generative densification method, named Hierarchical Pose AutoRegressive Transformer (HiPART), to generate hierarchical 2D dense poses from the original sparse 2D pose. Specifically, we first develop a multi-scale skeleton tokenization module to quantize the highly dense 2D pose into hierarchical tokens and propose a Skeleton-aware Alignment to strengthen token connections. We then develop a Hierarchical AutoRegressive Modeling scheme for hierarchical 2D pose generation. With generated hierarchical poses as inputs for 2D-to-3D lifting, the proposed method shows strong robustness in occluded scenarios and achieves state-of-the-art performance on the single-frame-based 3D HPE. Moreover, it outperforms numerous multi-frame methods while reducing parameter and computational complexity and can also complement them to further enhance performance and robustness.

% Current 2D-to-3D human pose estimation (HPE) struggles with occlusion issues, often relying on extra context like temporal and visual cues, which demand complex models and increase the computational cost. In this paper, we argue that the sparse 2D skeleton representation is ignored by existing works for addressing occlusion.  To address these, we propose a novel two-stage generative densification method, Hierarchical Pose AutoRegressive Transformer (HiPART), generating reliable hierarchical 2D dense poses from the original sparse 2D pose input. First, we introduce a multi-scale skeleton tokenization module to quantize the highly dense 2D pose into hierarchical tokens and  propose a Skeleton-aware Alignment to strengthen token connections. Second, we develop a Hierarchical AutoRegressive Modeling scheme for hierarchical 2D pose generation in a novel order specifically designed for skeletal structure data. With generated hierarchical poses as inputs for 2D-to-3D lifting, our method shows strong robustness in occluded scenarios and achieves state-of-the-art performance on the single-frame-based 3D HPE. Moreover, our method outperforms numerous multi-frame methods with reduced parameter and computational complexity and can be integrated into them to further enhance performance and robustness.
\end{abstract}

\section{Introduction}

3D human pose estimation (HPE) from monocular images has attracted wide attention due to its flexibility in adaptation on various devices. It is usually decoupled into the \emph{2D estimation} and \emph{3D lifting} stages in existing methods~\cite{martinez2017simple, zhao2019semantic, zheng2023actionprompt}, where the \emph{2D estimation} stage estimates 2D poses from input images via off-the-shelf 2D detectors~\cite{NewellYD16, chen2018cascaded, sun2019deep}, and subsequently, the \emph{3D lifting} stage achieves 2D-to-3D estimation with a lifting model. Such decoupling enables large-scale training with a large amount of 3D MOCAP data~\cite{martinez2017simple} by circumventing the scarce labeled 2D/3D ground truth poses for monocular images. However, the real-world application of 3D HPE faces a challenge of the occlusion problem (\emph{i.e.}, self-occlusion and object occlusion), where the unreliable occluded 2D poses could lead to corrupted 3D lifted poses.

\begin{figure}[!t]
\centering
\begin{subfigure}[b]{0.47\textwidth}
  \includegraphics[width=\textwidth]{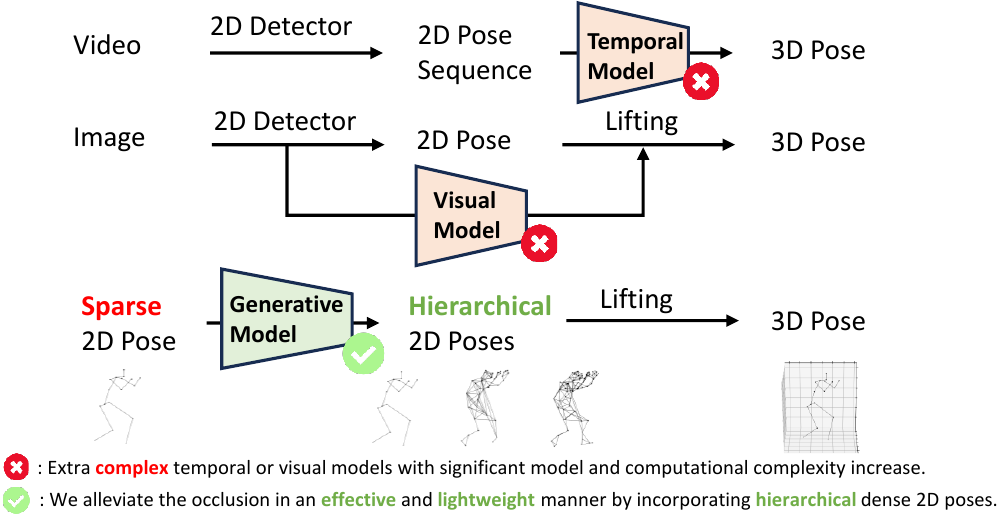}
\end{subfigure}\\
\vspace{5pt}
\begin{subfigure}[b]{0.42\textwidth}
  \includegraphics[width=\textwidth]{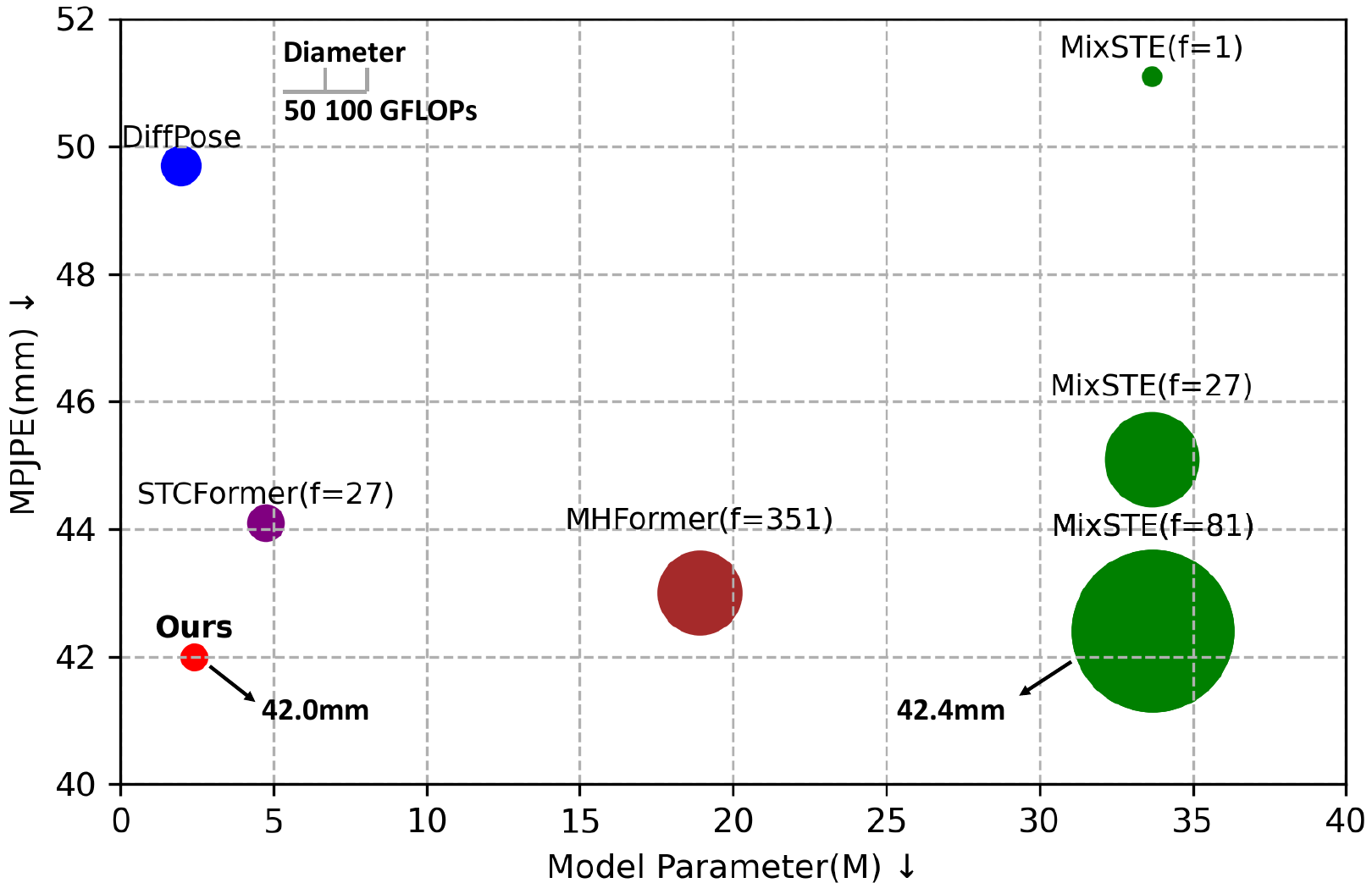}
\end{subfigure}
\vspace{-10pt}
\caption{\textbf{Top:} Comparison of temporal-based and visual-based methods with our densification approach at the framework level. These methods enrich information in the lifting stage, while ours address a more fundamental issue, \emph{i.e.}, the sparse 2D pose input.
\textbf{Bottom:} Comparison of parameters, GFLOPs, and MPJPE across various methods on Human3.6M.  The circle size indicates GFLOPs for inference. Our method achieves the  SOTA result with reduced complexity compared to the temporal-based methods.
}
\label{fig:motivation}
\vspace{-10pt}
\end{figure}
% 10-28-zhw: 模型复杂度和计算量比较。另外一种模型计算量和性能比较的图

Recent works have introduced additional information in the lifting stage to alleviate the occlusion issues. Some methods leverage topology priors (\emph{e.g.}, kinematic priors~\cite{radwan2013monocular, kundu2020kinematic, xu2020deep} and geometric constraints~\cite{chen2019unsupervised, yu2021towards})  to improve the structural understanding of joint correlations, but often prove ineffective in severe occlusion.  Others utilize enriched context (\emph{e.g.}, temporal \cite{pavllo20193d, zheng20213d, li2022mhformer} and visual cues~\cite{tekin2017learning, zhao2019semantic, liu2019feature, zhou2024lifting}) to improve the occluded joints inference, which require extra complex temporal or visual encoders with increased parameter and computational complexity  (see top of Fig.~\ref{fig:motivation}).  
In essence, all these methods primarily focus on enriching the information in the lifting stage while overlooking the inherent limitation of the current input 2D representation, \emph{\textbf{the sparse 2D skeleton representation}}, which fundamentally constrains the performance of 2D-to-3D lifting and worsens the occlusions.
Specifically, current 2D pose datasets typically represent the human body with few keypoints (\emph{e.g.}, 17 joints for Human3.6M~\cite{ionescu2013human3}). This sparse input for lifting inherently limits the ability to exploit local context. In this paper, we argue that \emph{developing a \textbf{hierarchical dense 2D skeleton representation} is essential for  exploring the intricate skeletal context, thus enhancing the robustness of occluded 2D-to-3D lifting.}

% Recent works have introduced additional information in the lifting stage to alleviate the occlusion issues. Some methods leverage topology priors (\emph{e.g.}, kinematic priors~\cite{radwan2013monocular, kundu2020kinematic, xu2020deep} and geometric constraints~\cite{chen2019unsupervised, yu2021towards})  to improve the structural understanding of joint correlations, but often prove ineffective in severe occlusion.  Others utilize enriched context (\emph{e.g.}, temporal \cite{pavllo20193d, zheng20213d, li2022mhformer} and visual cues~\cite{tekin2017learning, zhao2019semantic, liu2019feature, zhou2024lifting}) to improve the occluded joints inference. However, these methods require extra complex temporal or visual encoders with significant parameter and computational complexity increase (see top of Fig.~\ref{fig:motivation}).  
% In addition, existing works have ignored a key challenge of the occlusion issue: \emph{\textbf{the sparse 2D skeleton representation}}. Specifically, current 2D pose datasets typically represent the human body with few keypoints (\emph{e.g.}, 17 joints for Human3.6M~\cite{ionescu2013human3}). This sparse input for lifting inherently limits the ability to exploit local context, making occlusion even more difficult to address. In this paper, we argue that \emph{developing a \textbf{hierarchical dense 2D skeleton representation} is essential for  exploring the intricate skeletal context, thus enhancing the robustness of occluded 2D-to-3D lifting.}

\begin{figure}[t]
\setlength{\abovecaptionskip}{0pt}
\centering
\includegraphics[width=0.9\linewidth]{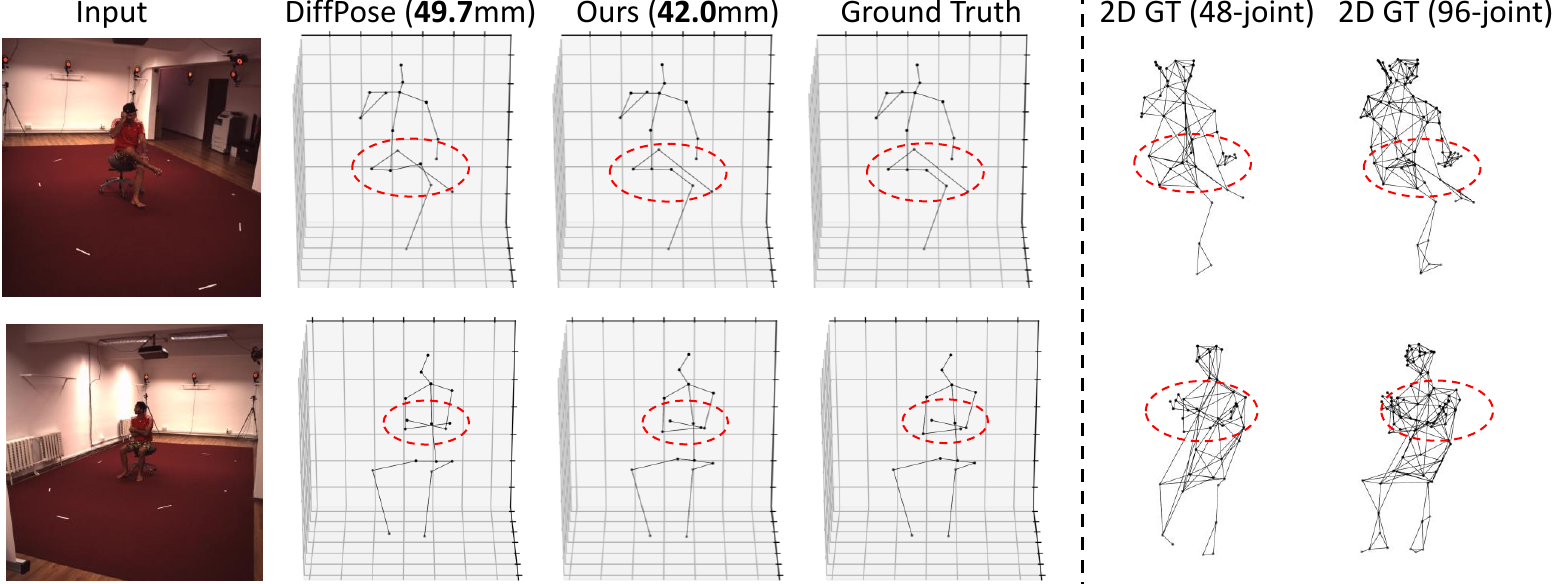}
    \caption{ Visualization of reconstructed 3D poses and hierarchical 2D poses on Human3.6M under occlusions. Ours outperforms DiffPose~\cite{gong2023DiffPose} due to the rich skeletal context in hierarchical poses.
    % Our method outperforms in predicting occluded joints (highlighted by the red circles) compared to DiffPose~\cite{gong2023DiffPose}. 
    % The hierarchical joints provide richer information around the invisible joints, effectively mitigating the occlusion problem.
    }
    \label{fig:motivation4}
    \vspace{-10pt}
\end{figure}

We conduct a toy experiment to demonstrate the benefit of this hierarchical representation. 
We collect the dataset  of hierarchical dense 2D poses by progressively coarsening the 3D ground truth mesh of Human3.6M (6890 vertices $\rightarrow$ 96 joints $\rightarrow$ 48 joints) and projecting them into the 2D pixel space. When integrating these hierarchical dense 2D poses into the 2D-to-3D lifting stage, MPJPE is significantly improved by 55\% from 37.6mm to 17.5mm, as shown in Appendix~\textcolor{red}{E}). This indicates the significance of richer skeletal information. For example, as shown in the bottom of Fig.~\ref{fig:motivation4}, when the right wrist is occluded,
% the sparse pose offers limited assistance  with only one elbow joint available, while the denser pose includes extra arm joints, capturing more detailed structure around the occlusion and aiding in predicting the occluded right wrist joint.
the sparse 2D pose only provides one elbow joint for reference,  while the denser 2D poses offer multiple arm joints to help predict the occlusion region.
However, acquiring such hierarchical dense 2D poses remains a challenge, as the 3D ground truth mesh is not available in real-world 3D HPE scenarios.
% Previous methods~\cite{choi2020pose2mesh, li2021hierarchical} combine the densification and lifting in one stage, resulting in inaccurate 2D and 3D predictions (see Fig.~\ref{fig:2D3DMPJPE}). Thus, \emph{there is a need to design a more effective densification approach to achieve more precise 2D results.}
% 10-31-zhw: 后面再加一句话过渡：需要更准确的densification方法
% Previous methods We compare the prediction accuracy of the fine 2D pose (96 joints) and lifted sparse 3D pose (17 joints) across various methods. Compared to 

Inspired by recent autoregressive models~\cite{van2017neural, razavi2019generating, esser2021taming,tian2024visual,li2024frequencyaware}, in this paper, we propose a novel hierarchical 2D pose densification method named Hierarchical Pose AutoRegressive Transformer (HiPART) to address the challenge. HiPART obtains reliable hierarchical dense 2D poses conditioned on the original sparse 2D pose input in a generative fashion with two main modules. First, we design a Multi-Scale Skeletal Tokenization (MSST) module to progressively quantize a highly dense 2D pose into hierarchical discrete tokens via VQ-VAE~\cite{van2017neural}, and introduce a Skeleton-aware Alignment strategy to strengthen connections across multi-scale tokens. Second, we develop a Hierarchical AutoRegressive Modeling (HiARM) scheme for hierarchical dense 2D pose generation. Unlike the conventional next-token prediction widely used in image and video generation~\cite{chen2020generative, esser2021taming, yu2023language}, we present novel \emph{sparse-to-dense} and \emph{center-to-periphery} strategies specifically designed for the non-Euclidean skeletal structure and show their efficiency. 
Therefore, HiPART generates hierarchical 2D poses to provide enhanced local context for subsequent 2D-to-3D lifting.

We perform extensive experiments to validate the effectiveness and efficiency of HiPART. Specifically, we show that, using only a \emph{vanilla spatial transformer} for lifting, HiPART achieves state-of-the-art performance in the single-frame setting on various 3D HPE benchmarks, including Human3.6M~\cite{ionescu2013human3}, 3DPW~\cite{von2018recovering}, and 3DPW-Occ~\cite{zhang2020object},  with strong robustness in occluded scenarios. Compared to methods that rely on complex temporal encoders to incorporate temporal cues, HiPART is superior or comparable with significantly reduced complexity, as shown in the bottom of Fig.~\ref{fig:motivation}. Note that HiPART is orthogonal to existing temporal methods and can integrate these methods to further enhance performance and robustness. The contributions of this paper are summarized below.

\begin{itemize}
\item To our best knowledge, we achieve the first hierarchical dense 2D pose generation scheme to address the occlusion problem for 3D HPE from single monocular image. 
    % We demonstrate the effectiveness of hierarchical poses for 3D HPE and propose an densification and lifting framework to address occlusion issues.
    % which leverages hierarchical information
    \item We develop a multi-scale skeleton quantization model via VQ-VAE along with a Skeleton-aware Alignment to enhance token connections.
    \item  We design a novel hierarchical pose autoregression strategy specifically for non-Euclidean skeleton structures.
    \item We employ the proposed 2D pose generation scheme in 2D-to-3D lifting to achieve state-of-the-art performance on the single-frame setting and outperforms numerous multi-frame methods.
    % using only a vanilla spatial transformer for lifting.
\end{itemize}

\begin{figure*}[t]
\setlength{\abovecaptionskip}{0pt}
    \centering
    \includegraphics[width=0.9\linewidth]{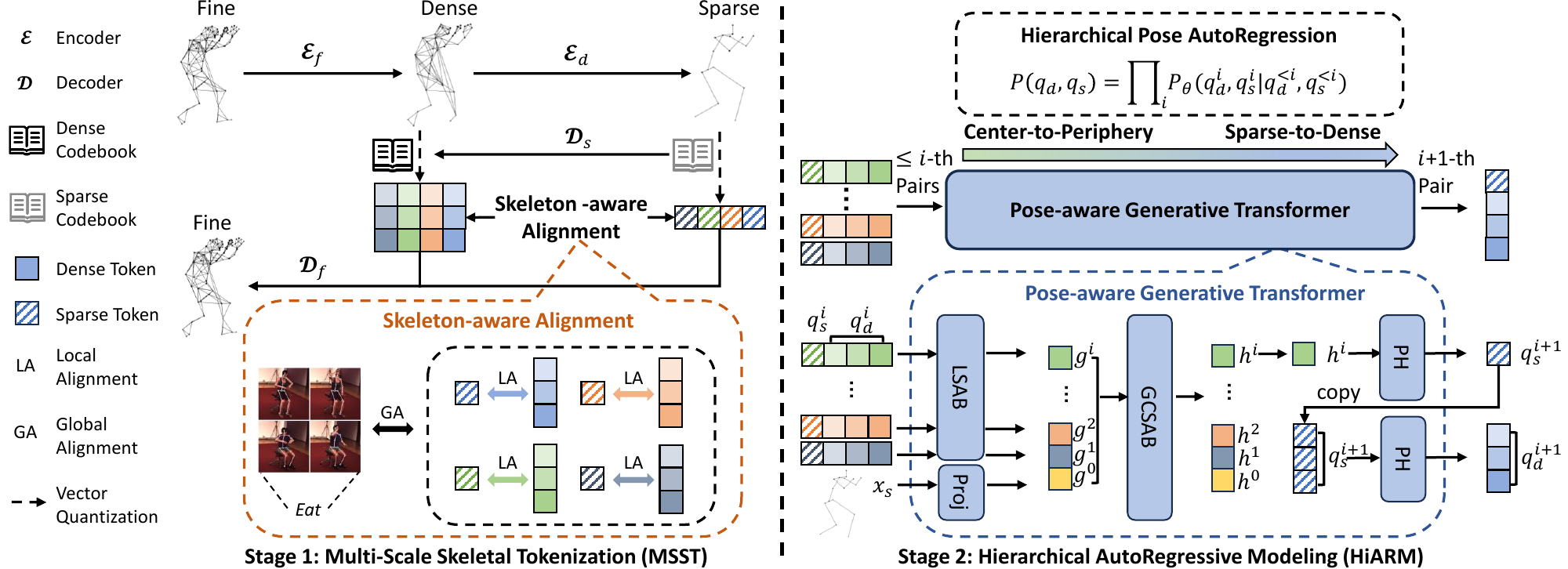}
    \caption{
    % The overview of the two-stage hierarchical 2D pose densification method, named Hierarchical Pose AutoRegressive Transformer (HiPART). 
    % In Stage 1, the Multi-Scale Skeletal Tokenization (MSST) module progressively quantizes the fine 2D pose into hierarchical discrete tokens. In Stage 2, we propose a Hierarchical AutoRegressive Modeling (HiARM) scheme to generate hierarchical dense 2D poses.
    The overview of the two-stage generative densification method, Hierarchical Pose AutoRegressive Transformer (HiPART). In Stage 1, the MSST module progressively quantizes the fine 2D pose into hierarchical tokens with Skeleton-aware Alignment. We omit the reconstruction of the dense 2D pose $\hat{\mathbf{x}}_d$ for simplicity.
    In Stage 2, we propose a HiARM scheme to generate $(i+1)$-th pair tokens based on a series of indices less than $i+1$. 
    One pair of discrete tokens contains  a single sparse token and $r$ dense tokens corresponding to the related part.
    Finally, generated hierarchical 2D poses are fed to a vanilla spatial transformer for subsequent 2D-to-3D lifting.
   }
    \label{fig:framework}
    \vspace{-10pt}
\end{figure*}

\section{Related Work}
% 遮挡问题 in 3DHPE：单帧利用结构信息、时序、多假设、图片信息
% hierarchy in HPE
% 生成式：VQ-VAE、PCT、difpose

\noindent\textbf{Occlusions in 3D HPE.} 
% Occlusion presents a significant challenge in 3D HPE, as missing or partially visible body parts often lead to inaccurate pose reconstruction~\cite{sarandi2018robust, cheng2019occlusion, kocabas2021pare, wang2022ocr}. 
Existing approaches typically incorporate additional information in the lifting stage. Topology priors (\emph{e.g.}, kinematic priors~\cite{radwan2013monocular, kundu2020kinematic, xu2020deep}  and geometric constraints~\cite{chen2019unsupervised, yu2021towards}) are utilized to enhance joint correlations, often failing in severely occluded scenarios. Other approaches introduce enriched context( \emph{e.g.}, temporal context~\cite{pavllo20193d, zheng20213d, li2022mhformer, tang20233d} and visual cues~\cite{liu2019feature, zhao2022graformer, zhou2024lifting}) to improve the prediction accuracy for occluded joints. 
While highly effective, these methods consist of advanced temporal or image encoders, significantly increasing model and computational complexity. 
We alleviate the occlusion problem in an effective and lightweight manner by incorporating hierarchical dense 2D poses as input. It addresses the sparse 2D pose input issue, which is overlooked by existing works.

\noindent\textbf{Hierarchy in 3D HPE.} Hierarchical information is essential in Human Mesh Recovery (HMR)~\cite{omran2018neural, kolotouros2019learning, kocabas2021pare,zhang2025hawor}, as the human body can be represented at multi-level, from  sparse joints to dense vertices, which contains rich skeletal context and can help address occlusions in 3D HPE. For instance, Pose2Mesh~\cite{choi2020pose2mesh} uses a learnable transformation matrix to estimate poses in a coarse-to-fine manner. HGN~\cite{li2021hierarchical} employs a hierarchical network to reconstruct poses, transitioning from sparse to dense representations. However, both methods regard the hierarchical 3D pose prediction as an auxiliary task, neglecting the accuracy of the hierarchical 2D poses, and leading to minimal performance gains. We propose a generative densification approach that enhances 2D hierarchical pose estimation and 3D lifting precision.

\noindent\textbf{Generative Models for 3D HPE.} Generative models have shown promising results in various domains, including image, text, and audio generation~\cite{ho2020denoising, ho2022cascaded, chen2023diffusiondet, huang2022prodiff, yang2023diffsound}, and have recently been applied in 3D HPE. For example, diffusion models progressively refine pose distributions, reducing uncertainty during estimation~\cite{choi2023diffupose, gong2023DiffPose, holmquist2023diffpose}. However, they suffer from low inference speed due to reliance on numerous sampling steps. 
% Intermediate results between denoising steps are also not reusable, unlike autoregressive (AR) models. 
A vector-quantized-based method, \emph{i.e.}, PCT~\cite{geng2023human},  uses a two-stage process to construct a codebook capturing sub-structures and casts pose estimation as a classification task. While effective, the classification approach struggles to fully model the latent distribution of discrete tokens. We address this in an auto-regression manner in Stage 2
and introduce Skeleton-aware Alignment to strengthen token connections. 
Besides, our method has a faster generation speed than diffusion models due to the posed-based auto-regression. More discussion is in the Appendix~\textcolor{red}{D}.

\section{Collection of Hierarchical 2D Poses}
The goal of our method is to generate reliable hierarchical 2D poses with denser joints conditioned on the original 
 sparse 2D pose $\mathbf{x}_s \in \mathbb{R}^{J_s \times 2}$ with $J_s$ joints, benefiting subsequent  2D-to-3D lifting.  However, existing datasets lack hierarchical dense 2D poses that could serve as ground truth for training. To address this, we use the 3D human mesh from Human3.6M~\cite{ionescu2013human3} and apply the dense vertices coarsening method~\cite{choi2020pose2mesh} along with camera projection to create the dense and fine (\emph{i.e.}, highly dense) 2D poses,  denoted as $\mathbf{x}_d \in \mathbb{R}^{J_d \times 2}$ and $\mathbf{x}_f \in \mathbb{R}^{J_f \times 2}$, where $J_f>J_d>J_s$.  The detailed process is described in the Appendix~\textcolor{red}{B}.

\section{Hierarchical Pose AutoRegressive Transformer (HiPART)}
% Previous methods adopt a single-stage manner to upsample the hierarchical joints, leading to inaccurate estimation. To address this issue, 
% The goal of HiPART is to generate reliable hierarchical dense 2D poses from the original 
%  sparse 2D pose, allowing for subsequent  2D-to-3D lifting. 
 % Fig.~\ref{fig:framework} shows the two-stage training of our HiPART. 
 The overview of our HiPART is  shown in Fig.~\ref{fig:framework}. We first learn the Multi-Scale Skeletal Tokenization (MSST) module to quantize the fine $\mathbf{x}_f \in \mathbb{R}^{J_f \times 2}$ into hierarchical discrete tokens, \emph{i.e.,} sparse tokens $\mathbf{q}_s \in \mathbb{Z}^{J_s}$ and dense token $\mathbf{q}_d \in \mathbb{Z}^{J_d}$, 
 % which represents skeletal information of different levels. 
 where different colors in Fig.~\ref{fig:framework} indicate different parts of the human body, and different saturations represent the number ($r$) of dense tokens.
 % in the group of $r$ dense tokens and one sparse token.
 Then, we learn a pose-aware generative transformer by the Hierarchical AutoregRessive Modeling (HiARM) scheme to generate hierarchical tokens conditioned on the sparse 2D pose $\mathbf{x}_s$. With the decoders of HiPART, we can reconstruct the hierarchical dense 2D poses from the hierarchical tokens $\mathbf{q}_s$ and $\mathbf{q}_d$. 
 % The pseudo code of HiPART is in the Appendix.~\textcolor{red}{C}.

% During inference, we adopt the HiARM decoding meth 
% 10-23-zhwdone: 句式and有点多

% \subsection{Stage 1: Learning hierarchical pose-based vector}
\subsection{Multi-Scale Skeletal Tokenization (MSST)}
% In Stage 1, we introduce Multi-Scale Skeletal Tokenization (MSST) to learn hierarchical codebooks and propose Skeleton-aware Alignment to enhance token connection.
\textbf{Multi-Scale Skeletal Quantized-Autoencoder.}
% Our approach employs multi-scale vector quantization (VQ) to encode the fine pose into two levels of discrete tokens: sparse tokens capture coarse-grained information, while dense tokens focus on joint-level details. 
% We utilize two pathways to learn these hierarchical tokens.
We use an architecture similar to VQ-VAE-2~\cite{razavi2019generating} to construct our multi-scale skeletal quantized autoencoder. Encoders and decoders are implemented with MLP-Mixer~\cite{tolstikhin2021mlp}, which is well-suited for skeletal data. Specifically, 
two  encoders $\mathcal{E}_f(\cdot)$ and $\mathcal{E}_d(\cdot)$  sequentially project the ground truth $\mathbf{x}_f$ into dense and sparse pose embeddings, $ \mathbf{z}_d \in \mathbb{R}^{J_d\times D}$ and $ \mathbf{z}_s \in \mathbb{R}^{J_s\times D}$, where $D$ is the embedding dimension.   

Next, $\mathbf{z}_s$ is quantized into sparse tokens $ \mathbf{q}_s \in \mathbb{Z}^{J_s} $ via a quantizer $ \mathcal{Q}(\cdot) $ using the sparse codebook $ \mathbf{C}_s \in \mathbb{R}^{K_s\times D} $, which contains $K_s$  learnable vectors. Each element of $ \mathbf{q}_s$ then looks up $ \mathbf{C}_s $ to obtain $ \mathbf{\hat{z}}_s$, \emph{i.e.}, the approximation of $\mathbf{z}_s$:

\begin{equation}
\mathbf{q}_s = \mathcal{Q}(\mathbf{z}_s;\mathbf{C}_s), \quad  \mathbf{\hat{z}}_s = \mathbf{C}_s[\mathbf{q}_s].
\end{equation}

Subsequently, $ \mathbf{\hat{z}}_s$ is upsampled into $ \mathbf{\hat{z}}^\prime_s \in \mathbb{R}^{J_d\times D} $ using decoder $\mathcal{D}_s(\cdot)$. This upsampling embedding is then concatenated with $ \mathbf{z}_d $ to form $ \mathbf{z}^\prime_d $, which is quantized into dense tokens $ \mathbf{q}_d \in \mathbb{Z}^{J_d} $ using the dense codebook $ \mathbf{C}_d \in \mathbb{R}^{K_d\times D} $, which contains $K_d$ vectors. The approximated dense pose embeddings $\mathbf{\hat{z}}_d$  is obtained by looking up $ \mathbf{C}_d $ from $ \mathbf{q}_d$.

\begin{equation}
\mathbf{q}_d = \mathcal{Q}(\mathbf{z}^\prime_d;\mathbf{C}_d), \quad \mathbf{\hat{z}}_d = \mathbf{C}_d[\mathbf{q}_d],
\end{equation}
% where $\mathbf{C}_s$  and $\mathbf{C}_d$ are distinct to reflect different levels of information.
Finally, the  reconstructed dense pose $ \hat{\mathbf{x}}_d $ is decoded  using decoder $\mathcal{D}_d(\cdot)$ given $\mathbf{\hat{z}}_s$, and  the  reconstructed fine pose $ \hat{\mathbf{x}}_f $ is decoded  using decoder $\mathcal{D}_f(\cdot)$ given $\mathbf{\hat{z}}_s$ and $\mathbf{\hat{z}}_d$.

% the dense approximation embedding is decoded into the reconstructed dense pose $ \hat{\mathbf{x}}_d $ via decoder $\mathcal{D}_d(\cdot)$, and the combined sparse and dense approximation embeddings are used to reconstruct the fine pose $ \hat{\mathbf{x}}_f $ via decoder $\mathcal{D}_f(\cdot)$. 
% All the encoders and decoders consist of MLP-Mixer~\cite{tolstikhin2021mlp} blocks.
% % zhwdone: sparse对dense是使用complement还是-

\noindent\textbf{Skeleton-aware Alignment.}
To enhance the tokenization process for skeletal data, we propose a Skeleton-aware Alignment strategy. This strategy incorporates skeletal structure into the tokenization process,  improving the mutual connections across different skeleton levels.
% The hierarchical tokens are learned independently through two distinct pathways, lacking mutual connections. To address this, we propose two Skeleton-aware Alignment strategies.
% 10-22-zhw: 关于现有codebook的缺点的描述还需要再想一下，pose structure有点奇怪

\noindent\emph{Part-wise Local Alignment (LA):} 
% The sparse and dense tokens lack necessary connections, leading to a disjointed relationship between the codebooks. To overcome this, 
The part-wise Local Alignment is introduced to align the sparse and dense tokens within the same local body part, forming a more unified representation. The Info Noise Contrastive Estimation (Info-NCE) loss is used to match part-wise pairs:
\begin{equation}
\mathcal{L}_{local} = \sum_{i=1}^{J_s} -\log\left(\frac{\exp(\hat{\mathbf{z}}_s^i \cdot \operatorname{avg}(\hat{\mathbf{z}}_d^i)/\tau)}{\sum_{k=1}^{J _s}\exp(\hat{\mathbf{z}}_s^i \cdot \operatorname{avg}(\hat{\mathbf{z}}_d^k)/\tau) }\right),
    \label{equ:3}
\end{equation}
where $\tau$ is the temperature parameter,  $ \operatorname{avg}(\cdot)$
% = \frac{1}{r}\sum_{k=1}^{r}\hat{\mathbf{z}}_d^{(i,k)}$ 
is the average function. The $r$ approximated dense pose embeddings $\hat{\mathbf{z}}_d^{i} = \{\hat{\mathbf{z}}_d^{(i,j)}\}_{j=1,2,\dots,r}$  correspond to the related part of single $\hat{\mathbf{z}}_s^i$, where the ratio $r=J_d/J_s$.

\noindent\emph{Action-wise Global Alignment (GA):} 
 We further introduce an action-wise Global Alignment to align the hierarchical tokens with the action label of the pose (contained in the Human3.6M~\cite{ionescu2013human3} dataset), thus offering consistent semantic information for the discrete spaces. We define the global loss $\mathcal{L}_{global}$ as a cross-entropy loss between the predicted classification vector $\mathbf{p}_A$ and the action label $\mathbf{y}_A$:
\begin{align}
\mathbf{p}_A &= \mathcal{P}_A(\operatorname{concat}(\hat{\mathbf{z}}_s, \hat{\mathbf{z}}_d)),\\
%\end{equation}
%\begin{equation}
\mathcal{L}_{global} &= \operatorname{CrossEntropy}(\mathbf{y}_A, \mathbf{p}_A),
    \label{equ:4}
\end{align}
where $\mathcal{P}_A(\cdot)$ is the action projector composed of two MLP-Mixer blocks, $\operatorname{concat}(\cdot)$ is the concatenate operation.

\noindent\textbf{Loss Function.} 
The Stage 1 loss $\mathcal{L}_1$ includes the reconstruction loss for the poses and hierarchical pose embeddings, as well as the loss of Skeleton-aware Alignment.
% The overall loss is as follows:
% \begin{equation}
% \begin{split}
% \mathcal{L}_1 = &\| \mathbf{x}_f - \hat{\mathbf{x}}_f \|_2^2 + \| \mathbf{x}_d - \hat{\mathbf{x}}_d \|_2^2 + \sum_{j \in \{s,d\}}  \Big(\|\operatorname{sg}(\mathbf{z}_j) - \hat{\mathbf{z}}_j\|_2^2 + \\ 
% &\beta \|\mathbf{z}_j - \operatorname{sg}(\hat{\mathbf{z}}_j)\|_2^2\Big) +  \lambda_{l} \cdot \mathcal{L}_{local} +  \lambda_{g} \cdot \mathcal{L}_{global}, \label{equ:5}
% \end{split} 
% \end{equation}
\begin{align}
\mathcal{L}_1 =&  \| \mathbf{x}_f - \hat{\mathbf{x}}_f \|_2^2 + \| \mathbf{x}_d - \hat{\mathbf{x}}_d \|_2^2 + \sum_{j \in \{s,d\}}  \Big(\|\operatorname{sg}(\mathbf{z}_j)\!-\! \hat{\mathbf{z}}_j\|_2^2 \nonumber \\
& + \beta \|\mathbf{z}_j\!-\! \operatorname{sg}(\hat{\mathbf{z}}_j)\|_2^2\Big) \!+\!  \lambda_{l} \cdot \mathcal{L}_{local} \!+\!  \lambda_{g} \cdot \mathcal{L}_{global}, \label{equ:5}
\end{align}

where $ \beta $ controls the commitment loss,  $\lambda_{l}$ and $\lambda_{g}$  are  weighting factors for the local and Global Alignment losses, $sg()$ is the stop gradient operation.
Codebooks are updated via Exponential Moving Average.

\subsection{Hierarchical AutoRegressive Modeling (HiARM)}
%10-21-lh 和生成的关系
% Stage 2 generates hierarchical tokens in an auto-regression manner. 
% Compared to the classification manner~\cite{geng2023human}, the generative method can sufficiently model the latent distribution of tokens.
% % 10-25-zhw: 是否需要提PCT的劣势
% We first define a pose-based auto-regressive objective function and then present the model details.
In Stage 2, we propose Hierarchical AutoRegressive Modeling (HiARM) for hierarchical dense 2D pose generation.
We first define a hierarchical pose auto-regressive objective function and then present the model structure details.
% \vspace{-10pt}

\noindent\textbf{Objective Function.}
% Given the hierarchical tokens$\{\mathbf{q}_s,\mathbf{q}_d\}$ obtained in MSST, we adopt the auto-regression manner to generate discrete tokens. 
We generate the discrete tokens in an auto-regression manner based on the learned codebooks in MSST.
Our objective function aims at maximizing the log-likelihood of the $i$-th pair discrete tokens based on a series of indices less than $i$:
% The likelihood can be defined as follows:
\begin{equation}
\mathbb{P}_{\theta}(\mathbf{q}_d, \mathbf{q}_s) = \prod_{i=1}^{J_s} \mathbb{P}_{\theta}(\mathbf{q}_d^{i}, q_s^{i} | \mathbf{q}_d^{< i}, \mathbf{q}_s^{< i} )
\end{equation}

\begin{figure}[t]
\setlength{\abovecaptionskip}{0pt}
    \centering
    \includegraphics[width=0.8\linewidth]{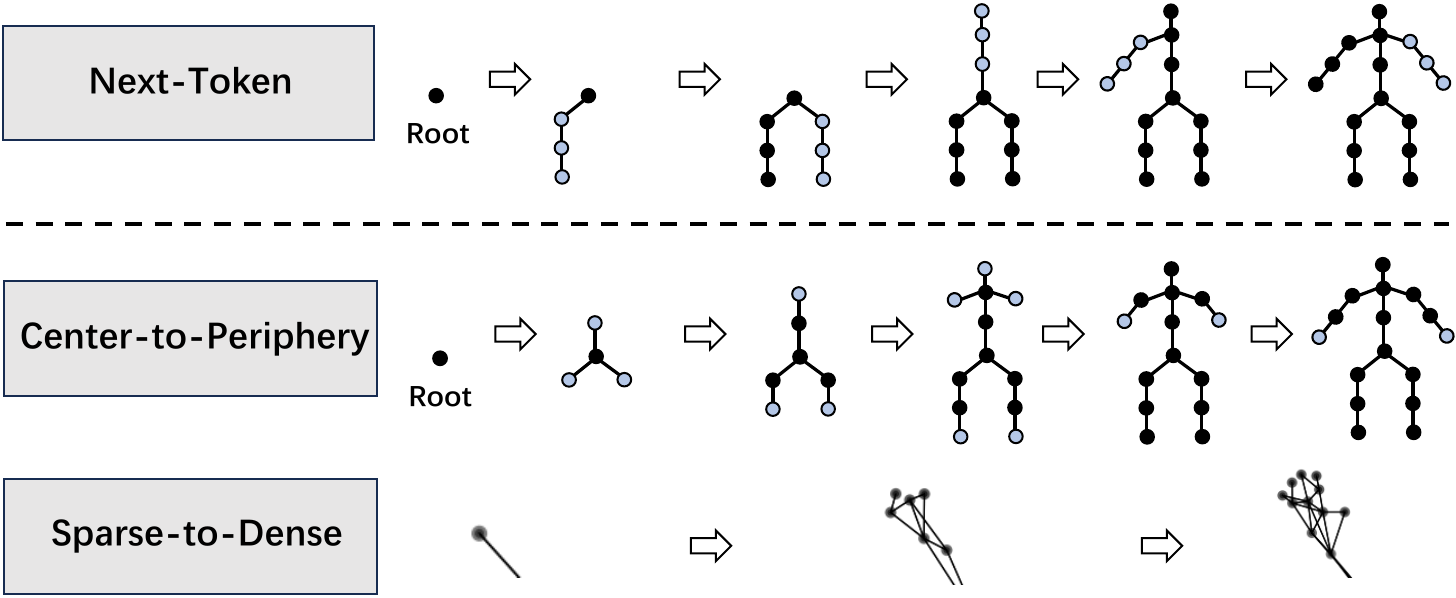}
    \caption{Standard auto-regressive modeling (\textbf{Top}) vs  Our proposed hierarchical pose auto-regressive modeling (\textbf{Bottom}).}
    \label{fig:order}
    \vspace{-16pt}
\end{figure}

However, we argue that the commonly used next-token prediction order in image generation~\cite{chen2020generative, esser2021taming} is unsuitable for pose generation. This is because it disregards the skeletal topology of human poses and suffers from slow generation speeds. As shown in Fig.~\ref{fig:order}, we propose a novel autoregressive manner specifically designed for the non-Euclidean skeletal structure.
% to exploit kinematic information and accelerate the generation process. 

\noindent\emph{Center-to-periphery.} \cite{li2023pose} has revealed that the joints farther from the root joint tend to exhibit larger uncertainty. Inspired by it, we  generate joints  progressively from the center to the periphery, thus ensuring joints with lower uncertainty are first generated and alleviating the depth uncertainty.

% center joints helps to reduce the uncertainty of the periphery joints.
% Fig.~\ref{fig:order} illustrates that we generate the joint based on the distance towards the root joint in a "center to periphery" manner.

\noindent\emph{Sparse-to-dense.} Sparse tokens $\mathbf{q}_s$ capture  comprehensive and coarse-grained information, while dense tokens $\mathbf{q}_d$ represent localized and fine-grained details. Based on the assumption that the local details of the pose can be generated conditionally on the global summary, we can transform the expression of the likelihood into the following form:
\begin{equation}
\mathbb{P}_{\theta}(\mathbf{q}_d^{i}, q_s^{i} | \mathbf{q}_d^{< i}, \mathbf{q}_s^{< i} ) = \mathbb{P}_{\theta}(\mathbf{q}_d^{i} |q_s^{i}, \mathbf{q}_d^{< i}, \mathbf{q}_s^{< i} ) \cdot \mathbb{P}_{\theta}(q_s^{i} | \mathbf{q}_d^{< i}, \mathbf{q}_s^{< i} ),
\end{equation}
where $r$ dense tokens are generated in parallel conditionally on the corresponding sparse token. We simplify the hierarchical generation process, reducing the original $1+r$ steps to $2$ steps, thus speeding up the pose generation process.

\noindent\textbf{Model Structure.}
% The pose-aware generative transformer consists of Local Self-Attention Blocks (LSAB), Global Causal Self-Attention Blocks (GCSAB), and Prediction Head (PH), which collectively model hierarchical tokens.
We detail the components of the pose-aware generative transformer as follows: 

% 10-23-zhwdone: 体现local和global的关系
\noindent\emph{Local Self-Attention Blocks (LSAB).} The approximation of $i$-th pair sparse and dense 
pose embeddings $\hat{\mathbf{z}}^i_s $ and $\{\hat{\mathbf{z}}^{(i, j)}_d\}_{j=1,2,\dots,r} $ are refined with the LSAB, which models hierarchical tokens within the local body part, fully leveraging the information contained in multi-scale embeddings:
\begin{equation}
\{\mathbf{g}^i_j\}_{j=0,1,\dots,r} \gets \operatorname{LSAB}(\hat{\mathbf{z}}^i_s, \hat{\mathbf{z}}^{(i,1)}_d, \dots, \hat{\mathbf{z}}^{(i,r)}_d),
\label{equ:8}
\end{equation}
where we adopt a bidirectional transformer with a self-attention block followed by an MLP block. 
The refined embeddings $\{\mathbf{g}^i_j\}_{j=0,1,\dots,r}$ are then averaged to obtain the average refined embedding $\mathbf{g}^i$.

\noindent\emph{Global Causal Self-Attention Blocks (GCSAB).} GCSAB generates a sequence of hidden embeddings $\{\mathbf{h}^k\}_{k=0,1,...,i} $, capturing spatial relationships across the overall pose.
Instead of using traditional next-token prediction order, we model the discrete tokens from the center to the periphery:
\begin{equation}
\{\mathbf{h}^k\}_{k=0,1,...,i} \gets \operatorname{GCSAB}(\mathbf{g}^0, \mathbf{g}^1, \ldots, \mathbf{g}^{i}),
\label{equ:9}
\end{equation}
where $\mathbf{g}^0$ represents the projection of sparse 2D pose $\mathbf{x}_s$ serving as the start-of-sentence (SOS) token.
% zhwdone: 解释为什么叫global

\noindent\emph{Prediction Head (PH).}
This module predicts sparse and dense tokens in a hierarchical order, starting from the sparse to the dense. The sparse token is generated from the hidden embedding, followed by the parallel generation of $r$ dense tokens based on the corresponding sparse token:
\begin{equation}
\{p^{i+1}_j\}_{j=0,1,\dots,r} \gets \operatorname{PH}(\mathbf{h}^i,\hat{\mathbf{z}}_s^i, \dots, \hat{\mathbf{z}}_s^i),
\label{equ:10}
\end{equation}
where $p^{i+1}_{0}$ and $\{p^{i+1}_j\}_{j=1,\dots,r}$ denote the $(i+1)$-th pair predicted vectors of the sparse token and dense tokens. PH consists of a self-attention block with an MLP block.

Finally, the Stage 2 loss $\mathcal{L}_2$ model the latent distributions of sparse and dense tokens via the cross-entropy loss:
\begin{equation}
\mathcal{L}_2\!=\!\operatorname{CrossEntropy}(\mathbf{q}_s,\mathbf{p}_s) \!+\! \lambda_d \cdot \operatorname{CrossEntropy}(\mathbf{q}_d,\mathbf{p}_d),
\label{equ:11}
\end{equation}
where $\mathbf{p}_s$ and $\mathbf{p}_d$ denote the predicted vectors for all sparse and dense tokens, and $\lambda_d$ controls the dense token loss.

\subsection{Inference Process}
As shown in Algorithm~\ref{alg:inference}, we first use the projection of the sparse 2D pose as a class-conditional SOS token to predict the initial sparse and dense tokens. The LSAB then encodes them for subsequent generations. This process is repeated
in a hierarchical pose auto-regression manner 
until all discrete tokens are generated. Finally, we reconstruct the hierarchical dense 2D poses via the MSST decoder.

\setlength{\textfloatsep}{15pt}
\renewcommand{\algorithmicrequire}{\textbf{Input:}}
\renewcommand{\algorithmicensure}{\textbf{Output:}}
\begin{algorithm}[ht]
\caption{Inference process.}
\label{alg:inference}
\begin{algorithmic}[1]

\REQUIRE The sparse 2D pose $\mathbf{x}_s$, the sparse and dense codebooks $\mathbf{C}_s$ and $\mathbf{C}_d$.
\ENSURE The reconstructed dense and fine 2D pose $\hat{\mathbf{x}}_d$, $\hat{\mathbf{x}}_f$.\\
\FOR{$i=0$ to $J_s$} 
\IF{$i=0$}  
\STATE \textbf{\{Obtain initial tokens based on a SOS token\}}
% \STATE Obtain class-conditional SOS token: $\mathbf{g}^0 \gets \mathcal{P}_c(\mathbf{x}_s)$
\STATE $\mathbf{g}^0 \gets \mathcal{P}_c(\mathbf{x}_s)$
\STATE $\mathbf{h}^0 \gets \operatorname{GCSAB}(\mathbf{g}^0) $
\STATE $q_s^{1}, \{q_d^{(1, j)}\}_{j=1,2,\dots,r}     \gets \operatorname{PH} (\mathbf{h}^0, \hat{\mathbf{z}}_s^0, \dots,  \hat{\mathbf{z}}_s^0) $
\ELSE
% \STATE \COMMENT{\textbf{Repeat in a hierarchical pose auto-regression manner}}
\STATE \textbf{\{Center-to-periphery and Sparse-to-dense\}}
\STATE $\hat{\mathbf{z}}_s^i \gets \mathbf{C}_s(q_s^i)$, $\hat{\mathbf{z}}_d^i \gets \mathbf{C}_d(\mathbf{q}_d^i)$
\STATE $\{\mathbf{g}^i_j\}_{j=0,1,\dots,r} \gets \operatorname{LSAB}(\hat{\mathbf{z}}^i_s, \hat{\mathbf{z}}^{(i,1)}_d, \dots, \hat{\mathbf{z}}^{(i,r)}_d) $
\STATE $\mathbf{g}^i = \operatorname{avg}(\mathbf{g}^i_0, \mathbf{g}^i_1, \dots, \mathbf{g}^i_r)$
\STATE $ \{\mathbf{h}^k\}_{k=0,1,\dots,i} \gets \operatorname{GCSAB}(\mathbf{g}^0, \mathbf{g}^1, \dots, \mathbf{g}^{i})$
\STATE $q_s^{i+1}, \{q_d^{(i+1, j)}\}_{j=1,2,\dots,r}   \gets \operatorname{PH} (\mathbf{h}^i, \hat{\mathbf{z}}_s^i, \dots,  \hat{\mathbf{z}}_s^i) $
\ENDIF
\ENDFOR
\STATE $\hat{\mathbf{x}}_d \gets \mathcal{D}_d(\mathbf{q}_d)$, $\hat{\mathbf{x}}_f \gets \mathcal{D}_f(\mathbf{q}_d, \mathcal{D}_s(\mathbf{q}_s))$ \textbf{\{Decode\}}
\RETURN $\hat{\mathbf{x}}_d$ , $\hat{\mathbf{x}}_f$
% \RETURN The reconstructed sparse and dense tokens $\mathbf{q}_s$ and $\mathbf{q}_d$ .

\end{algorithmic}
\end{algorithm}
% Specifically, the projection of the sparse 2D pose serves as the class-conditional SOS token, which is fed into GCSAB to obtain the hidden representation. PH then predicts the sparse token in the first step of the forward process, and the corresponding $r$ dense tokens are generated in parallel based on this sparse token in the next step. LSAB encodes the sparse and dense tokens into the refined representations for the next generation step. 
% This process is repeated in a hierarchical pose auto-regression manner until all discrete tokens are generated. Finally, we reconstruct the hierarchical dense 2D poses  from the generated discrete tokens via the MSS decoder.

\begin{table*}[!t]
\setlength{\tabcolsep}{2.5pt}
\setlength{\abovecaptionskip}{0pt}
\centering
\caption{Comparison of the proposed method with state-of-the-art methods on Human3.6M using MPJPE. The table is split into 3 groups. The top group uses SH detected~\cite{NewellYD16} 2D poses, the middle group uses CPN detected~\cite{chen2018cascaded} 2D poses, and the bottom group uses group truth 2D poses as inputs. $\dagger$ denotes using temporal context, $\star$ denotes using visual cues and $\ast$ denotes using hierarchical  information. $f$ denotes the number of frames used in temporal-based methods. Best in \textbf{bold}, second best in \underline{underlined}.}
\label{tab:sota1}
%\resizebox{0.95\textwidth}{!}{
\small
\begin{tabular}{l|cccccccccccccccccc|c}
\toprule[1pt]
Method                      & Dir.  & Disc. & Eat   & Greet & Phone & Photo & Pose  & Purch. & Sit   & SitD. & Smoke & Wait  & WalkD. & Walk  & WalkT. & Avg.  \\ \hline
Learning~\cite{fang2018learning} & 50.1& 54.3& 57.0& 57.1& 66.6 &73.3& 53.4 &55.7 &72.8 &88.6 &60.3& 57.7 &62.7& 47.5 &50.6 &60.4\\
SemGCN~\cite{zhao2019semantic}$\star$   & 47.3& 60.7& \underline{51.4} &60.5& 61.1 &49.9 &\underline{47.3}& 68.1 &86.2 &55.0& 67.8& 61.0 &42.1& 60.6& 45.3& 57.6 \\
Monocular~\cite{xu2021monocular}$\star$  & \underline{47.1}& 52.8& 54.2& 54.9& 63.8 &72.5 &51.7 &54.3& 70.9 &85.0& 58.7 &54.9& 59.7& \textbf{43.8} &47.1& 58.1 \\
Graformer~\cite{zhao2022graformer} &49.3 &53.9& 54.1 &55.0 &63.0 &69.8& 51.1& \underline{53.3} &\underline{69.4} &90.0 &58.0 &55.2& 60.3 &47.4&50.6 &58.7  \\
Lifting by Image~\cite{zhou2024lifting}$\star$ & 48.3 & \underline{51.5}&  \textbf{46.1} & \underline{48.5}&  \textbf{53.7}&  \textbf{42.8}&  \underline{47.3}&  59.9&  71.0&  \textbf{51.6} & \underline{52.7} & \underline{46.1}&  \textbf{39.8} & 53.0&  \underline{43.9}&  \underline{51.0} \\ 
 \hline 
Ours$\ast$ &\textbf{46.9} &\textbf{50.4}& 56.8 &\textbf{47.1}  &\underline{56.4} &\underline{46.2}& \textbf{42.3}& \textbf{49.2} &\textbf{68.7} & \underline{54.9} &\textbf{51.9} &\textbf{43.1}& \underline{41.0} &\underline{45.2}& \textbf{40.1} & \textbf{49.3} \\\hline 
 \hline
VideoPose~\cite{pavllo20193d}$\dagger$($f=243$) & 45.1& 47.4 &42.0& 46.0& 49.1& 56.7 &44.5& 44.4 &57.2& 66.1 &47.5& 44.8 &49.2 &32.6 &34.0& 47.1  \\
GraphSH~\cite{xu2021graph} & 45.2& 49.9 &47.5& 50.9& 54.9 &66.1& 48.5& 46.3 &59.7 &71.5 &51.4& 48.6& 53.9& 39.9& 44.1 &51.9  \\
% MGCN~\cite{zou2021modulated}     &45.4& 49.2 &45.7& 49.4& 50.4& 58.2 &47.9& 46.0& 57.5& 63.0& 49.7 &46.6 &52.2 &38.9 &40.8 &49.4  \\
POT~\cite{li2023pose} & 47.9& 50.0 &47.1& 51.3& 51.2& 59.5 &48.7& 46.9 &56.0 &61.9& 51.1 &48.9 &54.3 &40.0& 42.9& 50.5 \\
DiffPose~\cite{gong2023DiffPose}  & 42.8& 49.1& 45.2& 48.7 &52.1& 63.5& 46.3 &45.2& 58.6 &66.3& 50.4 &47.6 &52.0 &37.6& 40.2& 49.7 \\
Di\textsuperscript{2}Pose~\cite{wang2024text}&41.9& 47.8& 45.0& 49.0& 51.5 &62.2 &45.7 &45.6& 57.6& 67.1 &50.1& 45.3& 51.4& 37.3& 40.9& 49.2\\
Lifting by Image~\cite{zhou2024lifting}$\star$   & 44.9& 46.4 &42.4 &44.9 &48.7& \textbf{40.1}& 44.3& 55.0& 58.9& \textbf{47.1}& 48.2& 42.6& \textbf{36.9}& 48.8& 40.1& 46.4 \\ 
PoseFormer~\cite{zheng20213d}$\dagger$($f=81$)&41.5 &44.8& 39.8 &42.5& 46.5& 51.6 &42.1 &42.0 &53.3 &60.7& 45.5& 43.3& 46.1& 31.8 &32.2 &44.3\\
MHFormer~\cite{li2022mhformer}$\dagger$($f=351$) & \textbf{39.2}& 43.1& 40.1 &\underline{40.9}& 44.9 &51.2& \underline{40.6}& \textbf{41.3} &53.5 &60.3& \underline{43.7} &41.1 &43.8 & \underline{29.8}& \underline{30.6}& 43.0  \\
MixSTE~\cite{zhang2022mixste}$\dagger$($f=81$)& \underline{39.8}& \underline{43.0}& \underline{38.6}& \textbf{40.1}& \underline{43.4}& 50.6& \underline{40.6}& \underline{41.4}& \underline{52.2}& 56.7& 43.8& \underline{40.8}& 43.9& \textbf{29.4}& \textbf{30.3}& \underline{42.4}\\
 \hline
Ours$\ast$  &42.8& \textbf{42.7}& \textbf{38.1}& 41.3& \textbf{42.7}& \underline{46.3}& \textbf{37.2}& 44.2& \textbf{51.0}&\underline{51.4}& \textbf{40.9}& \textbf{38.3}& \underline{40.0}& 39.9& 33.7&  \textbf{42.0} \\ \hline
 \hline
SemGCN~\cite{zhao2019semantic}$\star$   & 37.8& 49.4& 37.6& 40.9& 45.1 &41.4 &40.1 &48.3& 50.1 &42.2& 53.5 &44.3& 40.5 &47.3 &39.0 &43.8  \\
VideoPose~\cite{pavllo20193d}$\dagger$($f=243$) & -&  -&  -&  -&  - & - & -&  -&  - & - & - & - & - & -&  - &37.2 \\
Pose2Mesh~\cite{choi2020pose2mesh}$\ast$ & 38.1 &41.7 &38.3 &37.5 &39.2& 45.4& 37.5& 36.2 &45.7& 50.1& 39.8 &39.2& 40.2& 35.2& 37.6&40.1\\
HGN~\cite{li2021hierarchical}$\ast$ &  35.4 &40.2 &31.1 &38.2 &38.3& 41.1& 36.1& 32.7 &42.1& 48.4& 37.1 &36.9& 37.1& 30.5& 32.4&37.2                      \\
% GraphSH~\cite{xu2021graph} & 35.8& 38.1& 31.0& 35.3& 35.8& 43.2& 37.3& 31.7& 38.4& 45.5& 35.4& 36.7& 36.8& 27.9 &30.7 &35.8  \\
% Graformer~\cite{zhao2022graformer} & 32.0 &38.0 &30.4 &34.4 &34.7& 43.3& 35.2& 31.4 &38.0& 46.2& 34.2 &35.7& 36.1& 27.4& 30.6& 35.2  \\
% POT~\cite{li2023pose} & 32.9& 38.3 &28.3 &33.8 &34.9 &38.7&37.2& 30.7& 34.5& 39.7& 33.9& 34.7 &34.3& 26.1 &28.9& 33.8 \\
DiffPose~\cite{gong2023DiffPose}   & 28.8 &32.7& 27.8& 30.9& 32.8& 38.9& 32.2 &28.3& 33.3& 41.0& 31.0& 32.1& 31.5& 25.9 &27.5& 31.6 \\
Lifting by Image~\cite{zhou2024lifting}$\star$   & 29.5  & 30.1  & \underline{25.0}  & 29.0  & \underline{28.5} &\textbf{ 28.6} & \textbf{26.9}  & 30.5  & 31.1 & \textbf{27.7}  & 32.4  & 27.7  & \textbf{24.8}   & 30.0  & 25.9   & 28.6  \\ 
PoseFormer~\cite{zheng20213d}$\dagger$($f=81$)& 30.0& 33.6& 29.9 &31.0& 30.2 &33.3& 34.8 &31.4& 37.8 &38.6 &31.7 &31.5 &29.0& 23.3 &23.1& 31.3 
 \\ 
MHFormer~\cite{li2022mhformer}$\dagger$($f=351$) & \underline{27.7}&  32.1 & 29.1&  28.9&  30.0 & 33.9&  33.0 & 31.2&  37.0&  39.3&  30.0 & 31.0&  29.4&  22.2 & \underline{23.0}&  30.5 \\
MixSTE~\cite{zhang2022mixste}$\dagger$($f=81$)& \textbf{25.6}& \textbf{27.8}& \textbf{24.5} & \textbf{25.7}&  \textbf{24.9}& 29.9& 28.6&  \textbf{27.4}&  \textbf{29.9} & \underline{29.0} &\textbf{26.1}& \textbf{25.0}& \underline{25.2}& \textbf{18.7}& \textbf{19.9}& \textbf{25.9}\\
 \hline
Ours$\ast$  & 30.4  & \underline{29.7}  & 26.3 & \underline{27.2}  & 28.7  & \underline{29.1}& \underline{28.2} & \underline{29.2}  & \underline{30.9}  & 33.1  & \underline{29.6}  & \underline{26.2}  & 27.2   & \underline{21.9}  & 26.2   & \underline{28.3}  \\
\toprule[1pt]
\end{tabular}
%}
%\vspace{-5pt}
%\end{table*}
% zhwdone: 不把详细动作列出来，放在附录，标清楚使用的方法和GFlops、2DMPJPE等信息
% zhwdone: 补充更多的多帧结果
%\begin{table*}[t]
\setlength{\abovecaptionskip}{0pt}
\centering
\caption{Comparison of our method with state-of-the-art methods on 3DPW, 3DPW-Occ, and 3DPW-AdvOcc using MPJPE (Protocol 1) and PA-MPJPE (Protocol 2). 
We use CPN detected~\cite{chen2018cascaded} 2D poses as inputs. The 40 and 80 in 3DPWAdv-Occ represent the occluded size. 
% Best in \textbf{bold}, second best in \underline{underlined}.
}
\label{tab:sota2}
\small
%\resizebox{0.8\linewidth}{!}{%
\begin{tabular}{lcccccccccccc}
\toprule[1pt]
\multirow{2}{*}{Method} & \multicolumn{2}{c}{3DPW} & \multicolumn{2}{c}{3DPW-Occ} & \multicolumn{2}{c}{3DPW-AdvOcc@40} & \multicolumn{2}{c}{3DPW-AdvOcc@80} \\
                         & MPJPE  & PA-MPJPE            & MPJPE  & PA-MPJPE                & MPJPE  & PA-MPJPE                 & MPJPE  & PA-MPJPE                 \\ \hline
STRGCN~\cite{cai2019exploiting}   & 112.9   & 69.6                 & 115.8   & 72.3                     & 241.1   & 101.4                     & 355.9   & 116.3                     \\
VideoPose~\cite{pavllo20193d} & 101.8 & 63.0                 & 106.7   & 67.1                     & 221.6   & 99.4                      & 334.3   & 112.9                     \\
GnTCN~\cite{cheng2021graph} & —      & 64.2                 & —       & 85.7                     & 279.4   & 113.2                     & 371.4   & 119.8                     \\
PoseFormer~\cite{zheng20213d} & 118.2  & 73.1                 & 132.8   & 80.5                     & 247.9   & 106.2                     & 359.6   & 115.5                     \\
Learning~\cite{zhang2023learning} & 91.1  & 54.3                 & 94.6    & 56.7                     & 142.5   & 83.5                      & 251.8   & 103.9                     \\
PCT~\cite{geng2023human} & 83.1  & 53.9                 & 82.8    & 53.7                     & 127.2   & 72.2                      & 192.5   & 92.1                      \\
DiffPose~\cite{gong2023DiffPose} & 82.7  & 53.8                 & 82.1    & 53.5                     & \underline{121.4}   & 70.9                      & 189.3   & 92.4                      \\ 

Di\textsuperscript{2}Pose~\cite{wang2024text}  & \underline{79.3}    & \underline{50.1}                & \underline{79.6}    & \underline{50.7}                     & \textbf{108.4}   & \textbf{59.8}                      & \underline{153.6}   & \underline{78.7}                      \\
\hline
Ours   & \textbf{77.2}    & \textbf{48.8}   & \textbf{75.4} & \textbf{47.3} & 124.6 & \underline{64.3} & \textbf{143.2} &  \textbf{72.1}                          
\\ 
\toprule[1pt]
\end{tabular}%
%}
\vspace{-15pt}
\end{table*}

\section{Experiments}
\subsection{Experimental Settings}
\noindent\textbf{Datasets.}
Human3.6M~\cite{ionescu2013human3} is the most commonly used indoor dataset for 3D HPE. We follow prior works~\cite{pavllo20193d,zhao2019semantic, zhao2022graformer} to use five subjects (S1, S5, S6, S7, S8) for training and two subjects (S9, S11) for testing. 
3DPW~\cite{von2018recovering} is a popular in-the-wild dataset. We train our model on Human3.6M and test it on 3DPW to evaluate the generalization ability. Furthermore, a subset of 3DPW, \emph{i.e.,}  3DPW-Occ~\cite{zhang2020object}, is used  to validate the model's robustness against occlusions.

\noindent\textbf{Evaluation Metrics.}  We calculate the Mean Per Joint Position Error (MPJPE) under two protocols: Protocol 1 measures MPJPE by aligning the root (pelvis) keypoint between predicted and ground truth poses, while Protocol 2 (PA-MPJPE) first aligns through translation, rotation, and scaling.
We also follow ~\cite{zhang20233d} to use the 3DPW-AdvOcc protocol to evaluate performance under occlusion. Textured patches from the Describable Textures Dataset (DTD)~\cite{cimpoi2014describing} are applied to input images to assess degradation on visible joints and identify the worst prediction. Two patch sizes, Occ@40 and Occ@80, are used with a stride of 10. 
% We adopt CPN~\cite{chen2018cascaded} as the 2D detector to generate 2D poses.

\noindent\textbf{Implementation Details.}
The encoder and decoder of MSST comprise four and one MLP-Mixer~\cite{geng2023human} blocks with the embedding dimensions of 128 and 96, respectively.
% We adopt 16 tokens for sparse and 48 for dense tokenization, where 
The joint quantities of the sparse, dense, and fine 2D pose are set as $J_s=16$, $J_d=48$, and $J_f=96$.
Both codebook sizes are 2048 with a dimension of 128. In HiARM, the LSAB, GCSAB, and PH consist of 1, 12, and 4 transformer blocks with the same dimension of 128. The lifting model is a 12-block vanilla spatial transformer with 6 heads and a dimension of 96. Detailed settings are in the Appendix~\textcolor{red}{C}.
% 详细超参设置见附录

% \noindent\textbf{Results on MPI-INF-3DHP.}
\subsection{Comparison with State-of-the-art Methods} 

\noindent\textbf{Results on Human3.6M.} We evaluate our method on Human3.6M against three types of SOTA methods that incorporate additional information. 
% Results using SH detected, CPN detected, and ground truth 2D poses are shown in Tab.~\ref{tab:sota1}. 
Table~\ref{tab:sota1} shows that our method achieves the best performance in the single-frame settings including the hierarchy-based (w/ $\ast$) and visual-based (w/ $\star$) methods. 
% Additionally, our results are comparable to those of temporal-based methods (w/ $\dagger$), demonstrating that hierarchical information can be as effective as temporal information.
Moreover, our approach outperforms the temporal-based (w/ $\dagger$) methods when using detected 2D poses, which is closely to the real-world scenarios. Table\ref{tab:lifting} further demonstrates that hierarchical information can complement temporal-based methods.
% demonstrating the effectiveness of hierarchical information.

\noindent\textbf{Results on 3DPW.}
We evaluate our model pretrained on Human3.6M to the 3DPW dataset. As shown in Table~\ref{tab:sota2}, our method achieves SOTA performance with an average improvement of 2.1mm in MPJPE and 1.3mm in PA-MPJPE over Di\textsuperscript{2}Pose~\cite{wang2024text}, demonstrating strong generation ability. Our method also maintains superiority on 3DPW-Occ. Under the extreme occlusion scenarios of 3DPW-AdvOcc, our method shows a smaller performance drop (77\% in MPJPE and 44\% in PA-MPJPE) compared to the average drop of 120\% and 67\%, highlighting the superiority of hierarchical information in enhancing robustness under occlusions.

\subsection{Ablation Studies and Analysis}
We perform ablation studies on Human3.6M and 3DPW  using CPN detected 2D poses to understand how our method enhances 3D HPE. More results are in the Appendix~\textcolor{red}{E}.

\noindent\textbf{Ablation Study on each Component of MSST.}
We verify each component of MSST by either removal or standard component replacement in Table~\ref{tab:PH-VAE}.
The accuracy on both datasets reduces sharply when replacing hierarchical codebooks with one shared codebook. It highlights the importance of hierarchical design in token learning. Then, we sequentially remove local and Global Alignment, which results in respective performance loss of 1.6mm and 0.4mm on Human3.6M, indicating that these alignments enhance discrete space learning and strengthen token connectivity.

\begin{table}[!t]
\setlength{\abovecaptionskip}{0pt}
\centering
\caption{Ablation study on each component of MSST. H36M denotes the Huamn3.6 dataset.}
\label{tab:PH-VAE}
\small
\begin{tabular}{@{}llcc@{}}
\toprule[1pt]
Method& &H36M&  3DPW\\
\hline
\multicolumn{2}{l}{Our method}&\multicolumn{1}{l}{\textbf{42.0}}   &  \multicolumn{1}{l}{\textbf{77.2}} \\
\multicolumn{2}{l}{\quad w/o hierarchical codebooks}& 45.3\textcolor{gray}{\footnotesize{$\Delta$3.3}} &  81.5\textcolor{gray}{\footnotesize{$\Delta$4.3}}  \\
\multicolumn{2}{l}{\quad w/o Local Alignment }& 43.6\textcolor{gray}{\footnotesize{$\Delta$1.6}}  &  80.2\textcolor{gray}{\footnotesize{$\Delta$3.0}}  \\
\multicolumn{2}{l}{\quad w/o Global Alignment}& 42.4\textcolor{gray}{\footnotesize{$\Delta$0.4}}  & 78.9\textcolor{gray}{\footnotesize{$\Delta$1.7}}\\
\toprule[1pt]
\end{tabular}
\vspace{-12pt}
\end{table}
\begin{figure}[t]
\setlength{\abovecaptionskip}{0pt}
    \centering
    \includegraphics[width=0.80\linewidth]{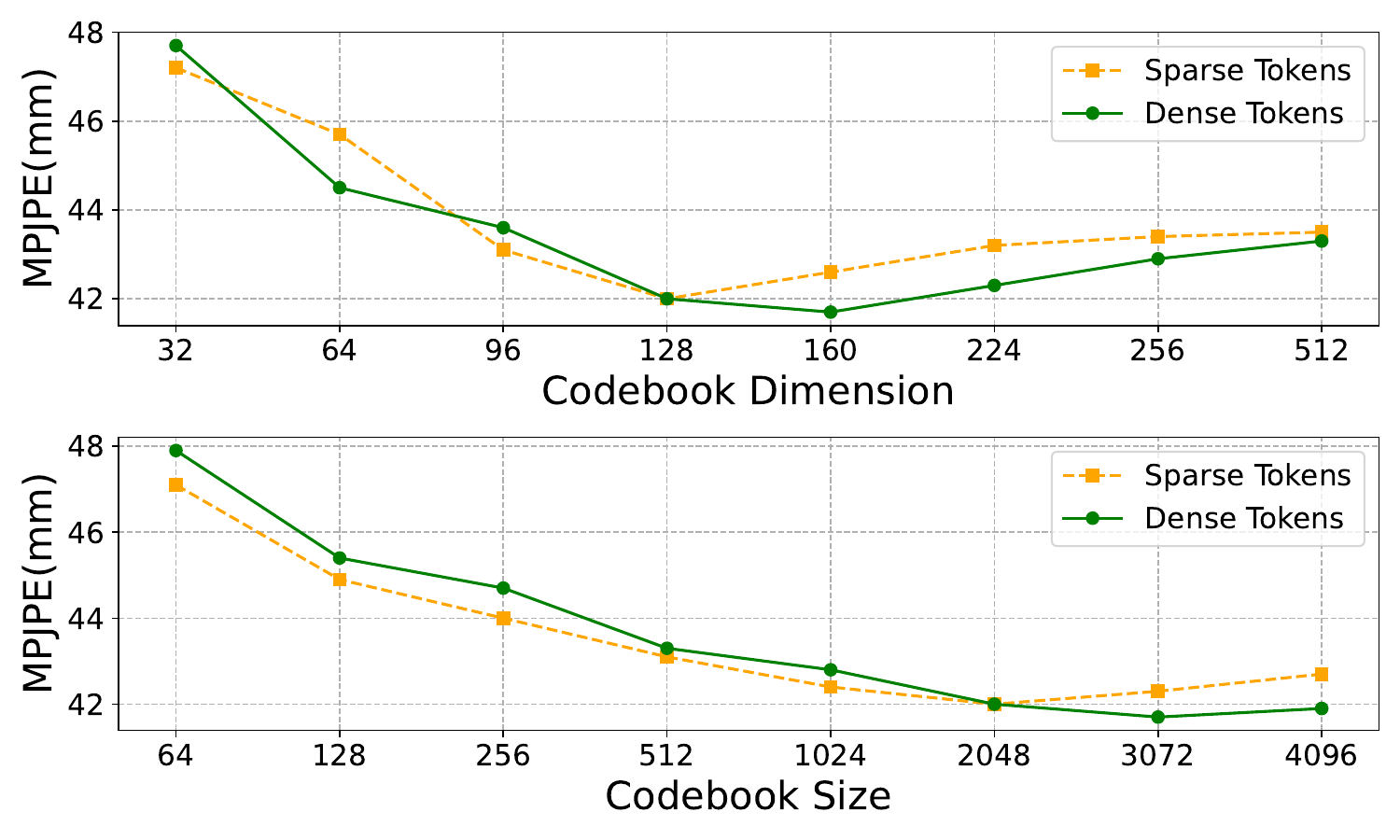}
    \caption{Impact of the codebook dimension (\textbf{top}) and size (\textbf{bottom}) for sparse and dense tokens on Humman3.6M.}
    \label{fig:codebook}
    \vspace{-10pt}
\end{figure}

\begin{figure*}[!t]
\setlength{\abovecaptionskip}{0pt}
\centering
\begin{subfigure}[b]{0.38\textwidth}
  \includegraphics[width=\textwidth]{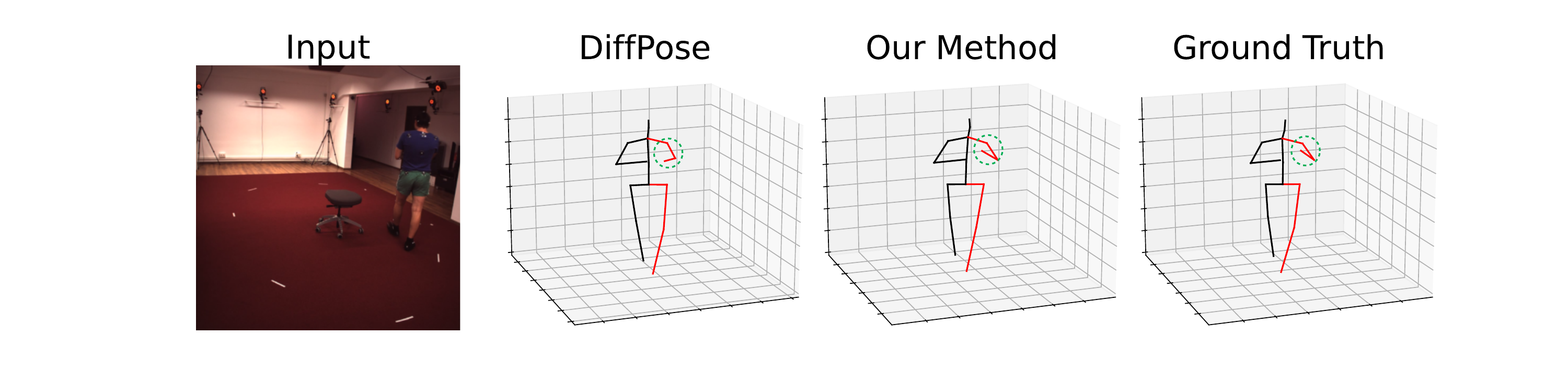}
\end{subfigure}
\begin{subfigure}[b]{0.38\textwidth}
  \includegraphics[width=\textwidth]{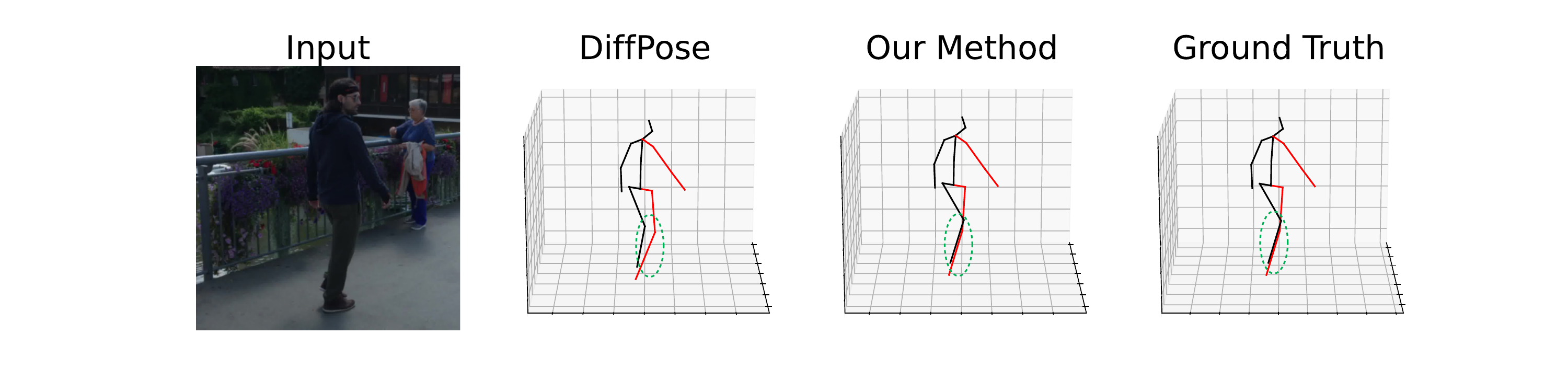}
\end{subfigure}\\
\begin{subfigure}[b]{0.38\textwidth}
  \includegraphics[width=\textwidth]{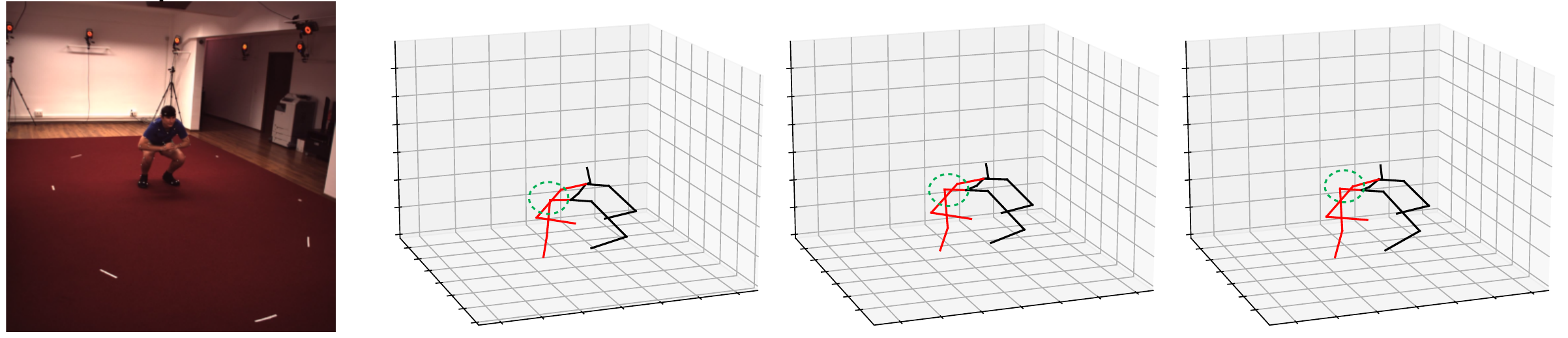}
\end{subfigure}
\begin{subfigure}[b]{0.38\textwidth}
  \includegraphics[width=\textwidth]{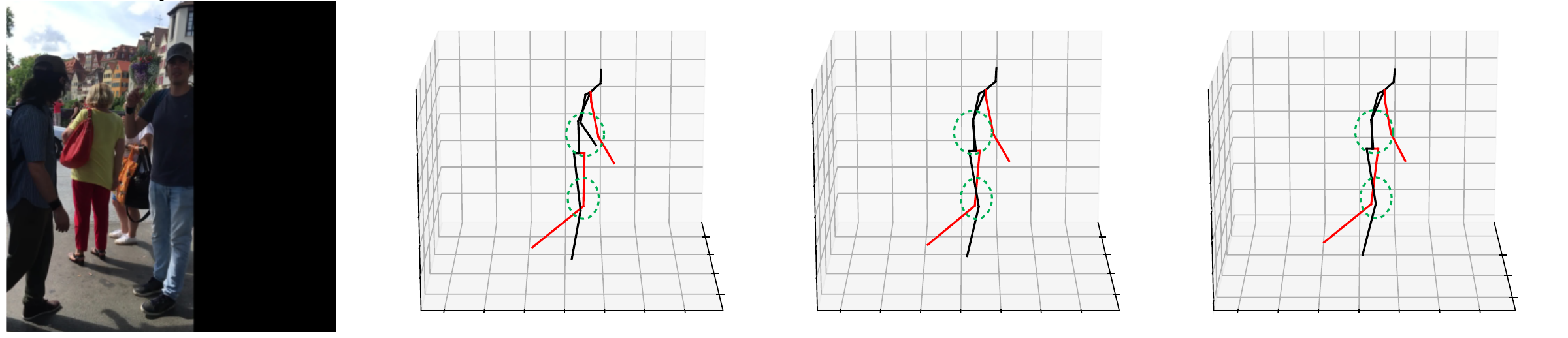}
\end{subfigure}
\\
\begin{subfigure}[b]{0.38\textwidth}
  \includegraphics[width=\textwidth]{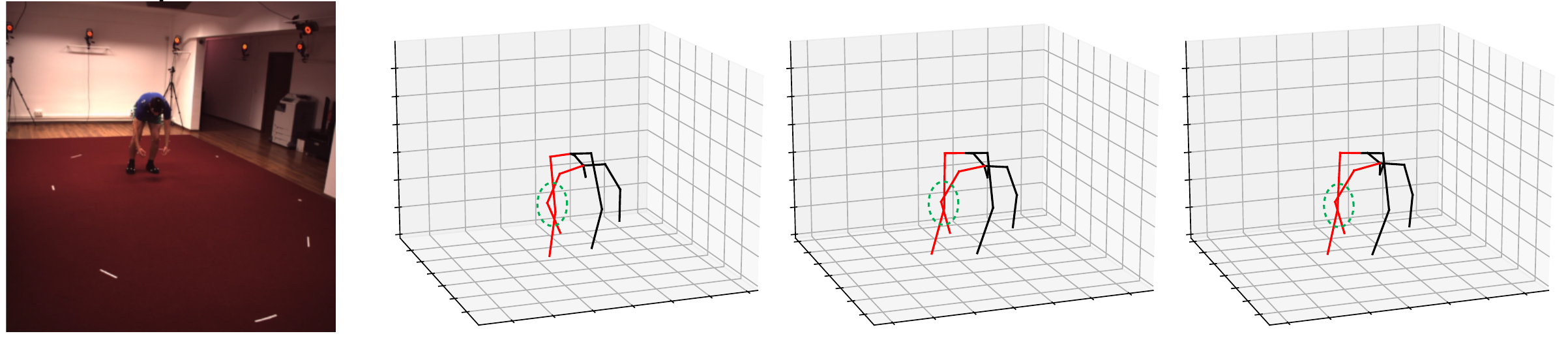}
\end{subfigure}
\begin{subfigure}[b]{0.38\textwidth}
  \includegraphics[width=\textwidth]{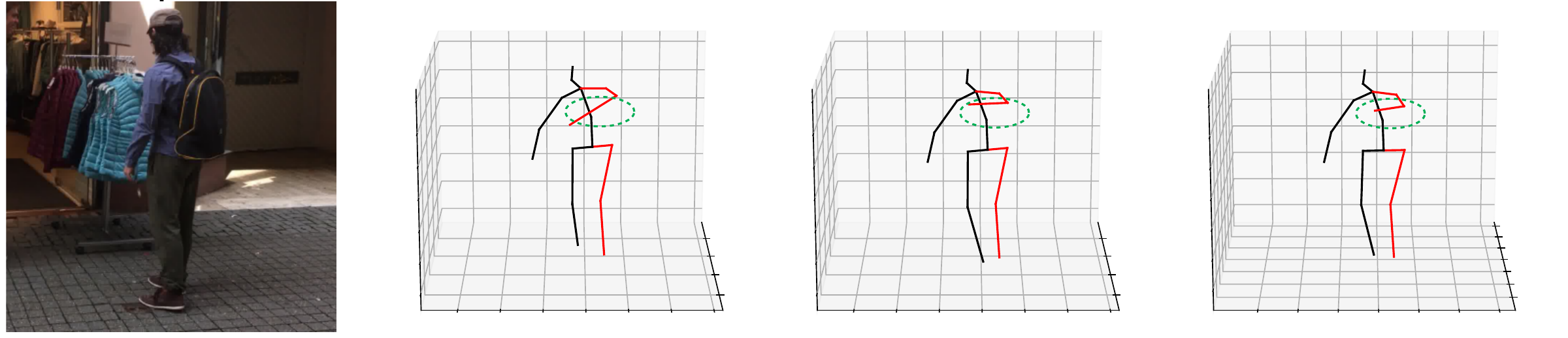}
\end{subfigure}
\caption{Qualitative results compared with DiffPose~\cite{gong2023DiffPose} on Human3.6M (\textbf{left}) and 3DPW (\textbf{right}).}
\label{fig:Qualitative}
\vspace{-10pt}
\end{figure*}

\noindent\textbf{Ablation Study on the Codebook Dimension and Size.} 
% We evaluate the impact of the token number and codebook size on Human3.6M. The results in Fig.~\ref{fig:codebook} indicate that increasing the sparse token number from $8$ to $24$ and the dense token number from $8$ to $48$ significantly improves performance. However, further increasing the number of tokens introduces redundancy and repetition, which does not lead to additional performance enhancements.
% Then, we adopt a range of codebook sizes for both sparse and dense tokens. We find that enlarging the codebook size can boost performance. Performance peaks when the codebook sizes for sparse and dense tokens reach $2048$ and $3072$, respectively. However, further increasing the codebook size raises the difficulty of learning the representation without substantially improving performance.
We examine the effects of the codebook dimension and size on Human3.6M. Fig.~\ref{fig:codebook} shows that increasing the sparse and dense codebook dimension from 32 to 128 leads to significant performance gains,
but further increases introduce redundancy without additional improvement.
Testing different codebook sizes reveals peak performance at 2048 for the sparse and 3072 for the dense. Larger sizes increase the learning complexity without notable performance gains.

% 画图类似PCT，每张图有2条线

\noindent\textbf{Ablation Study on HiARM.} We then conduct an ablation study on the framework type and auto-regression strategy of HiARM in Table~\ref{tab:HiARM}. Following PCT~\cite{geng2023human}, we reformulate pose estimation in Stage 2 as a classification task by predicting the categories of hierarchical tokens. When replacing it with an auto-regressive model (ARM) and employing a next-token prediction order, MPJPE reduces by 2.7mm, demonstrating that ARM better captures the token latent distribution. We further adopt different auto-regression strategies.   
``Center-to-periphery'', ``Sparse-to-dense'' and their combination yield great performance improvements, reaching a final best MPJPE of 42.0mm. ``Sparse-to-dense'' speeds up the ARM inference process by 49\%.

% \noindent\textbf{Model complexity and computation cost.}
% We compare the model complexity and computational cost of our method with single-frame methods and temporal-based methods. Tab.~\ref{tab:computation} shows that our method achieves better performance with a similar parameter and computational cost than single-frame methods like GraphMLP~\cite{li2025graphmlp} and POT~\cite{li2023pose}. Compared to temporal-based methods, our method is more lightweight and yet offers comparable or even superior performance. For instance, MHFormer~\cite{li2022mhformer} has lower accuracy than our method but requires 18.9M and 24.2 GFLOPs for lifting, while our method only consumes 2.4M and 1.3 GFLOPs (8 and 11 times reductions).

\begin{table}[!t]
\setlength{\abovecaptionskip}{0pt}
\setlength{\tabcolsep}{2pt}
\centering
\caption{Ablation study on HiARM. ``Both” denotes using both ``Center-to-periphery'' and ``Sparse-to-dense''  strategies. Frame per second (FPS) is based on one NVIDIA Tesla V100 GPU.}
\label{tab:HiARM}
\resizebox{\linewidth}{!}{%
\begin{tabular}{@{}llcccc@{}}
\toprule[1pt]
Framework Type& Auto-Regression Strategy& H36M &  3DPW& Params(M) & FPS\\
\hline
Classification& None &\multicolumn{1}{l}{46.3} & \multicolumn{1}{l}{81.9}&2.2  & 432\\
ARM& Next-token & 43.6\textcolor{darkgreen}{\footnotesize{$\downarrow$2.7}}  &80.2\textcolor{darkgreen}{\footnotesize{$\downarrow$1.7}} &2.4 & 265 \\
ARM& Center-to-periphery & 42.5\textcolor{darkgreen}{\footnotesize{$\downarrow$3.8}} & 78.4\textcolor{darkgreen}{\footnotesize{$\downarrow$3.5}} &2.4 & 265\\
ARM& Sparse-to-dense &43.1\textcolor{darkgreen}{\footnotesize{$\downarrow$3.2}}  & 79.3\textcolor{darkgreen}{\footnotesize{$\downarrow$2.6}}&2.4 & 396\\
ARM& Both & \textbf{42.0}\textcolor{darkgreen} {\footnotesize{$\downarrow$4.3}}&\textbf{77.2}\textcolor{darkgreen}{\footnotesize{$\downarrow$4.7}} &2.4 & 396\\
\bottomrule[1pt]
\end{tabular}
}
\vspace{-6pt}
%\end{table}
%\begin{table}[!t]
%\begin{center}
\setlength{\tabcolsep}{6pt}
\caption{Ablation study on the level of hierarchical input 2D poses.}
\label{tab:2d level}
\small
\begin{tabular}{@{}lll|cc@{}}
\toprule[1pt]
48&96&192& H36M &  Params(M)\\
\hline
& &  &  \multicolumn{1}{l}{52.1}&  0.9 \\
\checkmark& &  & \multicolumn{1}{l}{46.7\textcolor{darkgreen}{\footnotesize{$\downarrow$5.4}}} &  1.4\\
&\checkmark&&  \multicolumn{1}{l}{43.2\textcolor{darkgreen}{\footnotesize{$\downarrow$8.9}}} & 1.8\\
&&\checkmark&\multicolumn{1}{l}{42.7\textcolor{darkgreen}{\footnotesize{$\downarrow$9.4}}} & 2.5\\
\checkmark&\checkmark&&\multicolumn{1}{l}{42.0\textcolor{darkgreen}{\footnotesize{$\downarrow$10.1}}}& 2.4 \\
\checkmark&\checkmark&\checkmark&  \multicolumn{1}{l}{\textbf{41.4}\textcolor{darkgreen}{\footnotesize{$\downarrow$10.7}}} &4.4\\
\bottomrule[1pt]
\end{tabular}
%\end{center}
%\vspace{-20pt}
%\end{table}
%\begin{table}[!t]
%\begin{center}
\setlength{\tabcolsep}{2pt}
\caption{Ablation study on lifting models. We compare model complexity, inference speed, and MPJPE on Human3.6M. }
\label{tab:lifting}
\resizebox{0.99\linewidth}{!}{%
\begin{tabular}{@{}llccc@{}}
\toprule[1pt]
\multicolumn{2}{@{}l}{Lifting Model}&Params(M) &FPS &  MPJPE \\
\hline
\multicolumn{2}{@{}l}{vanilla spatial transformer }&0.9 &458 & \multicolumn{1}{l@{}}{52.1}\\
\multicolumn{2}{@{}l}{vanilla spatial transformer w/ ours}&2.0 &396 &42.0\textcolor{darkgreen}{\footnotesize{$\downarrow$10.1}}  \\
\hline
\multicolumn{2}{@{}l}{PoseFormer~\cite{zheng20213d} ($f=81$)}& 9.7&381 &\multicolumn{1}{l}{44.3} \\
\multicolumn{2}{@{}l}{PoseFormer~\cite{zheng20213d}  ($f=81$) w/ ours}& 10.7& 185& \multicolumn{1}{l@{}}{40.8\textcolor{darkgreen}{\footnotesize{$\downarrow$3.5}}}\\
\hline
\multicolumn{2}{@{}l}{MixSTE~\cite{zhang2022mixste} ($f=81$)}&33.7 &1134 &\multicolumn{1}{l}{42.4} \\
\multicolumn{2}{@{}l}{MixSTE~\cite{zhang2022mixste} ($f=81$) w/ ours}& 34.8&681 &\multicolumn{1}{l@{}}{39.3\textcolor{darkgreen}{\footnotesize{$\downarrow$3.1}}} \\
\bottomrule[1pt]
\end{tabular}
}
%\end{center}
\vspace{-10pt}
\end{table}

\begin{figure}[!t]
\setlength{\abovecaptionskip}{0pt}
\centering
\begin{subfigure}[b]{0.2\textwidth}
  \includegraphics[width=\textwidth]{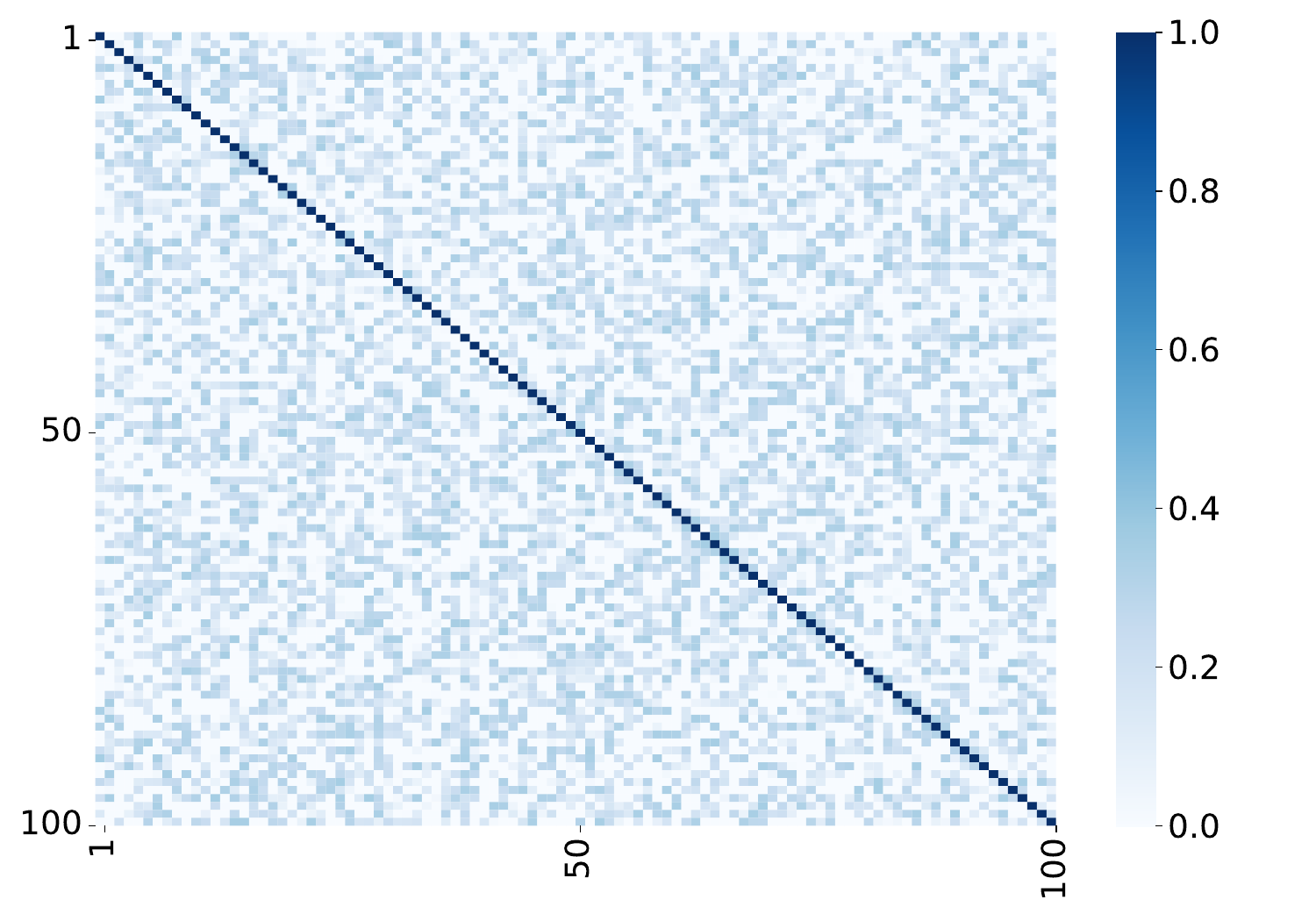}
\caption{Sparse tokens}
\end{subfigure}
\begin{subfigure}[b]{0.2\textwidth}
  \includegraphics[width=\textwidth]{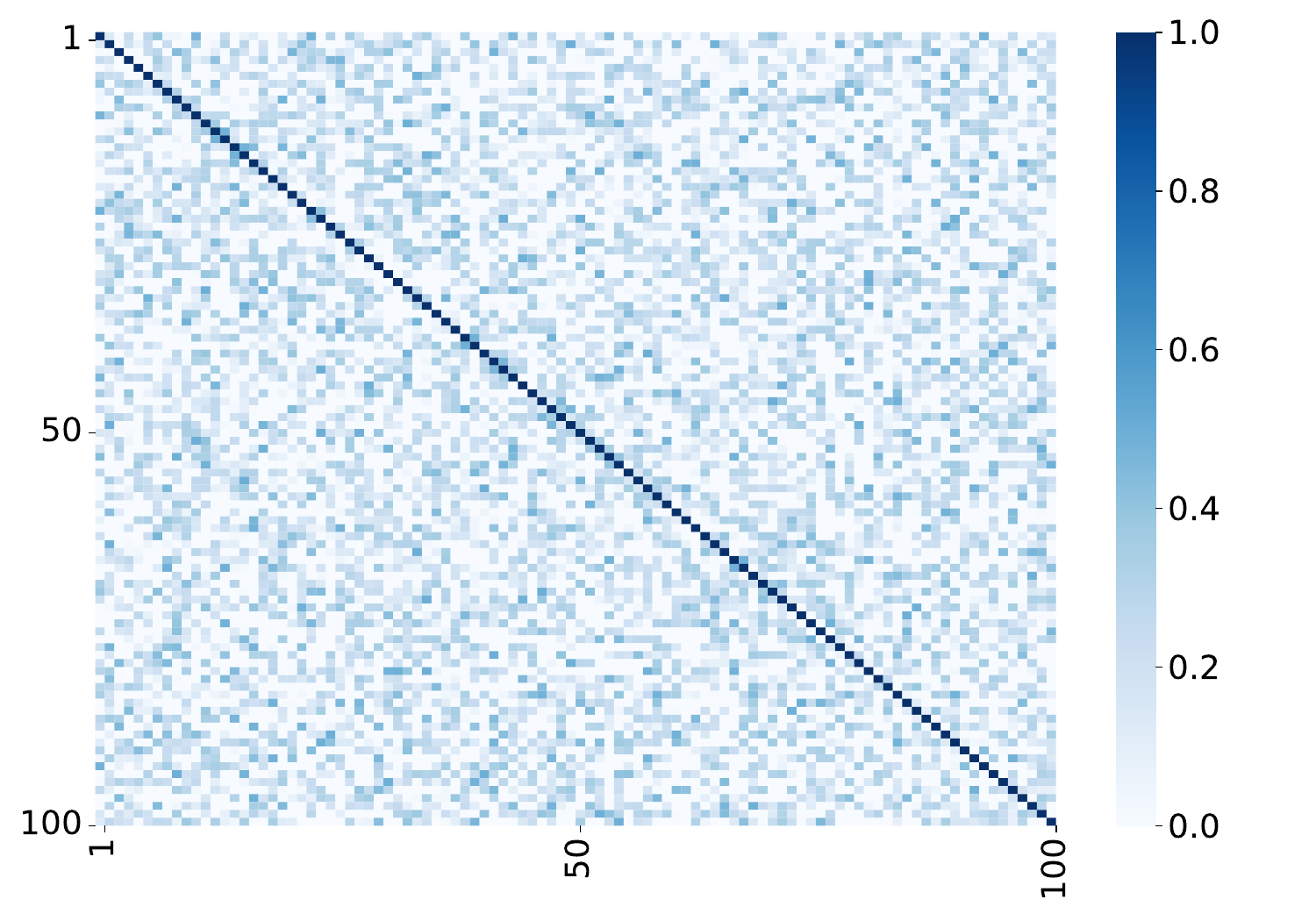}
\caption{Dense tokens}
\end{subfigure}\\
\begin{subfigure}[b]{0.2\textwidth}
  \includegraphics[width=\textwidth]{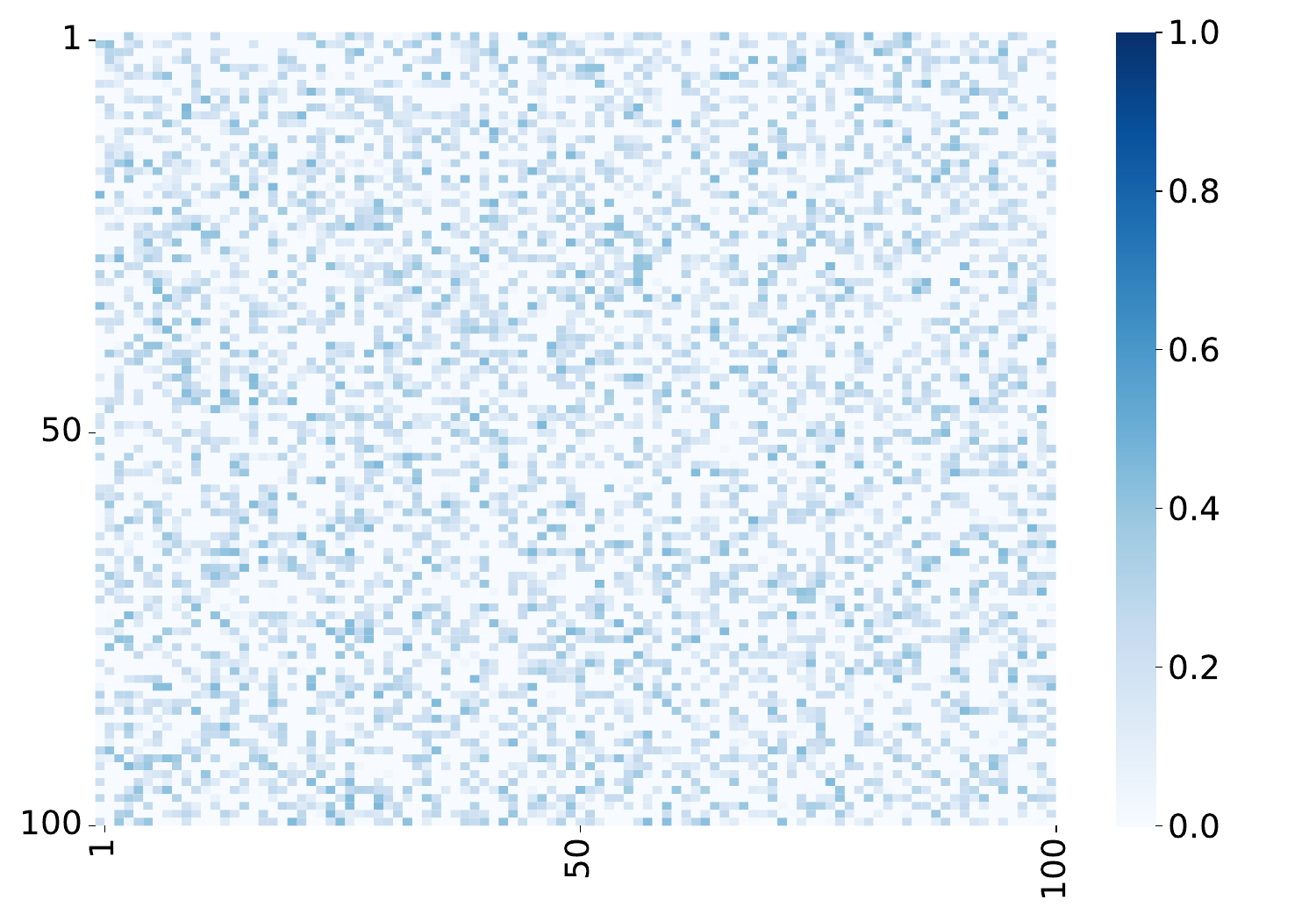}
\caption{Sparse and dense tokens w/o local alignment}
\end{subfigure}
\begin{subfigure}[b]{0.2\textwidth}
  \includegraphics[width=\textwidth]{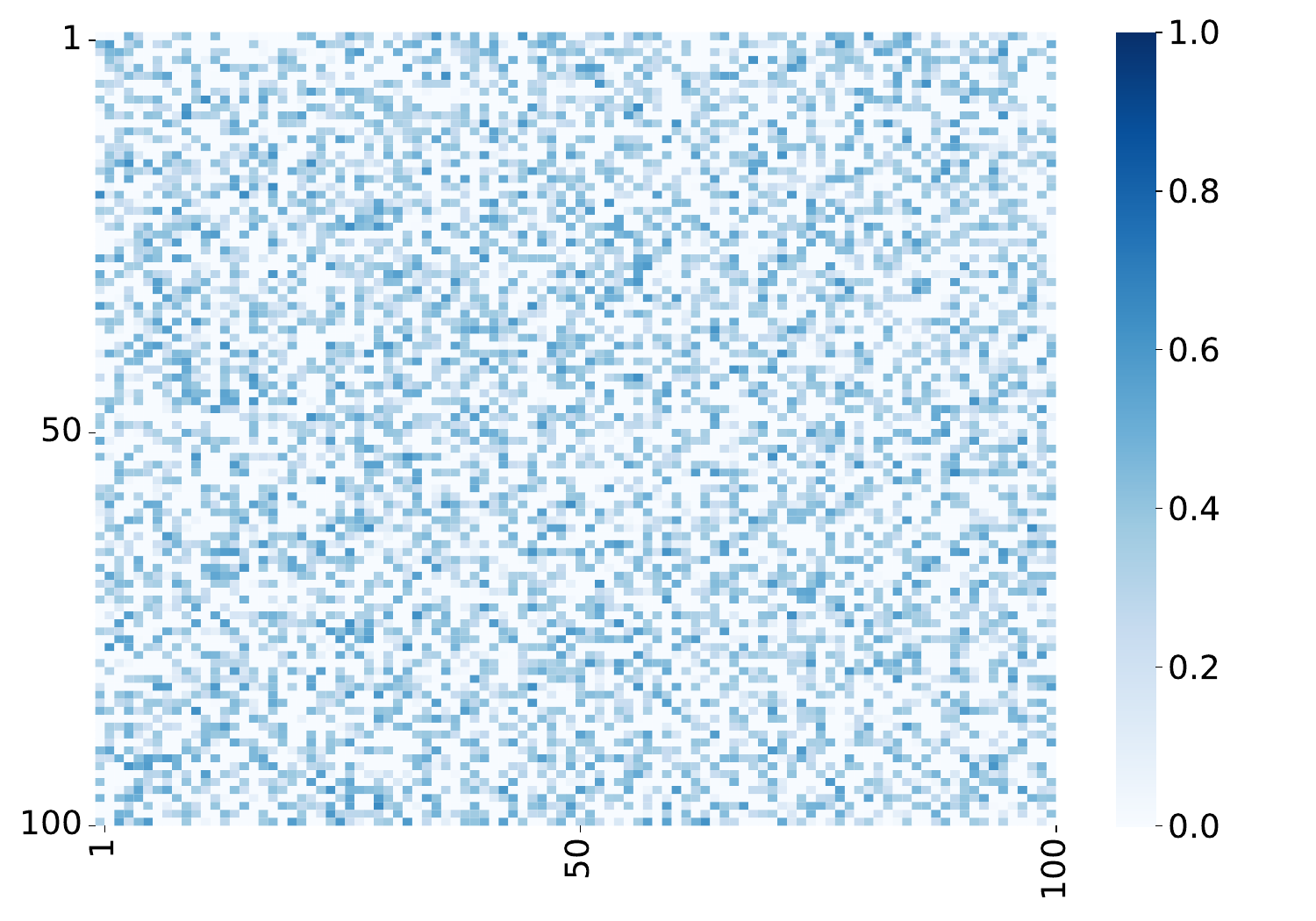}
\caption{Sparse  and dense tokens w/ local alignment}
\end{subfigure}
\caption{Hierarchical codebooks similarity analysis. The cosine similarity is calculated based on the first 100 tokens from both the sparse and dense codebooks. (c) (d) The x-axis represents sparse tokens, and the y-axis represents dense tokens.}
\label{fig:pearason}
\vspace{-5pt}
\end{figure}

\noindent\textbf{Ablation Study on the Effect of Different Levels of 2D Poses.} 
% To assess the impact of varying levels of 2D poses, we introduce three levels (48, 96, and 192) of 2D poses as additional inputs for the original sparse 2D pose (17 joints).
We validate the effect of different levels of 2D poses by incorporating the original sparse 2D pose (17 joints) with additional 2D poses at three levels (48, 96, and 192 joints).
Table~\ref{tab:2d level} reveals that denser 2D pose inputs  improve 3D HPE accuracy. Incorporating a 48-joint pose results in a 5.4mm gain (52.1mm to 46.7mm), and a 96-joint pose further improves 4.7mm (46.7mm to 42.0mm). However, adding a 192-joint pose only  enhances MPJPE by 0.6mm but significantly increases the complexity. We analyze that 
the distinct positional information from the 48-joint and 96-joint poses together boosts performance, but the already rich context in these two levels lets the 192-joint pose bring minimal gains.

\noindent\textbf{Extension to Temporal-based Lifting Models.} 
% We compare the performance of different lifting models, ranging from single-frame methods like the vanilla spatial transformer to multi-frame approaches such as PoseFormer~\cite{zheng20213d} and MixSTE~\cite{zhang2022mixste}.
We further extend the lifting model to temporal-based methods including PoseFormer~\cite{zheng20213d} and MixSTE~\cite{zhang2022mixste}. Table~\ref{tab:lifting} shows that the incorporation of hierarchical information leads to performance gains across various methods, achieving the SOTA result when combined with MixSTE with  a tolerable computational complexity. 
More discussion about the extension to temporal-based methods is in the Appendix~\textcolor{red}{G}.

% \begin{table}[!t]
% \begin{center}
% \caption{Comparison of model complexity and computation cost across different methods.}
% \label{tab:computation}
% \resizebox{0.9\linewidth}{!}{%
% \begin{tabular}{llccc}
% \toprule[1pt]
% \multicolumn{2}{l}{Method}&Params(M) &FLOPs(G)&  MPJPE\\
% \hline
% \multicolumn{2}{l}{GraphMLP~\cite{li2025graphmlp}}& 9.5& 0.3&34.2 \\
% \multicolumn{2}{l}{POT~\cite{li2023pose}}& 0.98 &1.6 &33.8 \\
% % \multicolumn{2}{l}{DiffPose~\cite{gong2023DiffPose} }& & &31.6 \\
% \multicolumn{2}{l}{PoseFormer~\cite{zheng20213d} ($f=81$) }&9.7 & 1.6 &31.3\\
% \multicolumn{2}{l}{MHFormer~\cite{li2022mhformer} ($f=351$) }&18.9 & 24.2&30.5\\
% \multicolumn{2}{l}{MixSTE~\cite{zhang2022mixste} ($f=243$) }& 33.8 & 277.4&21.6\\
% \hline
% \multicolumn{2}{l}{Ours } &2.4 & 2.2 & 28.3 \\
% \hline
% \toprule[1pt]
% \end{tabular}
% }
% \end{center}
% \vspace{-10pt}
% \end{table}

% 分类还是自回归，解码顺序

% zhwdone：比较不同方法的2D Mean Error噪声

We also conduct a series of qualitative analyses. 
% to illustrate the performance improvements achieved by our method. 
Additional visualizations are available in the Appendix~\textcolor{red}{F}.

\noindent\textbf{Qualitative Results.}  Fig.~\ref{fig:Qualitative} shows qualitative results on Human3.6M and 3DPW, comparing our method with DiffPose \cite{gong2023DiffPose}. Our method provides accurate pose predictions, especially in heavily occluded scenarios, demonstrating that the hierarchical information helps address the depth ambiguity due to occlusion.
% zhwdone: 对于遮挡，多层次节点的优势
% zhwdone: 3dpw的分析(附录详细分析)

\noindent\textbf{Analysis on the Similarity  of Hierarchical Codebooks}. We compute four cosine similarity matrices to evaluate the relationships among sparse tokens, dense tokens, and their interplay. Fig.~\ref{fig:pearason}(a) and (b) show that the connections within sparse and dense codebooks are generally low, indicating that each codebook contains relatively independent and non-redundant information. The slightly higher similarity among dense tokens compared to sparse tokens indicates that dense tokens capture more fine-grained, localized information, leading to stronger correlations.
Fig.~\ref{fig:pearason} (c) and (d) demonstrate that Local Alignment enhances the connection between sparse and dense tokens.

\begin{figure}[t]
\setlength{\abovecaptionskip}{0pt}
    \centering
    \includegraphics[width=0.9\linewidth]{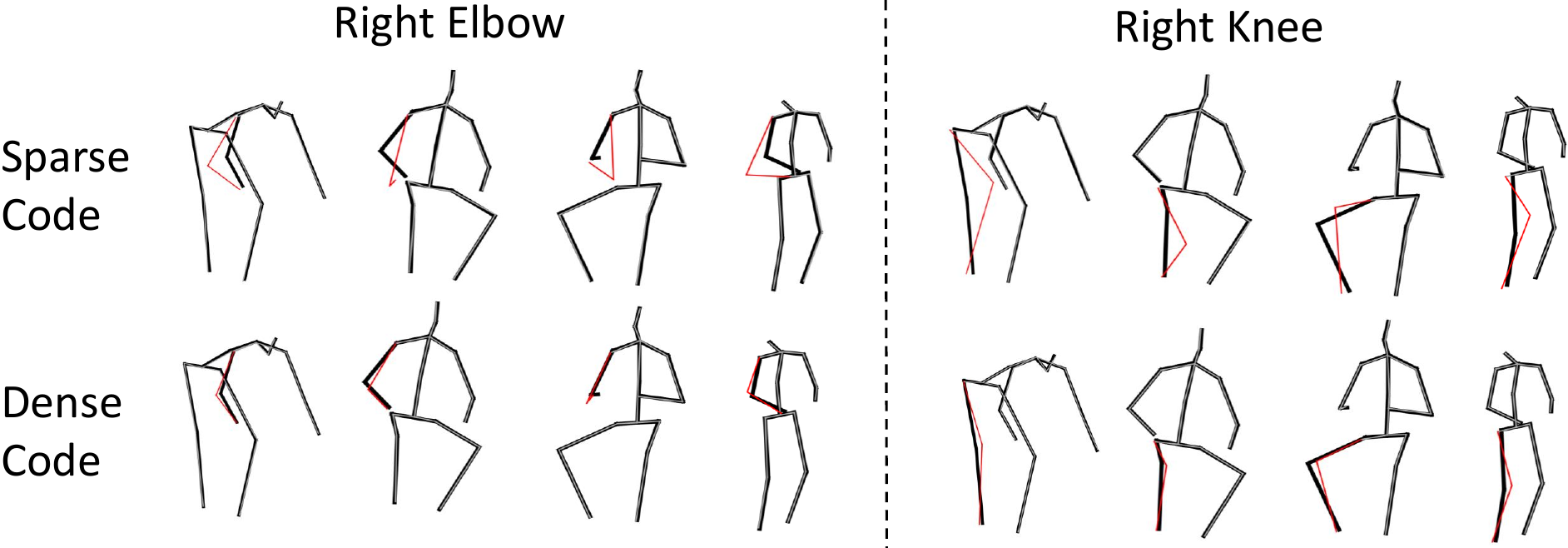}
    \caption{Visualization of the impact of hierarchical tokens on the sub-structure. We change the single sparse and dense token of two parts into another value, respectively, and the corresponding sub-structures consistently change (Highlighted in red). Black poses denote the original, while gray poses denote the change.}
    \label{fig:code_vis}
    \vspace{-5pt}
\end{figure}

\noindent\textbf{Impact of Hierarchical Tokens on the Sub-structure.}
% 参考PCT 
Fig.~\ref{fig:code_vis} visualizes the influence of hierarchical tokens on sub-structures in Human3.6M using 16 sparse and 48 dense tokens. The pose sub-structure, represented by a pair of joints, is primarily controlled by two corresponding tokens. Modifying the sparse token leads to significant changes in the sub-structure, indicating that it captures coarse-grained information, while changes to the dense token result in minor adjustments, suggesting it focuses on fine-grained details.

% 附录
% compuation cost
% Results on the H36M under occlusion
% \noindent\textbf{Ablation study on different lifting models.} 包含多帧结果
% lifting模块结果，附录
% 只用96或者48的joint作为输入的ablation study
% 使用sparse或者dense code重建的可视化结果, 只是用sparse，在精细处的缺失；只使用dense基本错误
% 更多可视化结果
% MPJPE distribution of different actions不同动作的详细mpjpe
% 2dmpjpe详细结果，不同action
% MPII数据集结果

\section{Conclusion}
% This paper presents a novel generative densification approach, named Hierarchical Pose AutoRegressive Transformer (HiPART), to obtain reliable hierarchical dense 2D poses based on the original sparse 2D pose. Specially, we progressively quantize a highly dense 2D pose into hierarchical tokens using a multi-scale skeleta tokenization (MSST) module and propose a Hierarchical AutoRegressive Modeling (HiARM) scheme for hierarchical dense 2D pose generation. As the first hierarchical dense 2D pose generation scheme for addressing the occlusion, HiPART achieves SOTA performance on the single-frame setting and is superior to numerous multi-frame methods.

This paper introduces the Hierarchical Pose AutoRegressive Transformer (HiPART), a novel generative densification approach that derives reliable hierarchical dense 2D poses from the sparse 2D pose. We employ a Multi-Scale Skeletal Tokenization module to progressively quantize a highly dense 2D pose into hierarchical tokens and establish a Hierarchical AutoRegressive Modeling scheme for the pose generation. As the first approach for hierarchical dense 2D pose generation tackling occlusions, HiPART achieves SOTA performance on the single-frame setting using only a vanilla spatial transformer for 2D-to-3D lifting and even outperforms numerous multi-frame methods.

\section*{Acknowledgement}
This work was supported in part by the National Natural Science Foundation of China under Grant 62125109, Grant 62431017, Grant U24A20251, Grant 62320106003, Grant 62371288, Grant 62401357, Grant 62401366, Grant 62301299, Grant 62120106007, and in part by the Program of Shanghai Science and Technology Innovation Project under Grant 24BC3200800.

{
    \small
    \bibliographystyle{ieeenat_fullname}
    \bibliography{main}
}

\clearpage
\setcounter{page}{1}
\maketitlesupplementary

\renewcommand\thesection{\Alph{section}}
\setcounter{section}{0}
% \section{Rationale}
% \label{sec:rationale}
% % 
% Having the supplementary compiled together with the main paper means that:
% % 
% \begin{itemize}
% \item The supplementary can back-reference sections of the main paper, for example, we can refer to \cref{sec:intro};
% \item The main paper can forward reference sub-sections within the supplementary explicitly (e.g. referring to a particular experiment); 
% \item When submitted to arXiv, the supplementary will already included at the end of the paper.
% \end{itemize}
% % 
% To split the supplementary pages from the main paper, you can use \href{https://support.apple.com/en-ca/guide/preview/prvw11793/mac#:~:text=Delete%20a%20page%20from%20a,or%20choose%20Edit%20%3E%20Delete).}{Preview (on macOS)}, \href{https://www.adobe.com/acrobat/how-to/delete-pages-from-pdf.html#:~:text=Choose%20%E2%80%9CTools%E2%80%9D%20%3E%20%E2%80%9COrganize,or%20pages%20from%20the%20file.}{Adobe Acrobat} (on all OSs), as well as \href{https://superuser.com/questions/517986/is-it-possible-to-delete-some-pages-of-a-pdf-document}{command line tools}.

% Exp
% computation cost
% Results on the H36M under occlusion
% \noindent\textbf{Ablation study on different lifting models.} 包含多帧结果
% 只用96或者48的joint作为输入的ablation study
% 2dmpjpe详细结果，不同action

% Vis
% 使用sparse或者dense code重建的可视化结果, 只是用sparse，在精细处的缺失；只使用dense基本错误
% 更多可视化结果
% 更多多层次节点的结果

\section{Pseudo-code for Our HiPART Algorithm}
We define the pseudo-code for Stage 1 and Stage 2 of our Hierarchical Pose AutoRegressive Transformer (HiPART) algorithm during training in Alg.~\ref{alg:stage1} and ~\ref{alg:stage2}.

\section{Detailed Coarsening Process}
\begin{figure}[t]
    \centering
\includegraphics[width=\linewidth]{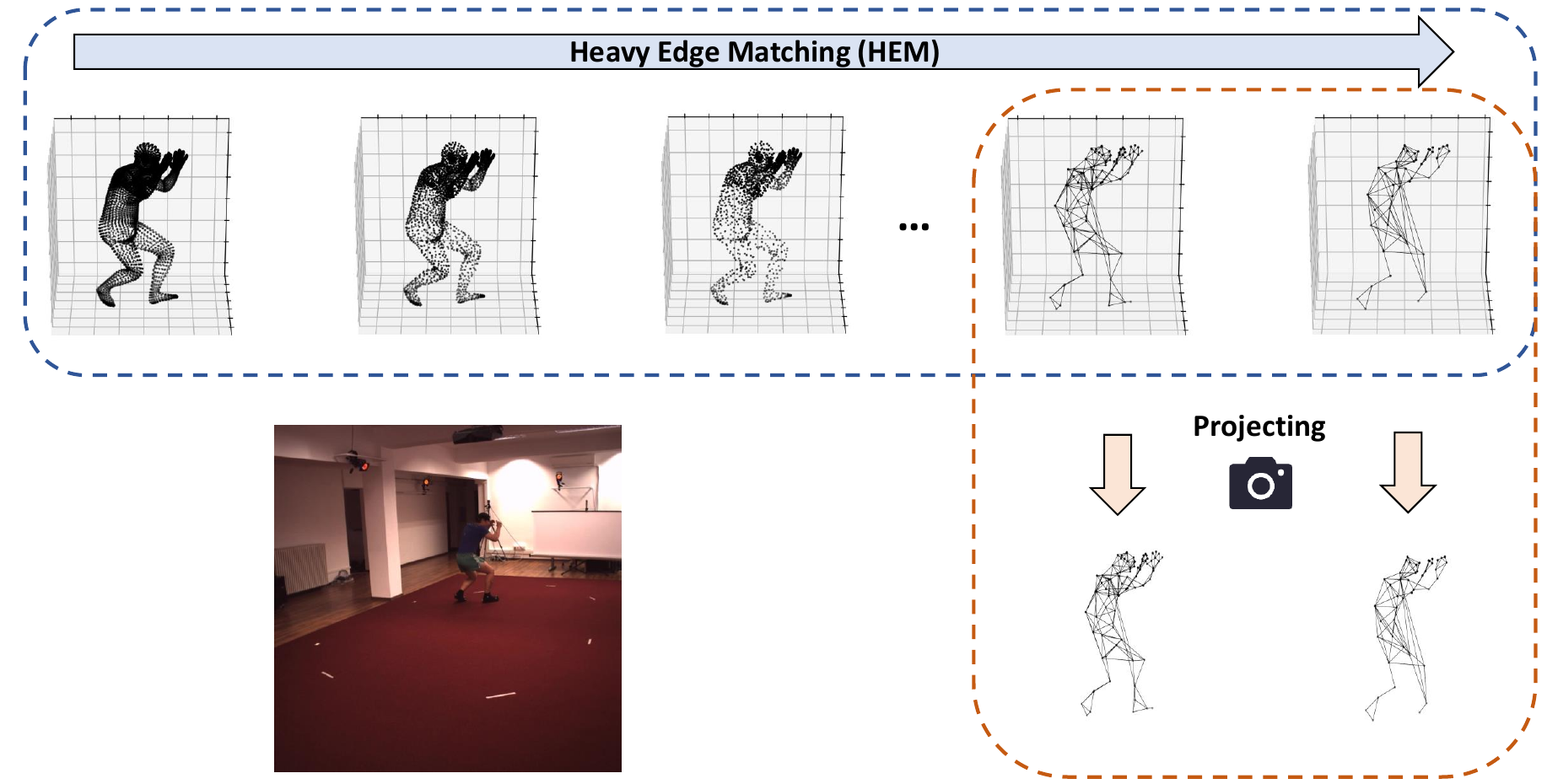}
    \caption{The overview of the coarsening process, consisting of two steps. We first progressively coarsen a human mesh with 6890 vertices via Heavy Edge Matching (HEM)~\cite{defferrard2016convolutional}. Then we project these 3D poses into 2D pixel space. }
    \label{fig_app:coarsening}
\end{figure}

Due to the absence of the hierarchical 2D pose dataset, we construct one as the ground truth for training. We use the pseudo-ground truth 3D mesh provided by Pose2Mesh~[7] for Human3.6M. It is widely used in 3D human mesh recovery. Following the dense vertices coarsening method from \cite{choi2020pose2mesh}, the process is split into two steps. As shown in Fig.~\ref{fig_app:coarsening}, we first progressively coarsen a human mesh graph with 6890 vertices via Heavy Edge Matching (HEM)~\cite{defferrard2016convolutional}, selecting 96 and 48 joints to represent different levels of human skeleton structure. Then, we project these 3D poses into 2D pixel space and obtain three levels of 2D poses: sparse, dense, and fine, denoted as ${\mathbf{x}_s, \mathbf{x}_d, \mathbf{x}_f}$, with $\mathbf{x}_s \in \mathbb{R}^{J_s \times 2}$, $\mathbf{x}_d \in \mathbb{R}^{J_d \times 2}$, and $\mathbf{x}_f \in \mathbb{R}^{J_f \times 2}$.

\section{Detailed Experimental Setup}
\begin{figure}[t]
    \centering
\includegraphics[width=0.95\linewidth]{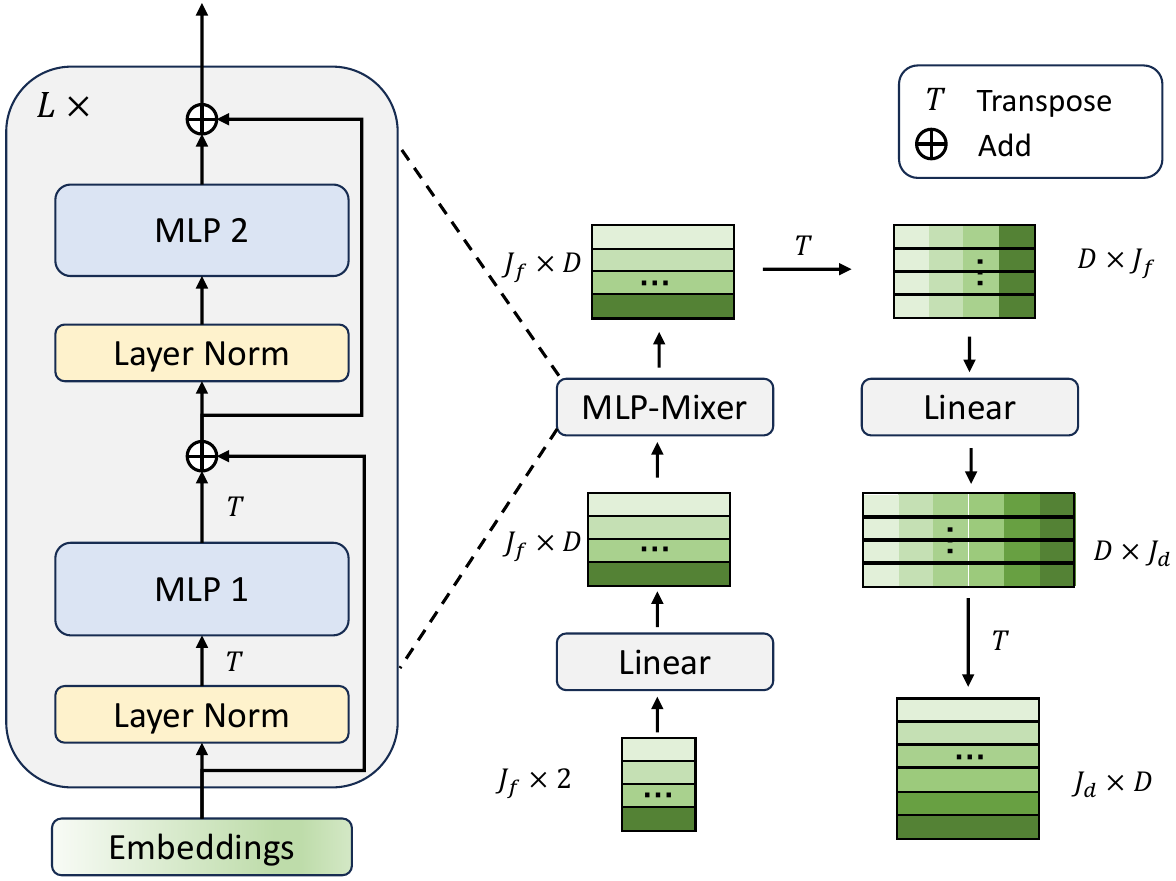}
    \caption{The detailed structure of the encoder of MSST, taking $\mathcal{E}_f$ as the example. Following PCT~\cite{geng2023human}, the fine pose is first fed to a linear projection layer to transform the embedding dimension. Subsequently, these enhanced embeddings are passed through $L$ MLP-Mixer blocks~\cite{tolstikhin2021mlp}, which deeply fuse the pose feature. We  can finally obtain encoded embeddings  by applying a linear projection along the joint axis and transposing the embeddings.}
    \label{fig_app:encoder}
\end{figure}

% \noindent\textbf{Implementation details.} 
% \beta \lambda_{local} \lambda_{global} \tau; bs lr wd epoch 
% \lamda_{d} bs lr wd epoch dropout
%  bs lr wd epoch dropout
% \noindent\textbf{Stage 1.} 
We train the MSST with a batch size of 128 for 20 epochs using the AdamW optimizer~\cite{loshchilov2017decoupled}. The learning rate is initialized at 1e-3  with a weight decay of 0.15, warmed up over the first 500 iterations, and subsequently decayed following a cosine schedule.  We set $\beta$, $\lambda_{local}$, $\lambda_{global}$, and $\tau$ to 0.25, 1.0, 0.3, and 0.07, respectively. The detailed structure of the encoder is shown in Fig.~\ref{fig_app:encoder}.

% \noindent\textbf{Stage 2.} 
We train the HiARM with a batch size of 64 for 50 epochs using the AdamW optimizer.  The learning rate is  initialized at 5e-4 with a weight decay of 0.03. The $\lambda_{d}$ is set to 1.5. The dropout rate of the transformer block is set to 0.25. 

For the inference process of HiPART, we select the index with the highest probability from the predicted vectors to generate discrete tokens.

\begin{figure}[t]
    \centering
\includegraphics[width=0.95\linewidth]{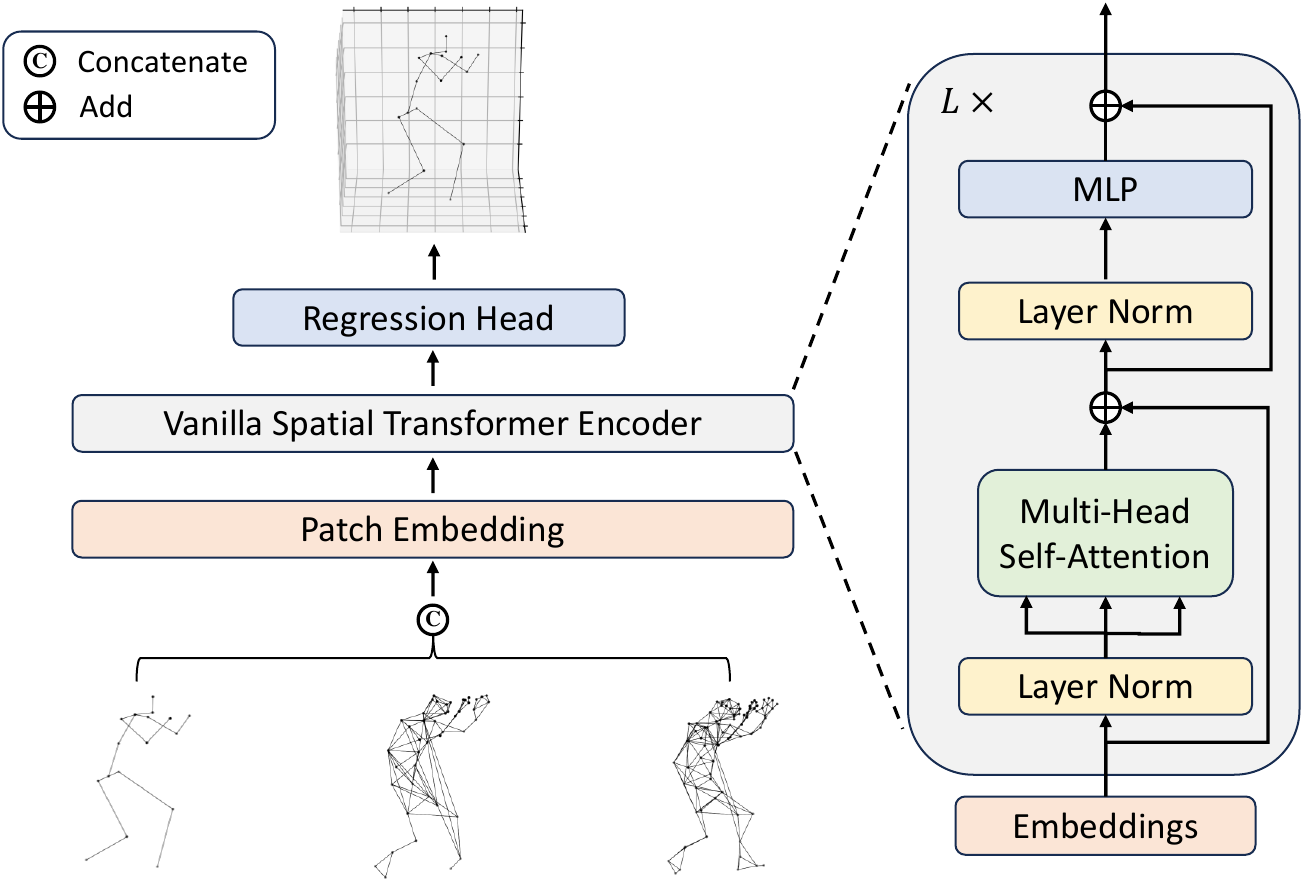}
    \caption{The overview of the lifting model, consisting of patch embedding, vanilla spatial transformer encoder, and regression head. We concatenate three levels of 2D poses along the joint axis as the input to patch embedding and vanilla spatial transformer encoder. Then the corresponding embeddings are regressed into the target 3D pose. }
    \label{fig_app:lifting}
\end{figure}

% \noindent\textbf{Lifting step.} 
For the lifting stage, we adopt the Adam~\cite{kingma2014adam} optimizer. The learning rate is
initialized to 1e-3 and decayed by 0.96 per 4 epochs, and we train the model for 25 epochs using a batch size of 256. The overview of the lifting model is shown in Fig.~\ref{fig_app:lifting}.

Our experiments are conducted on one NVIDIA Tesla V100 GPU with the CentOS 7 system, using PyTorch 1.11.0 and Torchvision 0.12.

\section{Additional Discussion on Related Work}

\setlength{\itemindent}{10em}
\renewcommand{\algorithmicrequire}{\textbf{Input:}}
\renewcommand{\algorithmicensure}{\textbf{Require:}}
\begin{algorithm}[t]
\caption{Stage 1: Multi-Scale Skeletal Tokenization (MSST).}
\label{alg:stage1}
\begin{algorithmic}[1]
\REQUIRE The dense and fine 2D poses $\mathbf{x}_d$ and $\mathbf{x}_f$, encoders $\mathcal{E}_d$ and $\mathcal{E}_f$, decoders  $\mathcal{D}_s$, $\mathcal{D}_d$ and $\mathcal{D}_f$, sparse and dense codebooks $\mathbf{C}_s$ and $\mathbf{C}_d$, action label $\mathbf{y}_A$, weighting factors $\beta$, $\lambda_{local}$ and $\lambda_{global}$, temperature parameter $\tau$, the number of iterations $T$.
% \ENSURE .\\

\FOR{$t=0$ to $T$}
\STATE \textbf{\{Forward pass\}}
% \STATE Obtain dense and sparse pose embeddings by encoding: $\mathbf{z}_d \gets \mathcal{E}_f(\mathbf{x}_f)$ and $\mathbf{z}_s \gets \mathcal{E}_d(\mathbf{x}_d)$.
% \STATE Generate sparse tokens and sparse token embeddings: $\mathbf{q}_s \gets \mathcal{Q}(\mathbf{z}_s)$, $\hat{\mathbf{z}}_s \gets \mathbf{C}_s(\mathbf{q}_s)$.
% \STATE \operatorname{Concat}enate dense pose embeddings with upsampled sparse token embeddings: $\mathbf{z}^\prime_d \gets \operatorname{Concat}(\mathbf{z}_d, \mathcal{D}_s(\hat{\mathbf{z}}_s))$.
% \STATE Generate dense tokens and dense token embeddings: $\mathbf{q}_d \gets \mathcal{Q}(\mathbf{z}^\prime_d)$, $\hat{\mathbf{z}}_d \gets \mathbf{C}_d(\mathbf{q}_d)$.
% \STATE Obtain reconstructed dense and fine poses by decoding: $\hat{\mathbf{x}}_d \gets \mathcal{D}_d(\hat{\mathbf{z}}_d)$ and $\hat{\mathbf{x}}_f \gets \mathcal{D}_f(\operatorname{Concat}(\hat{\mathbf{z}}_d, \mathcal{D}_s(\hat{\mathbf{z}}_s)))$.
% \STATE Obtain the action classification vector: $\mathbf{p}_A \gets \mathcal{P}_A(\operatorname{Concat}(\hat{\mathbf{z}}_d, \mathcal{D}_s(\hat{\mathbf{z}}_s)))$. 
\STATE $\mathbf{z}_d \gets \mathcal{E}_f(\mathbf{x}_f)$, $\mathbf{z}_s \gets \mathcal{E}_d(\mathbf{z}_d)$.
\STATE $\mathbf{q}_s \gets \mathcal{Q}(\mathbf{z}_s)$, $\hat{\mathbf{z}}_s \gets \mathbf{C}_s(\mathbf{q}_s)$.
\STATE $\mathbf{z}^\prime_d \gets \operatorname{Concat}(\mathbf{z}_d, \mathcal{D}_s(\hat{\mathbf{z}}_s))$.
\STATE  $\mathbf{q}_d \gets \mathcal{Q}(\mathbf{z}^\prime_d)$, $\hat{\mathbf{z}}_d \gets \mathbf{C}_d(\mathbf{q}_d)$.
\STATE  $\hat{\mathbf{x}}_d \gets \mathcal{D}_d(\mathbf{q}_d)$, $\hat{\mathbf{x}}_f \gets \mathcal{D}_f(\mathbf{q}_d, \mathcal{D}_s(\mathbf{q}_s))$.
\STATE  $\mathbf{p}_A \gets \mathcal{P}_A(\operatorname{Concat}(\hat{\mathbf{z}}_d, \mathcal{D}_s(\hat{\mathbf{z}}_s)))$. 
\STATE \textbf{\{Loss calculation\}}
\STATE Compute the local and global alignment losses according to Eq.~\textcolor{darkblue}{3}, ~\textcolor{darkblue}{5}.
\STATE Compute the Stage 1 loss $\mathcal{L}_1$ based on Eq.~\textcolor{darkblue}{6}.
\STATE Update the model based on $\nabla \mathcal{L}_1$.
\ENDFOR
\RETURN
% \RETURN The reconstructed dense and fine 2D pose $\hat{\mathbf{x}}_d$ and $\hat{\mathbf{x}}_f$.

\end{algorithmic}
\end{algorithm}

\renewcommand{\algorithmicrequire}{\textbf{Input:}}
\renewcommand{\algorithmicensure}{\textbf{Require:}}
\begin{algorithm}[t]
\caption{Stage 2: Hierarchcal AutoRegressive Modeling (HiARM).}
\label{alg:stage2}
\begin{algorithmic}[1]

\REQUIRE The sparse and dense tokens $\mathbf{q}_s$ and $\mathbf{q}_d$, the sparse 2D pose $\mathbf{x}_s$, the weighting factor $\lambda_{d}$, the number of iterations $T$.
\FOR{$t=0$ to $T$}
\STATE \textbf{\{Forward pass\}}
% \STATE Obtain the sparse and dense token embeddings: $\hat{\mathbf{z}}_s \gets \mathbf{C}_s(\mathbf{q}_s)$, $\hat{\mathbf{z}}_d \gets \mathbf{C}_d(\mathbf{q}_d)$
% \STATE Refine the sparse and dense token embeddings through LSAB via Eq.\ref{equ:8}.
% \STATE Compute the average refined embedding: $\mathbf{g}^i \gets avg(\{\mathbf{g}^i_j\}_{j=0,1,\dots,r})$.
% \STATE Generate the hidden embeddings by GCSAB from center to periphery via Eq.~\ref{equ:9}.
% \STATE Predict the sparse and dense tokens from sparse to dense via Eq.~\ref{equ:10}.
% \STATE \textbf{\{Loss calculation\}}
\STATE $\hat{\mathbf{z}}_s \gets \mathbf{C}_s(\mathbf{q}_s)$, $\hat{\mathbf{z}}_d \gets \mathbf{C}_d(\mathbf{q}_d)$
\STATE $\{\mathbf{g}^i_j\}_{j=0,1,\dots,r} \gets \operatorname{LSAB}(\hat{\mathbf{z}}^i_s, \hat{\mathbf{z}}^{(i,1)}_d, \dots, \hat{\mathbf{z}}^{(i,r)}_d) $
\STATE $\mathbf{g}^i = \operatorname{avg}(\mathbf{g}^i_0, \mathbf{g}^i_1, \dots, \mathbf{g}^i_r)$
\STATE $ \{\mathbf{h}^k\}_{k=0,1,\dots,i} \gets \operatorname{GCSAB}(\mathbf{g}^0, \mathbf{g}^1, \dots, \mathbf{g}^{i})$
\STATE $\{p^{i+1}_j\}_{j=0,1,\dots,r} \gets \operatorname{PH} (\mathbf{h}^i, \hat{\mathbf{z}}_s^i, \dots,  \hat{\mathbf{z}}_s^i) $
\STATE \textbf{\{Loss calculation\}}
\STATE Compute the Stage 2 loss $\mathcal{L}_2$ based on Eq.~\textcolor{darkblue}{12}.
\STATE Update the model based on $\nabla \mathcal{L}_2$.
\ENDFOR
\RETURN

\end{algorithmic}
\end{algorithm}

\noindent\textbf{Hierarchical AutoregRessive Models.}  
% The paradigm of autoregressive modeling was first proposed in language modeling and has shown promising results in image generation. 
VQ-VAE~\cite{van2017neural} has pioneered a two-stage image generation process, which involves initially quantizing images into discrete tokens, followed by their reconstruction in the subsequent stage. 
Based on VQ-VAE, many subsequent works leverage  hierarchical discrete tokens for coarse-to-fine image generation. For instance, VQ-VAE-2~\cite{razavi2019generating} uses models of different sizes for top and bottom tokens, while Hierarchical VQ-VAE~\cite{peng2021generating} creates two levels of tokens to disentangle structural and textural image information. Our method differs from these works in three key aspects: \textbf{(1)} Human skeletons, with their non-Euclidean structure, require a tailored model and regression prediction order distinct from those used in conventional image data. \textbf{(2)} Compared to the unengaged hierarchical tokens in image generation, we give specific meanings (\emph{i.e.}, representing the multi-level 2D poses) to the multi-scale discrete tokens  with the corresponding constraint.  \textbf{(3)} While hierarchical tokens in image generation balance code sequence length with image quality, our tokens provide multi-scale skeletal context specifically designed to tackle occlusions.

\noindent\textbf{Discrete Representation Models in 3D HPE.} Recently, several 3D human pose estimation (HPE) methods have adopted the two-stage approach to learn discrete representations. PCT~\cite{geng2023human} establishes a framework that learns a discrete codebook and then treats pose estimation as a classification problem. However, this classification approach fails to efficiently capture the latent distribution of discrete tokens. Di\textsuperscript{2}Pose~\cite{wang2024text} employs a diffusion model to generate discrete tokens, enhancing prediction accuracy, but suffers from slow inference speed due to the need for numerous sampling steps. In contrast, our method introduces a hierarchical autoregressive modeling scheme for faster and more reliable predictions. Moreover, instead of directly generating 3D poses, our approach focuses on producing hierarchical dense 2D poses in a two-stage process.

\noindent\textbf{Diffusion Models in 3D HPE.} In recent years, diffusion models have also been increasingly applied in 3D HPE. For instance, diffusion models are employed to progressively refine the pose distribution, reducing uncertainty from high to low throughout the estimation process~\cite{choi2023diffupose, feng2023DiffPose, gong2023DiffPose}. Other approaches leverage diffusion models to generate multiple pose hypotheses from a single 2D observation~\cite{holmquist2023diffpose, shan2023diffusion}, effectively addressing ambiguity in pose estimation. However, these models typically exhibit lower throughput and slower inference speeds compared to the autoregressive approach in our method, due to the extensive sampling steps needed for precision. We further explore this in our experiments detailed in Section~\ref{exp}.

\section{Additional Experiment Results}
\label{exp}
In this section, we conduct a series of additional experiments on Human3.6M to further demonstrate the effectiveness of our method. 

% \noindent\textbf{Model complexity and computation cost.}
% We compare the model complexity and computational cost of our method with those of single-frame methods and temporal-based methods. As shown in Tab.~\ref{tab:computation}, our method achieves better performance with a similar parameter count and computational load compared to single-frame methods like GraphMLP and POT. Compared to temporal-based methods, our method is more lightweight and yet offers comparable or even superior performance. For instance, MHFormer has lower accuracy than our method but requires 18.9M and 24.2 GFLOPs for lifting, while our method only consumes 2.4M and 2.2 GFLOPs, which is a 8 and 11 times reduction, respectively.

% \begin{table}[!t]
% \begin{center}
% \caption{Comparison of model complexity and computation cost across different methods. $f$ denotes the number of frames used in temporal-based methods.}
% \label{tab:computation}
% \resizebox{0.9\linewidth}{!}{%
% \begin{tabular}{llccc}
% \toprule[1pt]
% \multicolumn{2}{l}{Method}&Params(M) &FLOPs(G)&  MPJPE\\
% \hline
% \multicolumn{2}{l}{GraphMLP}& 9.5& 0.3&34.2 \\
% \multicolumn{2}{l}{POT}& 0.98 &1.6 &33.8 \\
% % \multicolumn{2}{l}{DiffPose~\cite{gong2023DiffPose} }& & &31.6 \\
% \multicolumn{2}{l}{PoseFormer ($f=81$) }&9.7 & 1.6 &31.3\\
% \multicolumn{2}{l}{MHFormer ($f=351$) }&18.9 & 24.2&30.5\\
% \multicolumn{2}{l}{MixSTE ($f=243$) }& 33.8 & 277.4&21.6\\
% \multicolumn{2}{l}{Ours } &2.4 & 2.2 & 28.3 \\
% \hline
% \toprule[1pt]
% \end{tabular}
% }
% \end{center}
% \vspace{-20pt}
% \end{table}

\noindent\textbf{Results on Human3.6M under Occlusion.} To validate the performance of our method under different occlusion conditions, we synthesize the occlusion scenarios by masking or cropping the test images. We investigate the cascaded pyramid network (CPN)~\cite{chen2018cascaded} with a ResNet-50~\cite{zagoruyko2016wide} backbone as the 2D keypoint detector to infer the 2D poses of the test set. We load the model weight from~\cite{pavllo20193d}, which is pretrained on COCO~\cite{lin2014microsoft}. After obtaining the 2D results, we compare the performance of our method with DiffPose~\cite{gong2023DiffPose}.  Tab.~\ref{tab:mask} shows that our method performs better under different occlusion conditions and exhibits stronger robustness when the occlusion worsens.

\begin{table}[t]
\centering
\caption{Results on the H36M test set under occlusion (\emph{i.e.}, mask and crop).}
\resizebox{\linewidth}{!}{%
\begin{tabular}{l|lllll}
\toprule[1pt]
Mask Ratio & 0.0 & 0.2 & 0.4 & 0.6 & 0.8 \\
\midrule
DiffPose~\cite{gong2023DiffPose}&49.7 & 64.2\textcolor{gray}{\footnotesize{$\Delta$14.5}} & 83.9\textcolor{gray}{\footnotesize{$\Delta$34.2}} & 140.3\textcolor{gray}{\footnotesize{$\Delta$90.6}} &  284.6\textcolor{gray}{\footnotesize{$\Delta$234.9}} \\
Ours &42.0 & 53.2\textcolor{gray}{\footnotesize{$\Delta$11.2}} & 77.5\textcolor{gray}{\footnotesize{$\Delta$35.5}} & 125.1\textcolor{gray}{\footnotesize{$\Delta$83.1}} &  269.4\textcolor{gray}{\footnotesize{$\Delta$227.4}}\\
\midrule
Crop Ratio & 0.0 & 0.1 & 0.2 & 0.3 & 0.4 \\
\midrule
% PCT & 50.8 & 50.9 & 51.2 & 55.0 & 74.8 \\
DiffPose~\cite{gong2023DiffPose}&49.7& 50.0\textcolor{gray}{\footnotesize{$\Delta$0.3}} &50.9\textcolor{gray}{\footnotesize{$\Delta$1.2}} & 58.5\textcolor{gray}{\footnotesize{$\Delta$8.8}} & 72.7\textcolor{gray}{\footnotesize{$\Delta$23.0}} \\
Ours &42.0 & 42.2\textcolor{gray}{\footnotesize{$\Delta$0.2}} & 42.8\textcolor{gray}{\footnotesize{$\Delta$0.8}} & 48.4\textcolor{gray}{\footnotesize{$\Delta$6.4}} & 61.5\textcolor{gray}{\footnotesize{$\Delta$19.5}}  \\
\toprule[1pt]
\end{tabular}
}
\label{tab:mask}
\end{table}

\noindent\textbf{Comparison with Diffusion Models in 3D HPE.} To compare our autoregressive method with diffusion-based methods including DiffPose~\cite{gong2023DiffPose} and DiffuPose~\cite{choi2023diffupose}, we compare the inference speed and MPJPE in Tab.~\ref{tab:diffusion}. Results show that our method incorporating a vanilla spatial transformer significantly outperforms these two diffusion models on FPS and MPJPE, demonstrating the accuracy and efficiency of our autoregressive method. Further integrating into the temporal-based method, \emph{i.e.}, MixSTE~\cite{zhang2022mixste}, achieves the best inference speed and prediction accuracy.

\begin{table}[!t]
\begin{center}
\caption{Comparison with diffusion models in 3D HPE. We compare inference speed (frame per second (FPS)), and MPJPE on Human3.6M.}
\label{tab:diffusion}
\resizebox{0.9\linewidth}{!}{%
\begin{tabular}{llcc}
\toprule[1pt]
\multicolumn{2}{l}{Method} &FPS $\uparrow$&  MPJPE $\downarrow$ \\
\hline
\multicolumn{2}{l}{DiffPose~\cite{gong2023DiffPose}}&173 &49.7  \\
\multicolumn{2}{l}{DiffuPose~\cite{choi2023diffupose}}&188 &49.4  \\
\multicolumn{2}{l}{vanilla spatial transformer w./ ours} &396 &42.0  \\
\multicolumn{2}{l}{MixSTE~\cite{zhang2022mixste} ($f=81$) w./ ours}&\textbf{681} &\textbf{39.3} \\
\hline
\toprule[1pt]
\end{tabular}
}
\end{center}
\vspace{-20pt}
\end{table}

\begin{figure}[t]
\centering
\begin{subfigure}[b]{0.95\linewidth}
  \includegraphics[width=\textwidth]{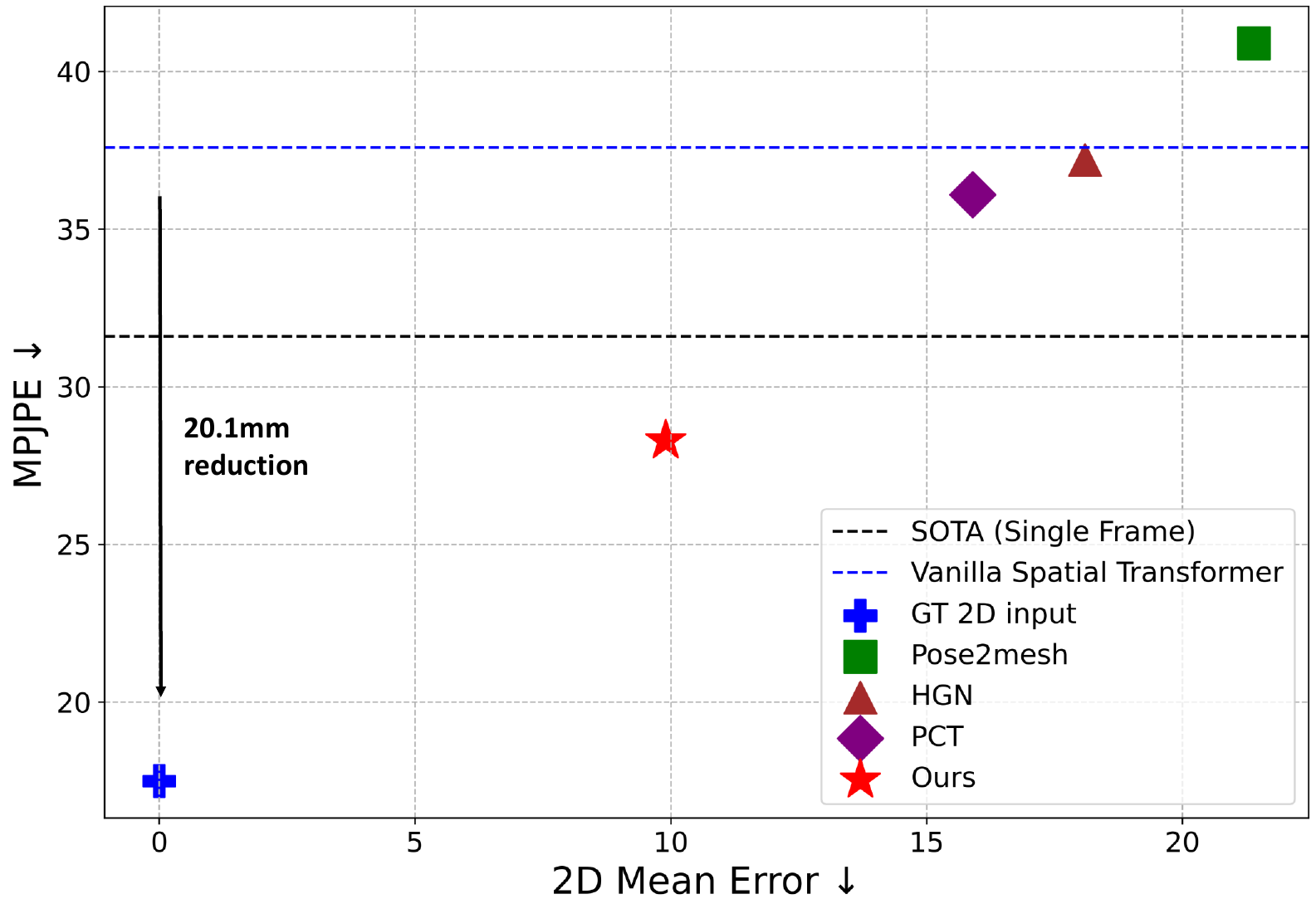}
\end{subfigure}\\
\begin{subfigure}[b]{0.95\linewidth}
  \includegraphics[width=\textwidth]{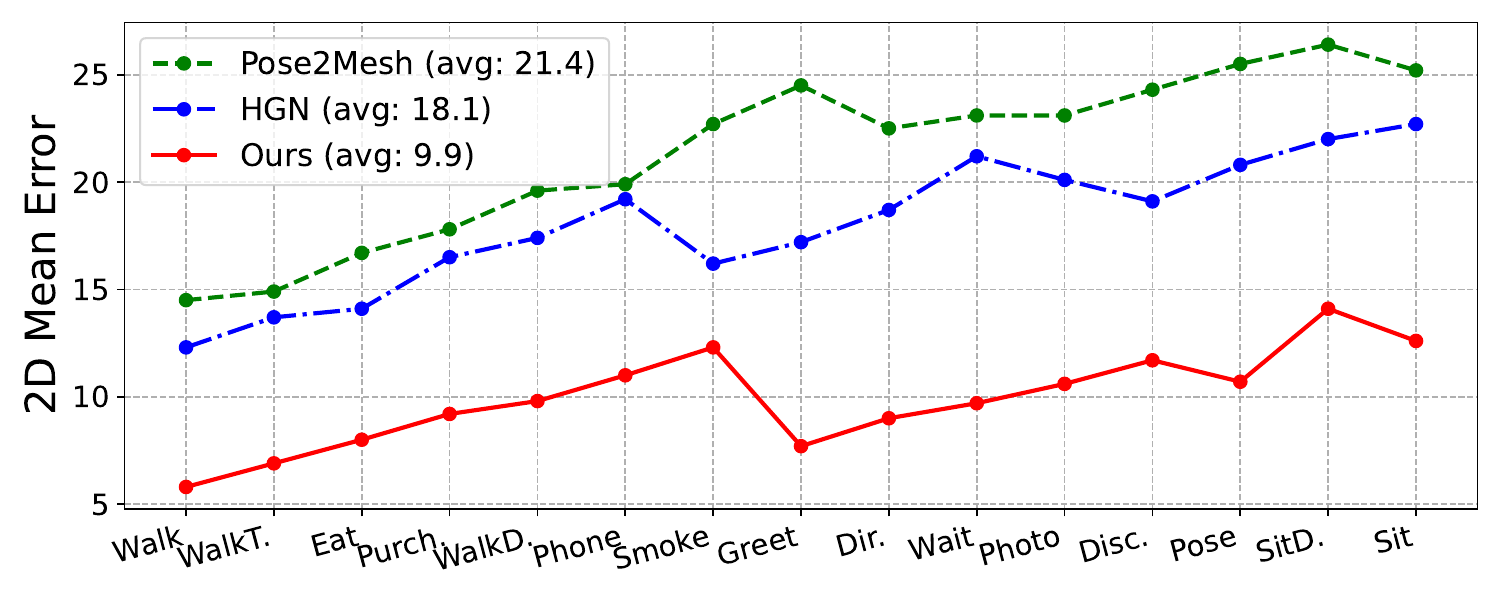}
\end{subfigure}
\caption{\textbf{Top:} The prediction accuracy of the fine 2D pose (96 joints) and lifted 3D pose (17 joints) across various methods. \textbf{Bottom:} Detailed densification results across various actions for three methods on Human3.6M using the ground truth sparse 2D pose (\emph{i.e.}, 17-joint pose) as the input.}
\label{fig_app:2dmpjpe}
\end{figure}

\noindent\textbf{Detailed Densification Results.} To evaluate the effectiveness of the hierarchical dense 2D poses, we conduct a toy experiment on Human3.6M with the ground truth 2D sparse pose. As shown in the top of Fig.~\ref{fig_app:2dmpjpe}, adding the ground truth hierarchical 2D dense poses  into a vanilla spatial transformer brings a 20.1 mm improvement of MPJPE. In real-world applications, our method achieves the best performance in 2D mean error and MPJPE compared with Pose2Mesh~\cite{choi2020pose2mesh}, HGN~\cite{li2021hierarchical}, and PCT~\cite{geng2023human}. 
Furthermore, The bottom of Fig.~\ref{fig_app:2dmpjpe}  illustrates the 2D Mean Error of the fine 2D pose across various actions for three different methods. Two conclusions can be intuitively concluded: (1) As the complexity of the action increases, \emph{i.e.}, when there is a higher frequency of occlusions, the densification performance deteriorates. (2) Our method enhances the densification performance, with an average improvement of 11.5mm for Pose2Mesh and 8.2mm for HGN. This enhancement is particularly evident for actions with severe occlusions, such as the Sit action, where our method achieves an improvement of 13.2mm for Pose2Mesh and 11.2mm for HGN.

\noindent\textbf{Ablation Study on Different Parameters of HiPART.}
Tab.~\ref{tab:parameter setting} details the effects of various parameters on our model's performance and complexity. Optimal results are achieved with 4 MLP-Mixer blocks of the MSST encoder and 12 blocks in GCSAB, with no significant improvements from adding more layers. Additionally, the results show that increasing the embedding dimension from 96 to 128 enhances performance, but dimensions larger than 128 do not yield further benefits. Therefore, we establish the default settings as $L_1 = 4$, $L_2 = 12$, and $D_1 = D_2 = 128$.

\begin{table}[!t]
\begin{center}
\caption{Ablation study on different parameters of HiPART. $L_1$ and $L_2$ denote the number of blocks of the MSST encoder and GCSAB, respectively. $D_1$ and $D_2$ are the embedding dimensions of the MSST encoder and HiARM, respectively.}
\label{tab:parameter setting}
\resizebox{0.9\linewidth}{!}{%
\begin{tabular}{cccc|ccc}
\toprule[1pt]
$L_1$&$L_2$&$D_1$&$D_2$& Params(M) &FLOPs(G)&  MPJPE\\
\hline
2& 6 & 128 & 128 &1.8 &1.41 &44.1\\
3 & 8 & 128 & 128&2.1 &1.82 & 43.5\\
4 & 12 & 128 &128 &2.4 &2.24 &\textbf{42.0}\\
5 & 16 & 128 & 128&2.7 &2.65 &42.2 \\
\hline
4 & 12 & 96 & 96 &2.0 &1.77 & 43.9\\
4 & 12 & 128 & 128&2.4 &2.24 &\textbf{42.0}\\
4 & 12 & 256 & 256 &3.6  &4.14 & 42.5 \\

\hline
\toprule[1pt]
\end{tabular}
}
\end{center}
\end{table}

\noindent\textbf{Extra Sequence-based Results.} 
% We fully agree that the results of sequence-based DiffPose and MixSTE ($f$ = 243) should be presented. 
Table~\ref{tab_rebuttal:1} provides extra results for sequence-based DiffPose and MixSTE with $f=243$. Our method is best in MPJPE under both frame and sequence based settings. 

\setlength{\textfloatsep}{5pt plus 2pt minus 2pt}

\begin{table}[!t]
\setlength\tabcolsep{3.5pt}
\setlength\abovecaptionskip{0pt}
\centering
\caption{Comparison with DiffPose and MixSTE on Human3.6M under frame (\textbf{left}) and sequence (\textbf{right}) based settings.}\label{tab_rebuttal:1}
%\begin{minipage}[t]{\textwidth}
%\begin{minipage}[t]{0.18\textwidth}
%\centering % 新增，使表格在minipage中居中
%\makeatletter\def\@captype{table}
% \caption{Comparison of test accuracy on supervised learning.}
% \vspace{-0.3cm}
%\resizebox{0.95\textwidth}{!}{
\footnotesize
\begin{tabular}{@{}lcc|lcc@{}}
\hline
Method & MPJPE &  FPS & Method & MPJPE &  FPS\\
%Method& &MPJPE&  FPS\\
\hline
MixSTE ($f$=1)& 51.1 & 358 & MixSTE ($f$=243) &40.9 & \textbf{1055}\\
DiffPose ($f$=1)&49.7 & 173 & DiffPose ($f$=243)&36.9 & 671\\
Ours ($f$=1)&\textbf{42.0} & \textbf{396} & Ours+MixSTE ($f$=243)&\textbf{36.7} & 577 \\
\hline
\end{tabular}
\vspace{-6pt}
%\end{minipage}
\end{table}

\section{Additional Visualization Analysis}
In this section, we provide additional visualization analysis to better understand our approach.

\begin{figure}[!t]
\centering
\setlength\abovecaptionskip{0pt}
\begin{subfigure}[b]{0.48\linewidth}
  \includegraphics[width=\linewidth]{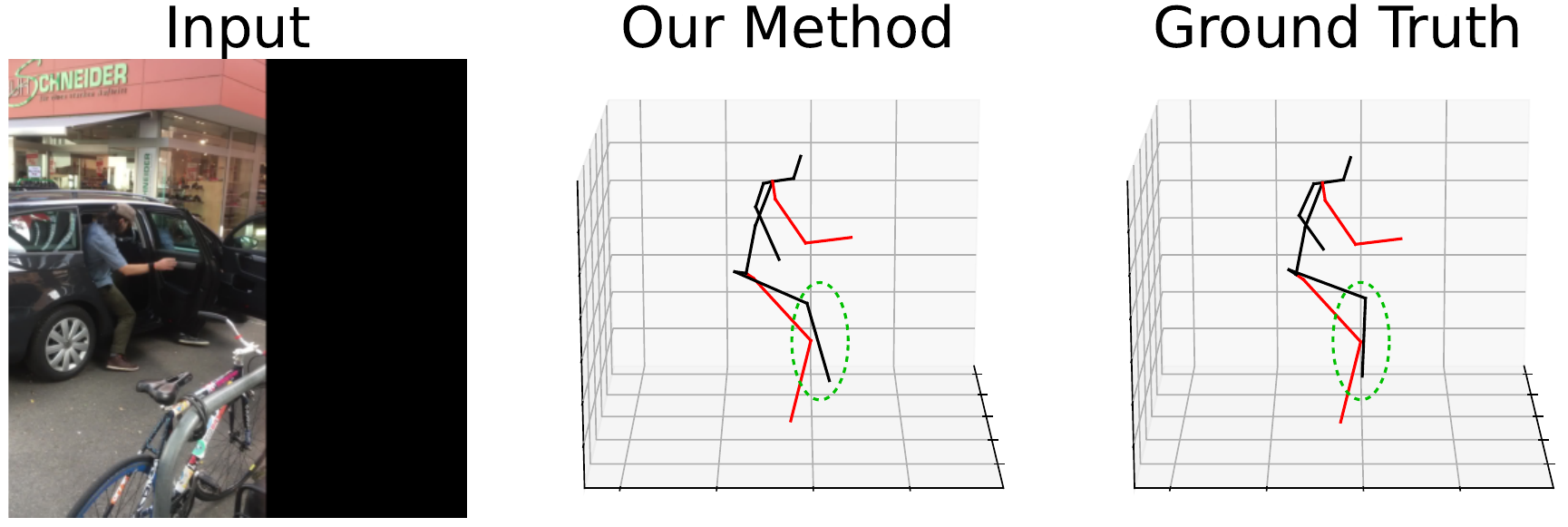}
\end{subfigure}
\begin{subfigure}[b]{0.48\linewidth}
  \includegraphics[width=\linewidth]{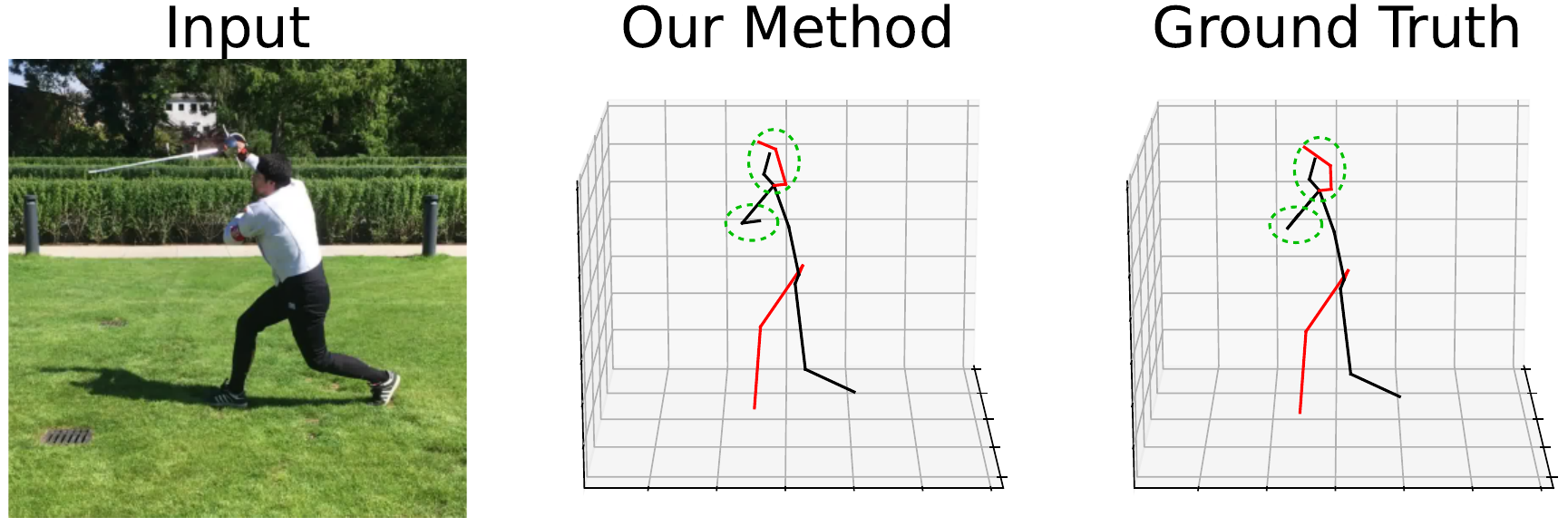}
\end{subfigure}
% \vspace{-0.1cm}
\caption{Failure cases on 3DPW. \textbf{Left:} severe occlusion where even human perception struggles to infer 3D poses. \textbf{Right:} unseen fencing motion which is absent from the training set.}
% \textbf{Left:} severe occlusion, \textbf{right:} unseen motion.
\label{figs_appendix:failure_case}
% \vspace{-10pt}
\end{figure}

\noindent\textbf{Discussion of Failure Cases.} Fig.~\ref{figs_appendix:failure_case} shows the failure cases on 3DPW. 
(\emph{i.e.}, severe occlusion and rare poses). 
% On the left side, most parts of the human body are occluded. Even for the human eye, it is difficult to estimate the accurate 3D pose, implying the domain gap problem of the learned codebook.
% Thus, we need to construct hierarchical codebooks on more datasets to cover various actions.
% This motivates expanding hierarchical codebooks across diverse datasets to enhance action coverage.

\begin{figure*}[!t]
\centering
\begin{subfigure}[b]{0.45\textwidth}
  \includegraphics[width=\textwidth]{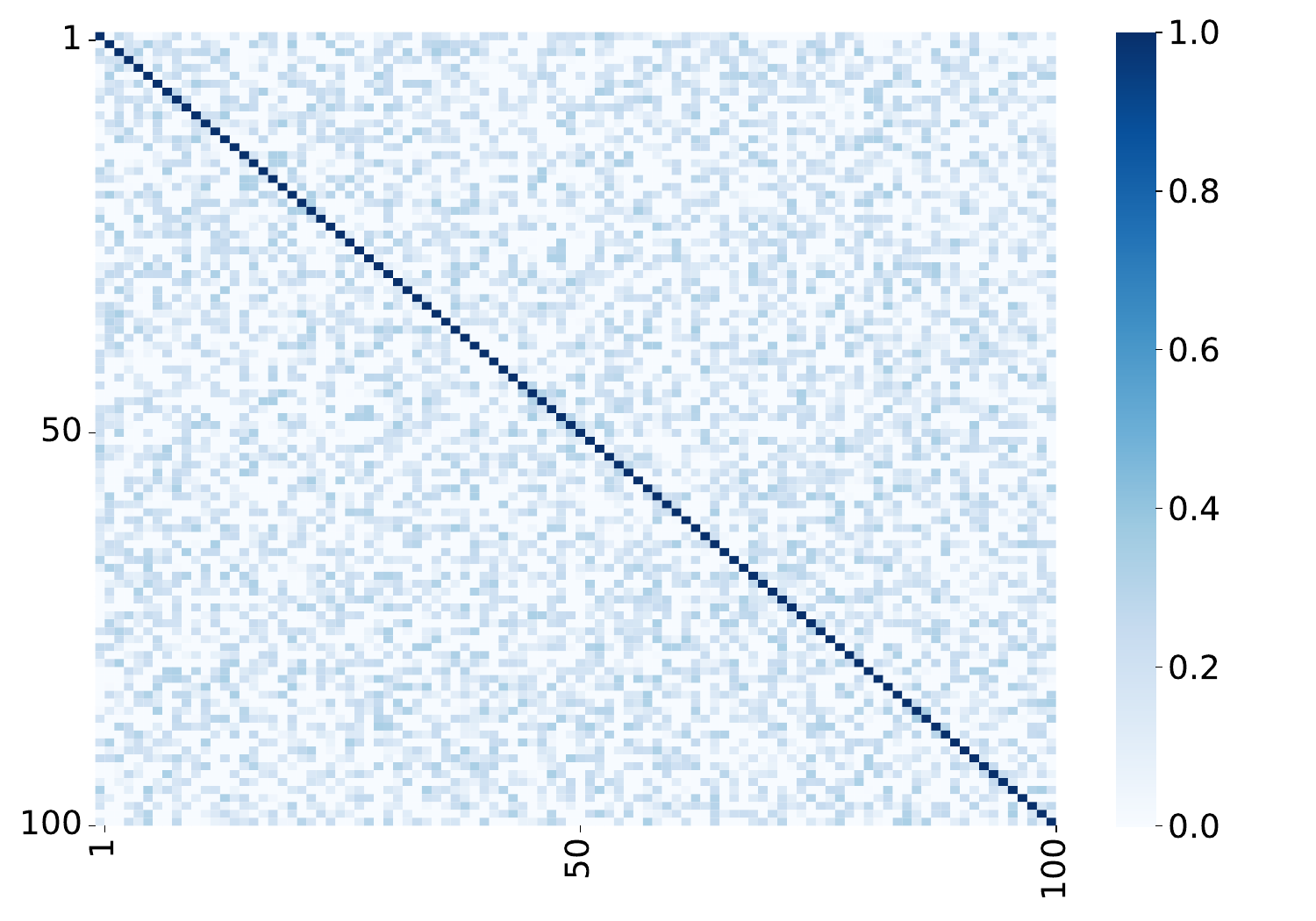}
\caption{Cosine similarity matrix of sparse tokens}
\end{subfigure}
\begin{subfigure}[b]{0.45\textwidth}
  \includegraphics[width=\textwidth]{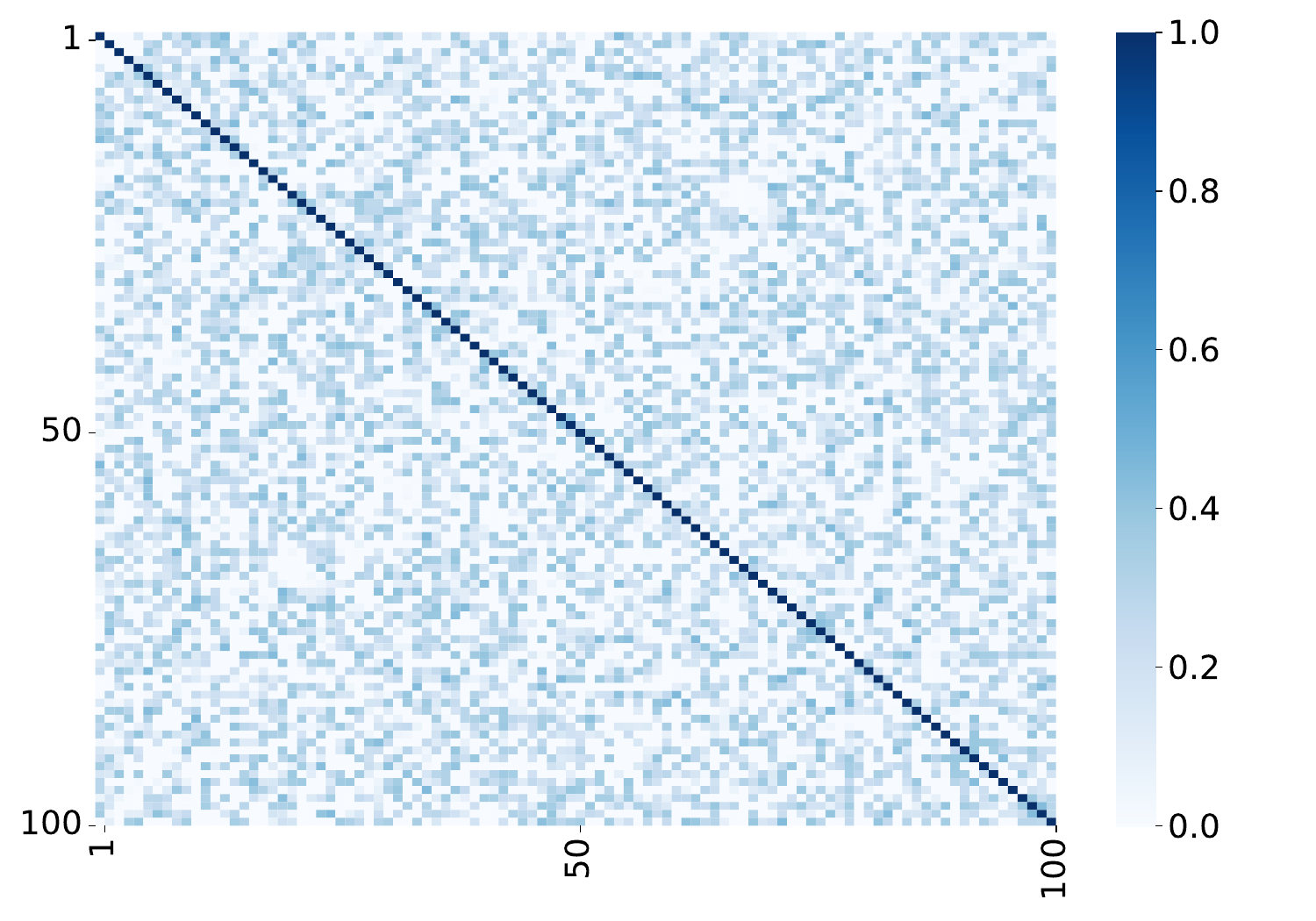}
\caption{Cosine similarity matrix of dense tokens}
\end{subfigure}\\
\begin{subfigure}[b]{0.45\textwidth}
  \includegraphics[width=\textwidth]{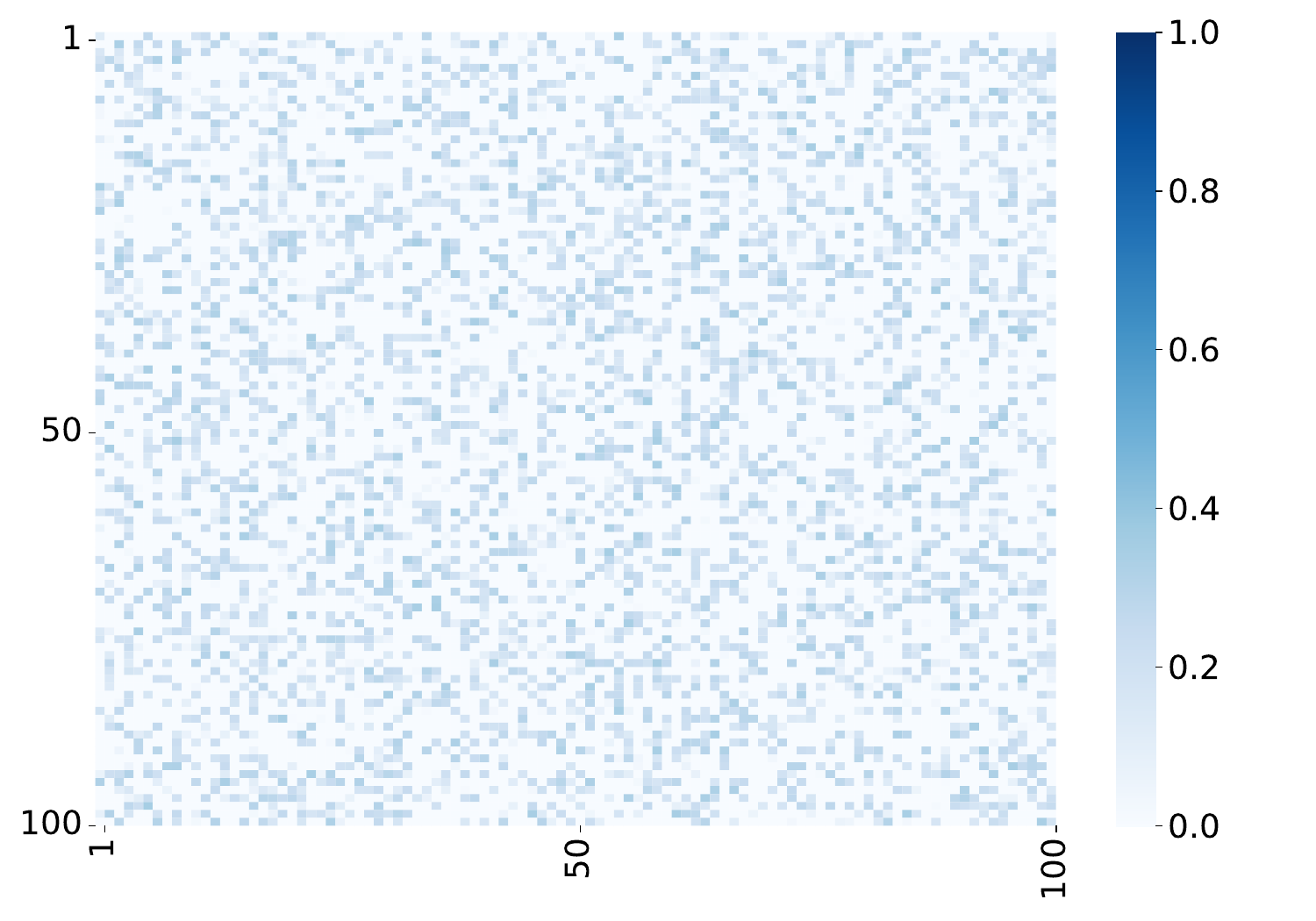}
\caption{Cosine similarity matrix of sparse  and dense tokens (w./o. Local Alignment)}
\end{subfigure}
\begin{subfigure}[b]{0.45\textwidth}
  \includegraphics[width=\textwidth]{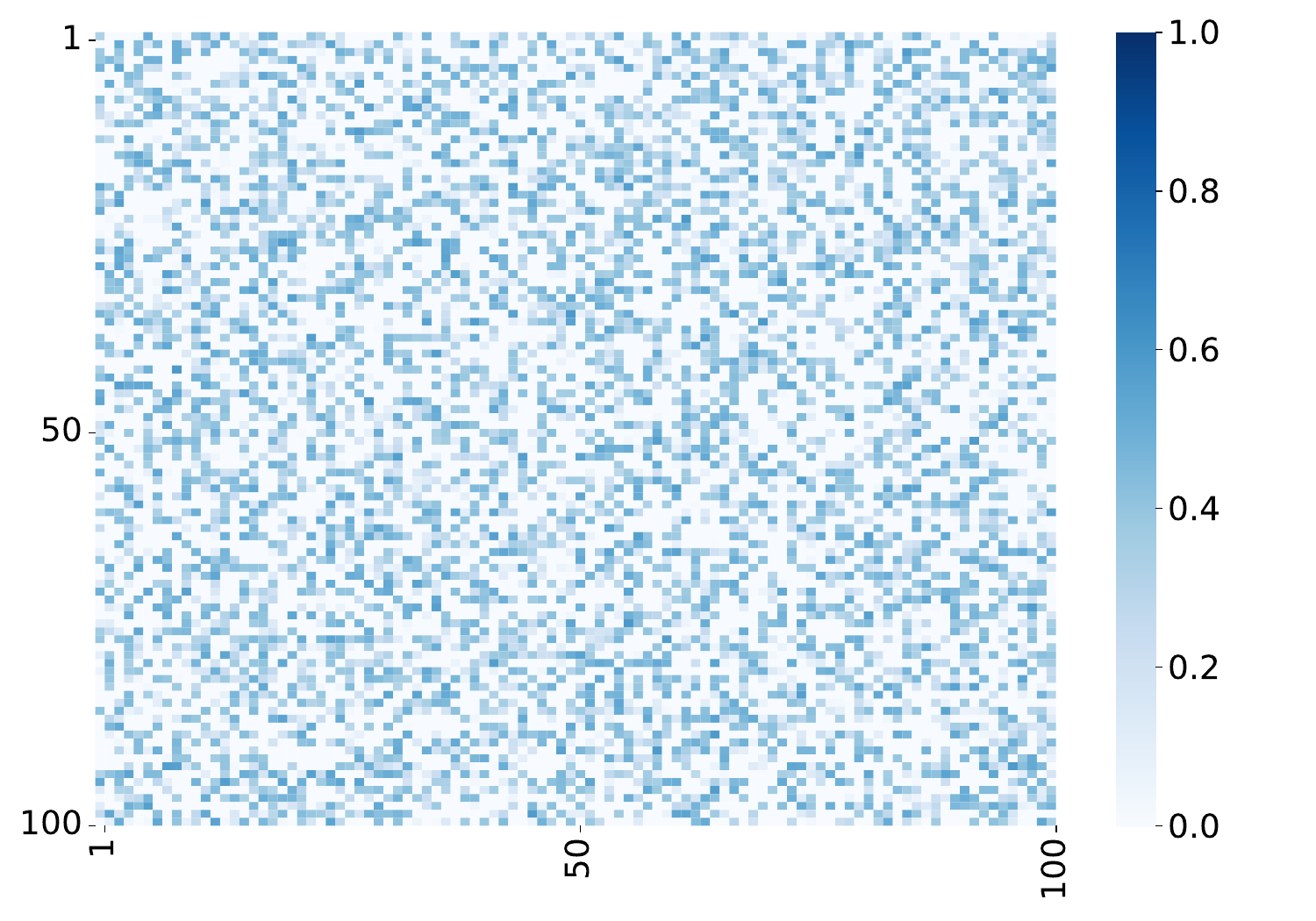}
\caption{Cosine similarity matrix of sparse  and dense tokens (w./ Local Alignment)}
\end{subfigure}
\caption{Hierarchical codebooks similarity analysis. The cosine similarity is calculated based on the \textbf{random selection of 100 tokens} from both the sparse and dense codebooks. (c) (d) The x-axis represents sparse tokens, and the y-axis represents dense tokens.}
\label{fig_app:pearason}
\end{figure*}

\noindent\textbf{More Analysis on the Similarity of Hierarchical Codebooks.} We supplement the cosine similarity matrices calculated from the randomly selected 100 tokens. As shown in Fig.~\ref{fig_app:pearason}, we can draw similar conclusions to those in the main text.

% \noindent\textbf{More visualization of the impact of hierarchical tokens on sub-structure.}

\begin{figure*}[!t]
\centering
\begin{subfigure}[b]{0.9\textwidth}
  \includegraphics[width=\textwidth]{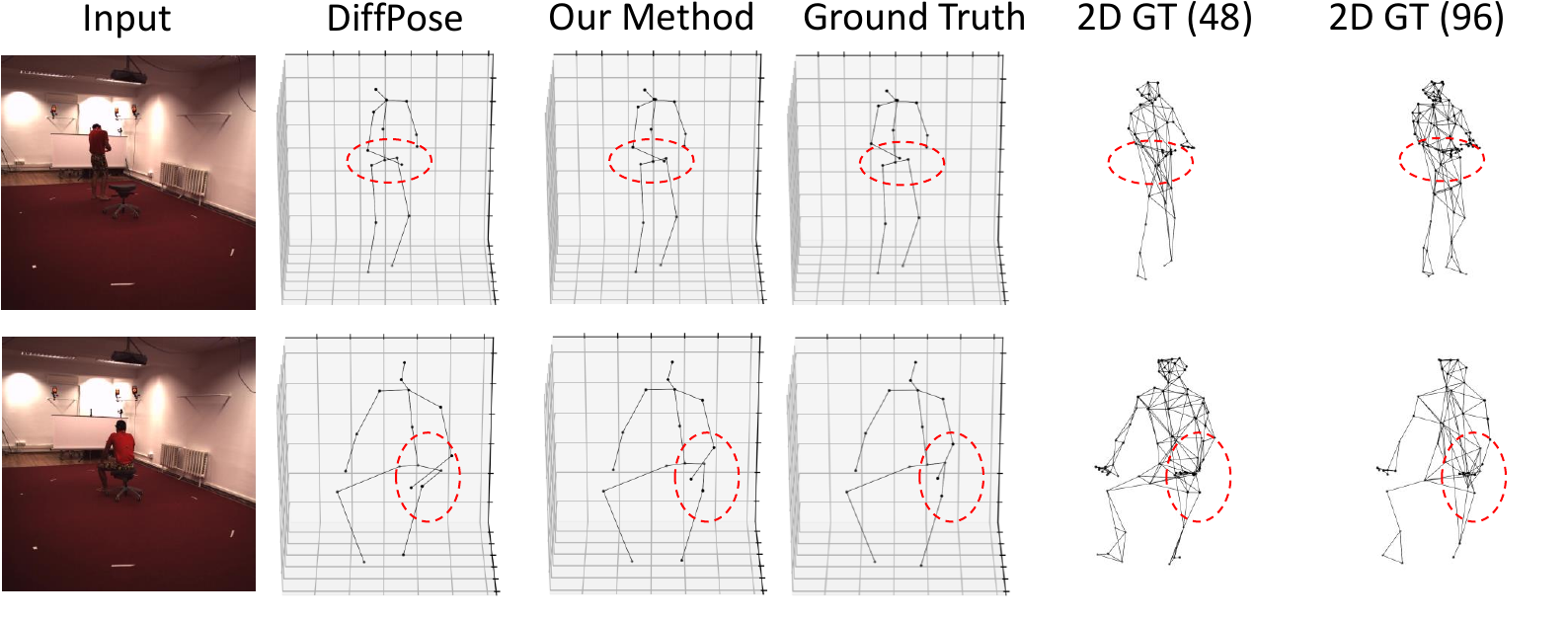}
\end{subfigure}\\
\begin{subfigure}[b]{0.9\textwidth}
  \includegraphics[width=\textwidth]{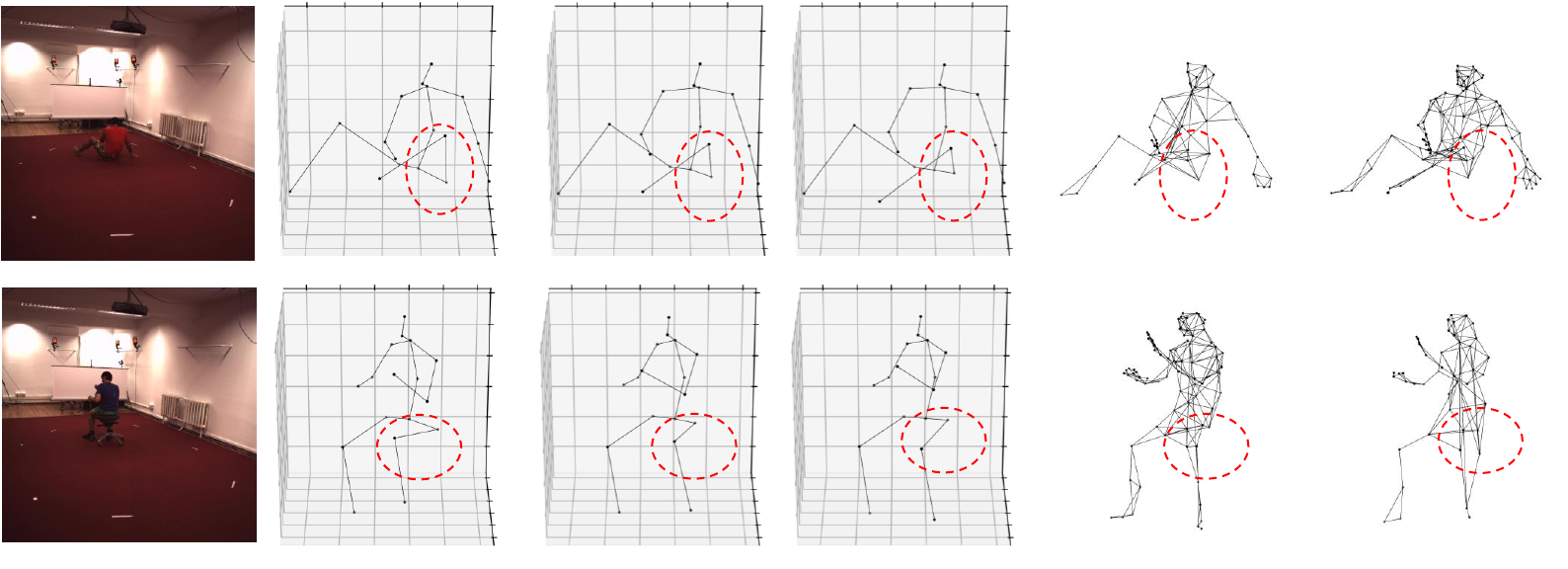}
\end{subfigure}\\
\begin{subfigure}[b]{0.9\textwidth}
  \includegraphics[width=\textwidth]{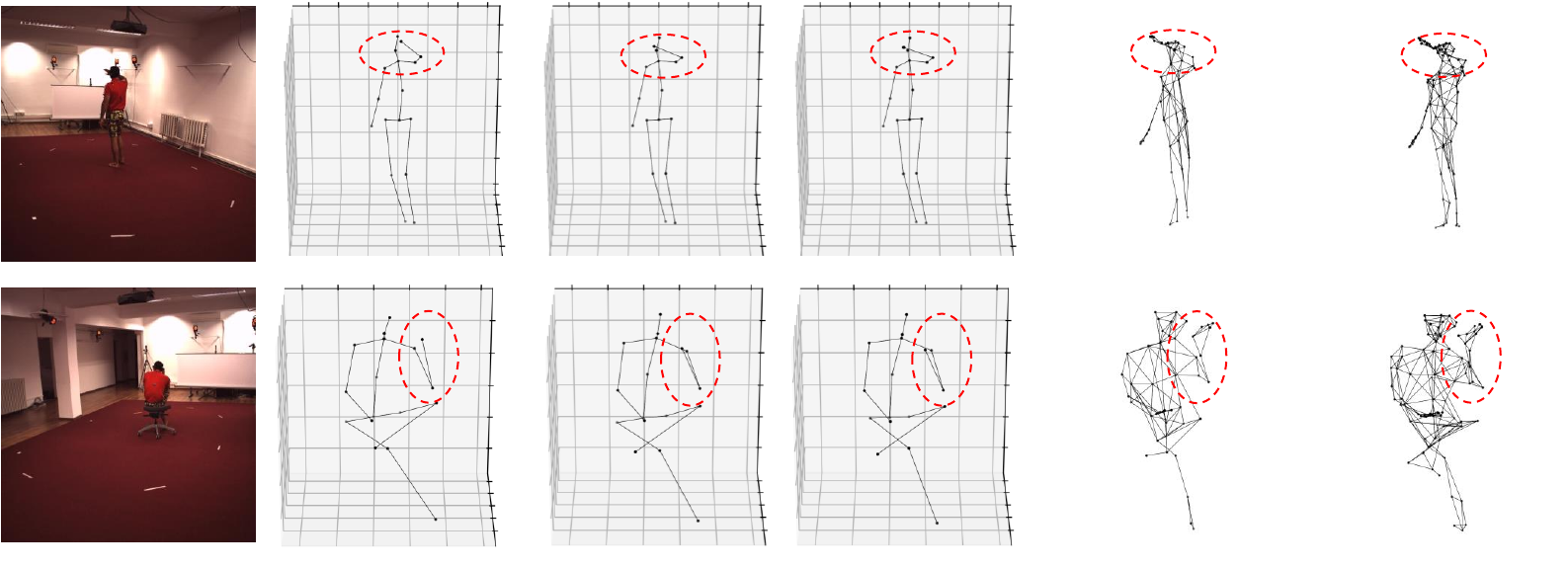}
\end{subfigure}
\caption{Qualitative results of reconstructed 3D poses and hierarchical 2D poses on Human3.6M under occlusions. }
\label{fig_app:hierarchical}
\end{figure*}

\noindent\textbf{More Visualization of Hierarchical Poses.} 
Qualitative results in Fig.\ref{fig_app:hierarchical} further demonstrate that additional joints around occluded areas provide richer skeletal information. For instance, when the right leg is occluded as shown in the fourth row of Fig.~\ref{fig_app:hierarchical}, the sparse pose offers limited support with only two joints (knee and ankle) available, while the denser pose includes multiple leg joints, capturing a more detailed structure around the occlusion and aiding in predicting the occluded right leg.

\begin{figure*}[!t]
\centering
\begin{subfigure}[b]{0.45\textwidth}
  \includegraphics[width=\textwidth]{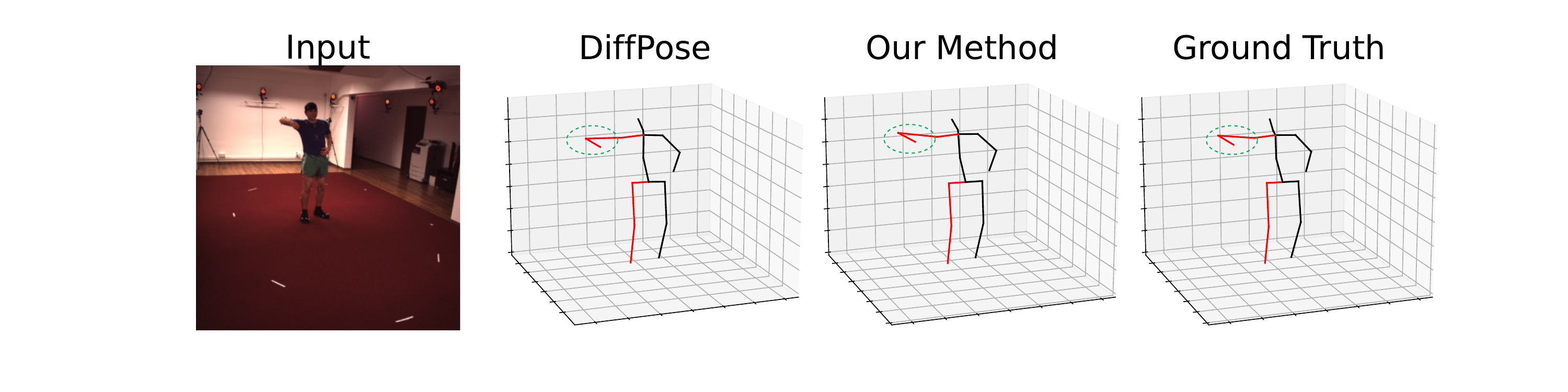}
\end{subfigure}
\begin{subfigure}[b]{0.45\textwidth}
  \includegraphics[width=\textwidth]{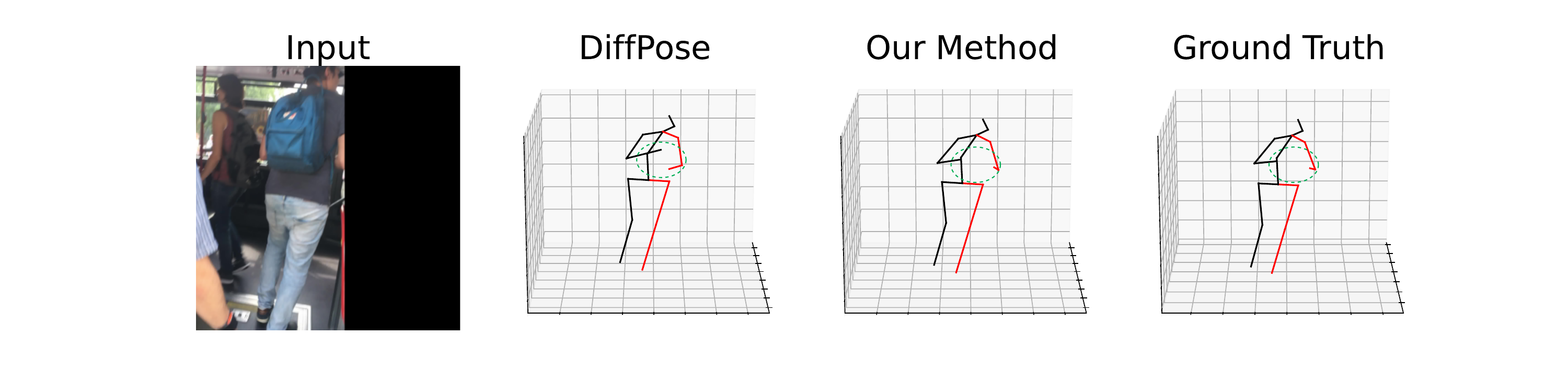}
\end{subfigure}\\
\begin{subfigure}[b]{0.45\textwidth}
  \includegraphics[width=\textwidth] {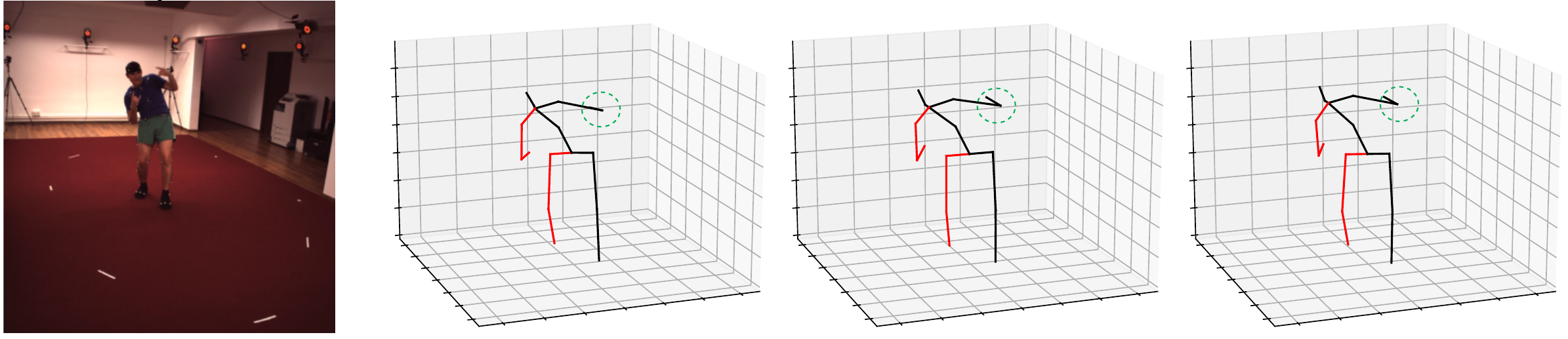}
\end{subfigure}
\begin{subfigure}[b]{0.45\textwidth}
  \includegraphics[width=\textwidth]{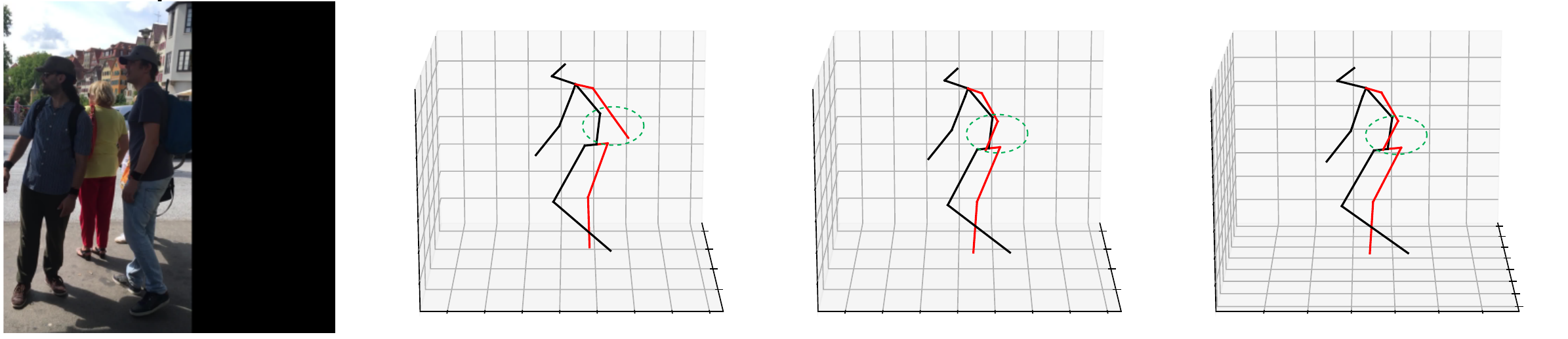}
\end{subfigure}\\
\begin{subfigure}[b]{0.45\textwidth}
  \includegraphics[width=\textwidth]{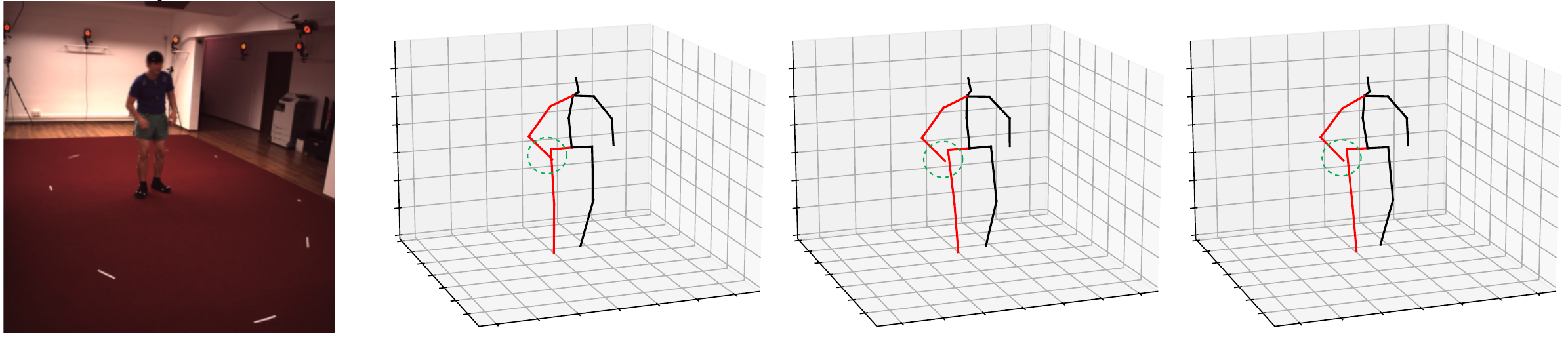}
\end{subfigure}
\begin{subfigure}[b]{0.45\textwidth}
  \includegraphics[width=\textwidth]{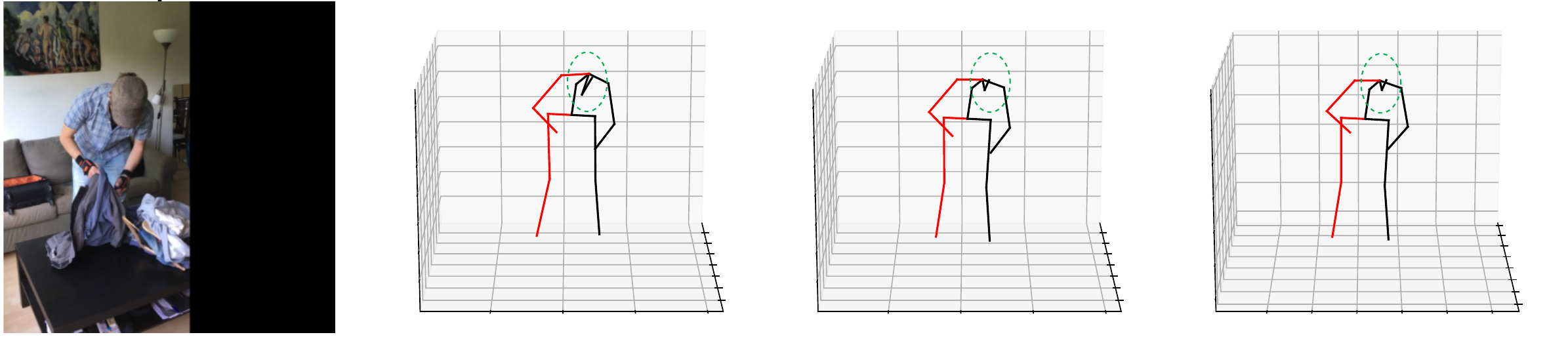}
\end{subfigure}\\
\begin{subfigure}[b]{0.45\textwidth}
  \includegraphics[width=\textwidth]{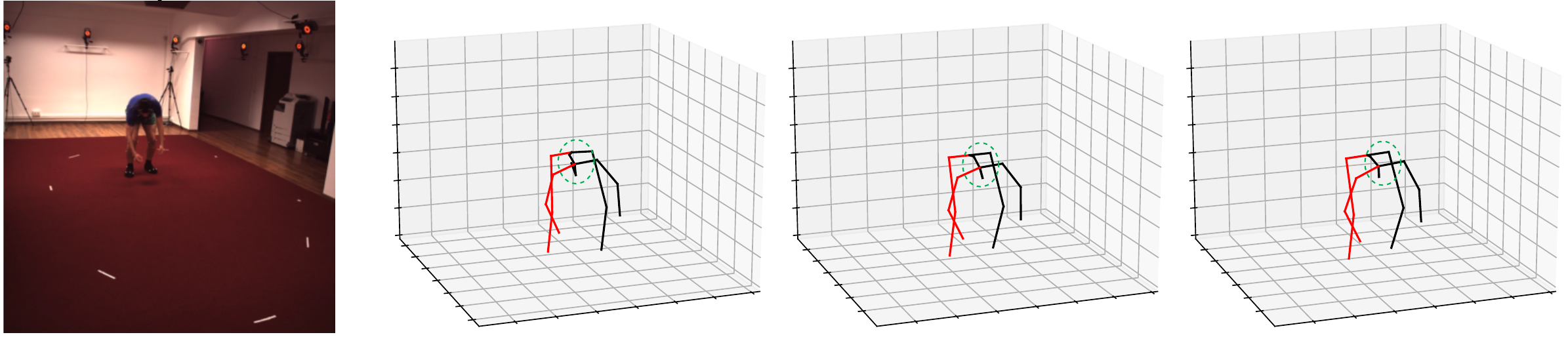}
\end{subfigure}
\begin{subfigure}[b]{0.45\textwidth}
  \includegraphics[width=\textwidth]{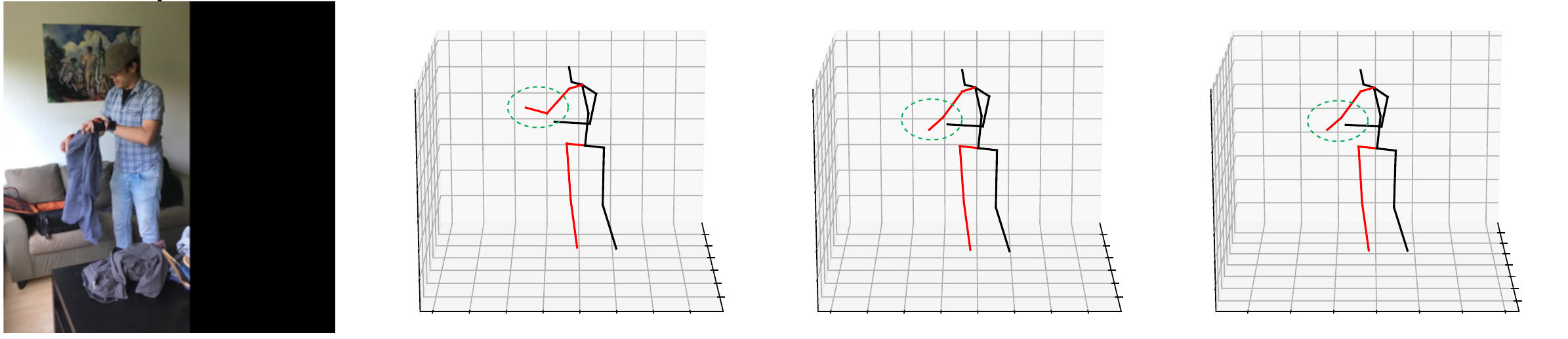}
\end{subfigure}\\
\begin{subfigure}[b]{0.45\textwidth}
  \includegraphics[width=\textwidth]{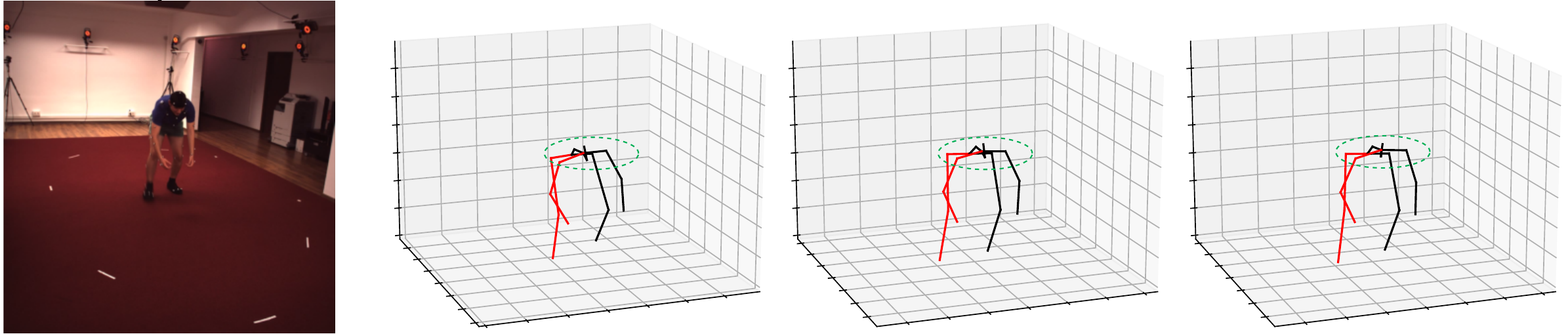}
\end{subfigure}
\begin{subfigure}[b]{0.45\textwidth}
  \includegraphics[width=\textwidth]{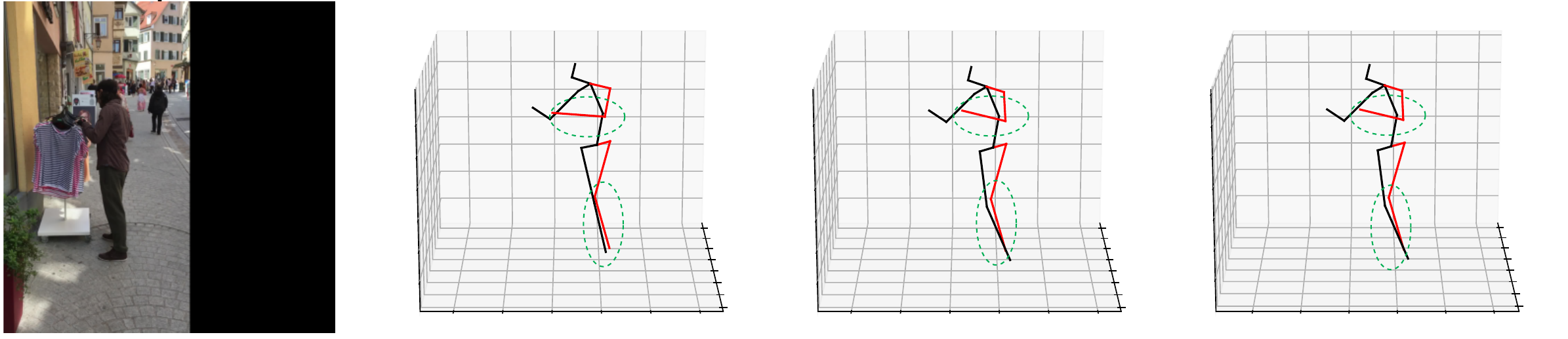}
\end{subfigure}\\
\begin{subfigure}[b]{0.45\textwidth}
  \includegraphics[width=\textwidth]{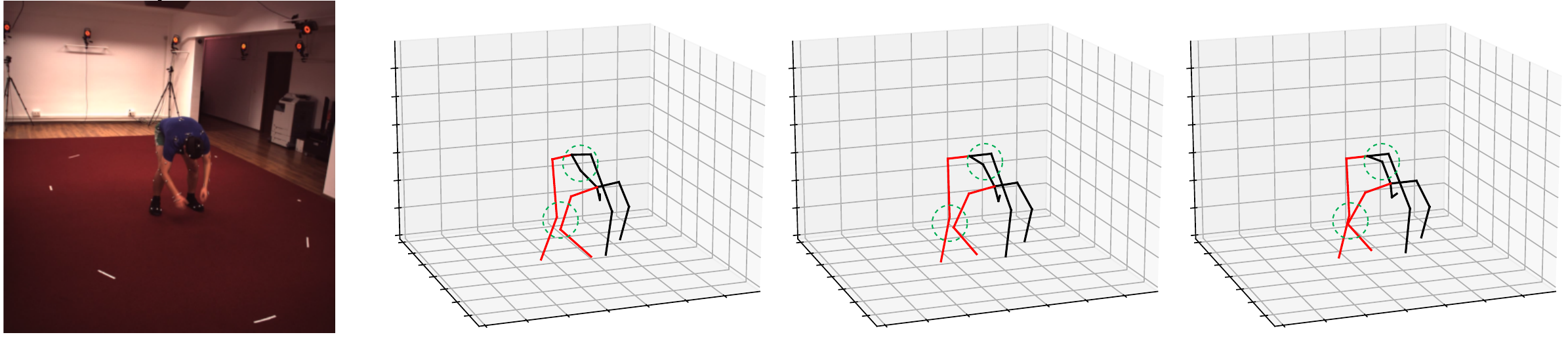}
\end{subfigure}
\begin{subfigure}[b]{0.45\textwidth}
  \includegraphics[width=\textwidth]{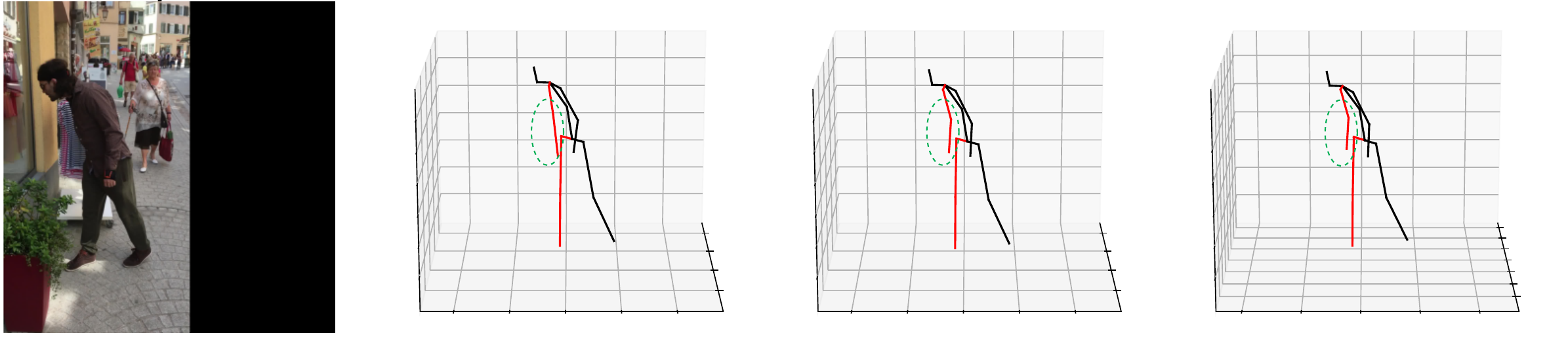}
\end{subfigure}\\
\begin{subfigure}[b]{0.45\textwidth}
  \includegraphics[width=\textwidth]{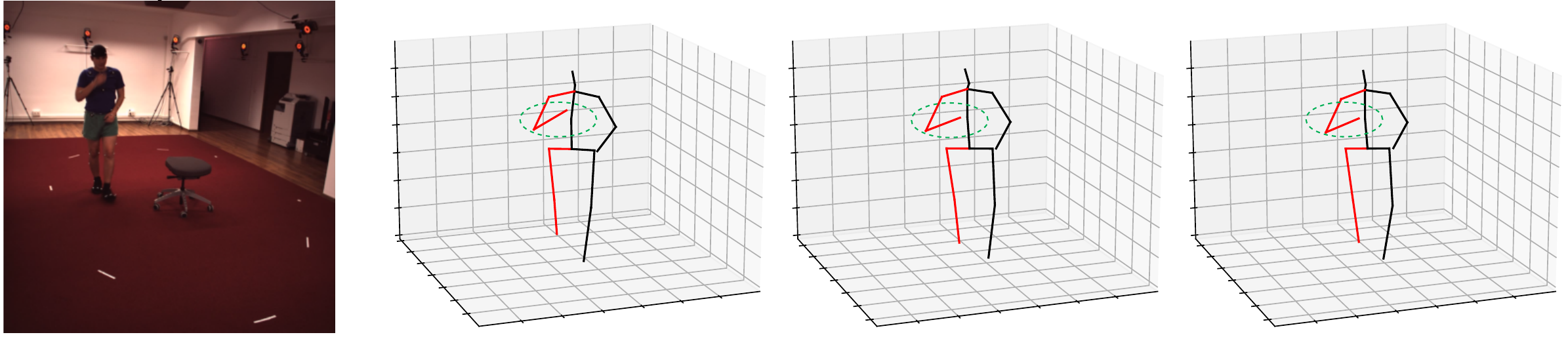}
\end{subfigure}
\begin{subfigure}[b]{0.45\textwidth}
  \includegraphics[width=\textwidth]{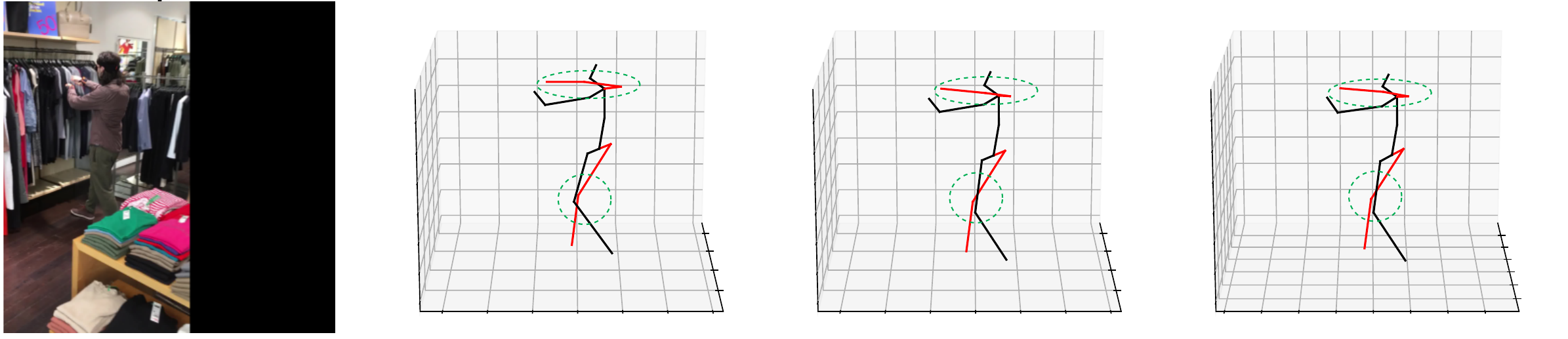}
\end{subfigure}\\
\begin{subfigure}[b]{0.45\textwidth}
  \includegraphics[width=\textwidth]{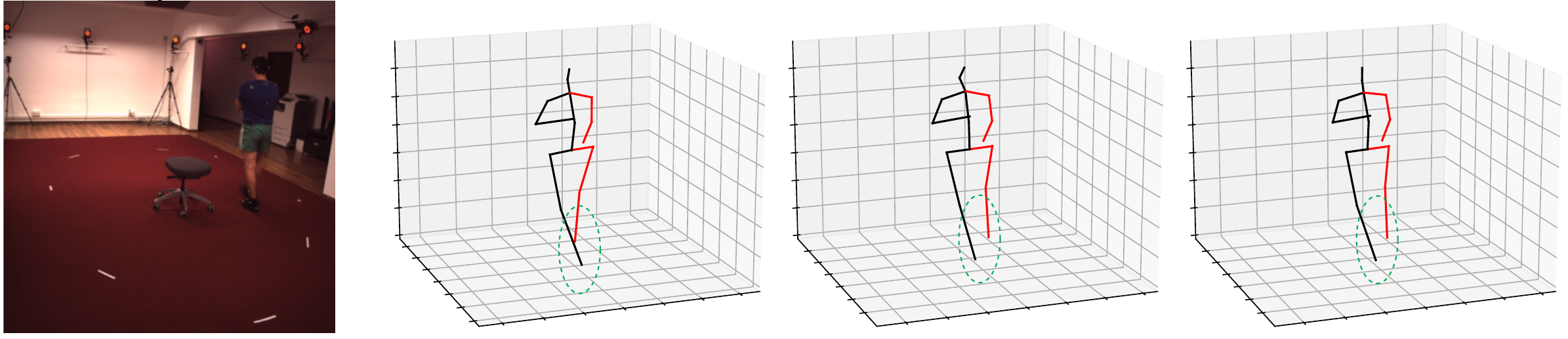}
\end{subfigure}
\begin{subfigure}[b]{0.45\textwidth}
  \includegraphics[width=\textwidth]{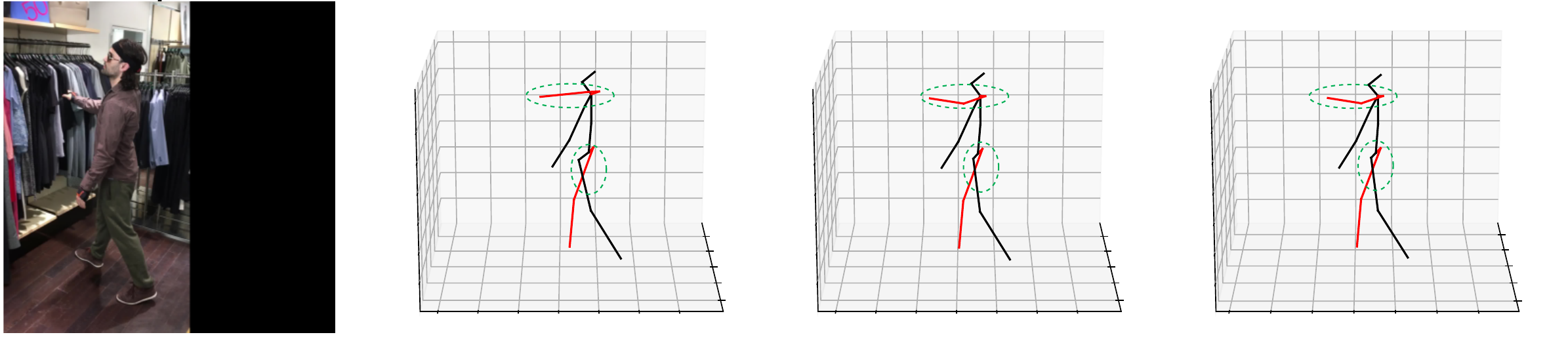}
\end{subfigure}
\caption{Qualitative results compared with DiffPose~\cite{gong2023DiffPose} on Human3.6M (\textbf{left}) and 3DPW (\textbf{right}).}
\label{fig_app:Qualitative}
\end{figure*}

\noindent\textbf{More Visualization of Poses under Occlusion.} 
As shown in Fig.\ref{fig_app:Qualitative}, we provide additional qualitative results on Human3.6M and 3DPW, comparing our method with DiffPose\cite{gong2023DiffPose}. DiffPose can predict decent results on Human3.6M, but its predictions in occluded areas become significantly poorer when generalized to the more occlusion-heavy 3DPW dataset. In contrast, our method maintains strong occlusion robustness on 3DPW. For instance, in the first row of the 3DPW data, where the back severely occludes both arms, DiffPose's predictions for the occluded area deviate substantially from the ground truth. Our approach leverages hierarchical information near the arms to aid in inference, effectively predicting the joints at the occluded locations.

\section{Limitation and Future Work} 
% 多帧的结合
% 物体遮挡，结合image context
% 数据集的获取
A current limitation of our approach is its reliance on the single-frame based methods~\cite{li2023pose} for the lifting model. Applying our method directly to the temporal-based lifting models~\cite{zhang2022mixste} would slow down inference, limiting our ability to further utilize temporal information and indicating room for improvement in our approach.
This is attributed to the expansion of input joint quantities. The conventional attention mechanisms have a computational complexity that scales significantly with the joint quantities.

Moving forward, we aim to develop a temporal-based lifting model compatible with hierarchical 2D poses. Our future work will involve strategies such as sampling the key joints from hierarchical 2D poses and optimizing the computation of attention mechanisms to manage the joint quantity increase effectively. By doing so, we expect to integrate the strengths of temporal-based models with our hierarchical pose approach, thereby improving the accuracy of pose estimation while maintaining computational efficiency.

Besides, Tab.~\textcolor{darkblue}{6} in the main paper shows that our method greatly improves single-frame lifting models more than multi-frame methods, suggesting considerable potential for optimizing how our densification approach combines with temporal information, which is straightforward in our experiments.
% we straightforwardly merged the new temporal dimension with the batch dimension for the densification process.
Firstly, each frame of the 2D pose sequence is fed into HiPART to generate hierarchical 2D poses. These poses are concatenated along the temporal and joint dimensions to form a tensor of size $T\times (J_s+J_d+J_f)\times2$, which is input for temporal-based lifting models to infer the final 3D pose.
In the future, we plan to delve into more effective densification methods for integrating hierarchical spatial and temporal information to better exploit their interplay and further boost the performance of 3D HPE.

% {
%     \small
%     \bibliographystyle{ieeenat_fullname}
%     \bibliography{main}
% }

\end{document}